\documentclass[namedate,webpdf]{ima-authoring-template-blank}
\usepackage[T1]{fontenc}
\usepackage{lmodern}
\usepackage{amssymb,amsmath,amsthm}
\usepackage{mathtools}
\usepackage{bm, bbm}
\usepackage{hyperref}
\usepackage{graphicx}
\usepackage{cases}
\usepackage{multirow, multicol, adjustbox, makecell}
\usepackage{array, booktabs}

\usepackage{tikz-cd}
\usepackage{algpseudocode, algorithm, algorithmicx}
\algrenewcommand\algorithmicrequire{\textbf{Input:}}
\algrenewcommand\algorithmicensure{\textbf{Output:}}

\newcommand*{\supp}{\mathrm{supp}}

\newcommand*{\argmin}[1]{{\underset{#1}{\operatorname{argmin}}}}

\newcommand*{\spn}{\mathrm{span}}
\newcommand*{\Lip}{\mathrm{Lip}}  % Lipschitz

\def\D{\,\mathrm{d}}

\newcommand*{\HS}{\textup{HS}}

\newcommand*{\nm}[1]{{\left\|#1\right \|}}
\newcommand*{\nmd}[1]{{\|#1\|}}    %default

\newcommand{\abs}[1]{\left|#1\right|}
\newcommand{\absd}[1]{|#1|}    %default

\newcommand*{\ip}[2]{\left\langle #1, #2 \right\rangle}
\newcommand*{\ipd}[2]{\langle #1, #2 \rangle}    %default
\newcounter{counter}
\numberwithin{counter}{section}
\numberwithin{equation}{section}

\theoremstyle{plain}
\newtheorem{theorem}[counter]{Theorem}
\newtheorem{lemma}[counter]{Lemma}

\newtheorem{proposition}[counter]{Proposition}
\newtheorem{assumption}[counter]{Assumption}
\newtheorem{result}{Result}

\newtheorem{remark}[counter]{Remark}

\newtheorem*{theorem*}{Theorem}
\newtheorem*{lemma*}{Lemma}
\newtheorem*{corollary*}{Corollary}
\newtheorem*{conjecture*}{Conjecture}
\newtheorem*{proposition*}{Proposition}
\newtheorem*{assumption*}{Assumption}
\newtheorem*{remark*}{Remark}
\newtheorem*{notation*}{Notation}

\theoremstyle{definition}
\newtheorem{definition}[counter]{Definition}
\newtheorem{example}[counter]{Example}

\newtheorem*{definition*}{Definition}
\newtheorem*{example*}{Example}
\newtheorem*{discussion*}{Discussion}

\newcommand*{\cD}{\mathcal{D}}
\newcommand*{\cE}{\mathcal{E}}
\newcommand*{\cF}{\mathcal{F}}

\newcommand*{\cH}{\mathcal{H}}

\newcommand*{\cK}{\mathcal{K}}
\newcommand*{\cL}{\mathcal{L}}
\newcommand*{\cM}{\mathcal{M}}
\newcommand*{\cN}{\mathcal{N}}

\newcommand*{\cP}{\mathcal{P}}

\newcommand*{\cT}{\mathcal{T}}
\newcommand*{\cV}{\mathcal{V}}

\newcommand*{\cX}{\mathcal{X}}
\newcommand*{\cY}{\mathcal{Y}}
\newcommand*{\cZ}{\mathcal{Z}}

\newcommand*{\bbE}{\mathbb{E}}

\newcommand*{\bbN}{\mathbb{N}}

\newcommand*{\bbR}{\mathbb{R}}

\newcommand*{\wtC}{\widetilde{C}}

\newcommand*{\wtF}{\widetilde{F}}

\newcommand*{\wtH}{\widetilde{H}}

\newcommand*{\wtY}{\widetilde{Y}}

\newcommand*{\wtd}{\widetilde{d}}

\newcommand*{\whH}{\widehat{H}}

\DefineNamedColor{named}{Purple}{cmyk}{0.45,0.86,0,0}

\DefineNamedColor{named}{JungleGreen} {cmyk}{0.99,0,0.52,0}

\DefineNamedColor{named}{orange} {rgb}{1,0.55,0}

\usepackage{natbib}
\usepackage{soul, cancel}

\usepackage[right, pagewise, mathlines]{lineno}

\theoremstyle{thmstyletwo}
\begin{document}

%\linenumbers

%\DOI{DOI HERE}
\copyrightyear{2026}
%\vol{00}
%\pubyear{2021}
%\access{Advance Access Publication Date: Day Month Year}
\appnotes{Paper}
%\copyrightstatement{Published by Oxford University Press on behalf of the Institute of Mathematics and its Applications. All rights reserved.}
%\firstpage{1}

%\subtitle{Subject Section}

\title[Sample Complexity of Learning Lipschitz Operators w.r.t. Gauss. Measures]{The Sample Complexity of Learning Lipschitz Operators with respect to Gaussian Measures}

\author{Ben Adcock
\address{\orgdiv{Department of Mathematics}, \orgname{Simon Fraser University}, \orgaddress{\street{8888 University Drive}, \postcode{V5A 1S6}, \state{Burnaby~BC}, \country{Canada}}}}
\author{Michael Griebel and Gregor Maier*
\address{\orgdiv{Institute for Numerical Simulation}, \orgname{University of Bonn}, \orgaddress{\street{Friedrich-Hirzebruch-Allee 7}, \postcode{53115 Bonn}, \country{Germany}}}
\address{\orgname{Fraunhofer Institute for Algorithms and Scientific Computing SCAI}, \orgaddress{\street{Schloss Birlinghoven 1}, \postcode{53754 Sankt Augustin}, \country{Germany}}}}

\authormark{Adcock et al.}

\corresp[*]{Corresponding author: \href{email:maier@ins.uni-bonn.de}{maier@ins.uni-bonn.de}}

\received{Date}{0}{Year}
\revised{Date}{0}{Year}
\accepted{Date}{0}{Year}

%\editor{Associate Editor: Name}

\abstract{
Operator learning, the approximation of mappings between infinite-dimensional function spaces using machine learning, has gained increasing research attention in recent years. Operator approximations can serve as efficient surrogate models for problems in computational science and engineering, complementing traditional methods. However, despite their empirical success, our understanding of the underlying mathematical theory is in large part still incomplete. 
In this paper, we study the approximation of Lipschitz operators with respect to Gaussian measures. We prove higher Gaussian Sobolev regularity of Lipschitz operators and establish lower and upper bounds on the Hermite polynomial approximation error. 
We then study general reconstruction strategies of Lipschitz operators from \(m\) arbitrary (potentially adaptive) linear samples. As a key finding, we tightly characterize the corresponding sample complexity, that is, the smallest achievable worst-case error among all possible choices of (adaptive) sampling and reconstruction strategies, in terms of~\(m\). As a consequence, we identify an inherent curse of sample complexity: No method to approximate Lipschitz operators based on \(m\) linear samples can achieve algebraic convergence rates in \(m\). On the positive side, we prove that a sufficiently fast spectral decay of the covariance operator of the underlying Gaussian measure guarantees convergence rates which are arbitrarily close to any algebraic rate. Overall, by tightly characterizing the sample complexity, our work confirms the intrinsic difficulty of learning Lipschitz operators, regardless of the data or learning technique.}

\keywords{Operator learning; Sample complexity; Lipschitz operators; Gaussian measures.}

\maketitle

%%%%%%%%%%%%%%%%%%%%%%%%%%%%%%%%%%%%%%%%%%%%%%%%%%%%%%%%%%%%%%%%%%%%%%%%%%%%%%%%%%%%%%%%%%
%%%%%%%%%%%%%%%%%%%%%%%%%%%%%%%%%%%%%%%%%%%%%%%%%%%%%%%%%%%%%%%%%%%%%%%%%%%%%%%%%%%%%%%%%%

\section{Introduction}

We study the approximation of generic Lipschitz operators which map between (infinite-) dimensional Hilbert spaces. The approximation error is measured in \(L^2\) with input samples drawn from a Gaussian probability distribution. We commence with a detailed literature review in Section~\ref{subsec: motivation and literature review}, where we put our work in the context of operator learning and motivate the Gaussian setting as a natural framework for analyzing Lipschitz operators. We subsequently summarize our main contributions in Section~\ref{subsec: contributions} and give an overview of the organization of the remainder of the paper. Limitations and future work are discussed in Section~\ref{subsec: limitations and future work}. Basic notation is recalled in Section~\ref{subsec: notation}.

\subsection{Motivation and literature review}
\label{subsec: motivation and literature review}

With the rise of machine learning, in particular deep learning, in computational science and engineering (CSE), operator learning has recently emerged as a new paradigm for the data-driven approximation of mappings between infinite-dimensional function spaces. Multiple deep learning architectures, typically referred to as \emph{neural operators}, such as DeepONet~\citep{lu_LearningNonlinearOperatorsDeepONet_2021}, FNO~\citep{li_FourierNeuralOperatorParametric_2021}, general neural operators~\citep{kovachki_NeuralOperatorLearningMaps_2023}, and PCA-Net~\citep{bhattacharya_ModelReductionNeuralNetworks_2021}, have been proposed and their efficiency has been demonstrated in various practical applications. We refer to the recent reviews by~\citet{kovachki_OperatorLearningAlgorithmsAnalysis_2024} and~\citet{boulle_MathematicalGuideOperatorLearning_2024} and references therein. However, the empirical success of these neural operators has to a large extent not yet been supported by a general mathematical theory. A thorough understanding of theoretical approximation guarantees is important though for a reliable deployment of operator learning methods in CSE applications.

\subsubsection{Theory of operator learning}

A typical starting point in the theoretical analysis of operator learning are universal approximation results. They guarantee the existence of a neural operator of a certain type which approximates a target operator up to some arbitrarily small error~\citep{chen_UniversalApproximationNonlinearOperators_1995, lanthaler_ErrorEstimatesDeepONetsDeep_2022, kovachki_UniversalApproximationErrorBounds_2021, lanthaler_NonlocalityNonlinearityImpliesUniversality_2025, lanthaler_OperatorLearningPCANetUpper_2023}. Albeit necessary for assessing the basic utility of a neural operator, mere existence results are of limited use in practical applications, where instead questions about quantitative approximation guarantees and explicit convergence rates are of greater importance. 

To address the latter, two quantities are of key interest. On the one hand, the \emph{parametric complexity} quantifies the convergence of the approximation error in terms of the number of tunable parameters employed by the approximation method. In the context of (deep) neural network (NN) approximations, this is often referred to as \emph{expression rates}. On the other hand, the \emph{sample complexity} quantifies the convergence of the approximation error in terms of the number of samples used for fitting the parameters to data. Previous research efforts mainly focused on deriving expression rates for NN approximations of specific (classes of) operators, whereas there has been comparably little work on sample complexity estimates.

Holomorphic operators have been widely studied in operator learning. They arise, for example, as parameter-to-solution mappings for parametric partial differential equations (PDEs), see, e.g.,~\citet{cohen_ApproximationHighdimensionalParametricPDEs_2015} and~\mbox{\citet[Ch. 4]{adcock_SparsePolynomialApproximationHighDimensional_2022}} and references therein. Such operators can be learned with algebraic or (on finite-dimensional domains) even exponential parametric complexity with NNs~\citep{opschoor_ExponentialReLUDNNExpression_2022, herrmann_NeuralSpectralOperatorSurrogates_2024, schwab_DeepLearningHighDimension_2023}. Moreover, they can be approximated with (near-)optimal sample complexity with least-squares and compressed sensing methods~\citep{adcock_EfficientAlgorithmsComputingNearBest_2024, bartel_SamplingRecoveryBochnerSpaces_2024, adcock_OptimalApproximationInfinitedimensionalHolomorphic_2024, adcock_OptimalApproximationInfinitedimensionalHolomorphic_2025} as well as with NNs~\citep{adcock_OptimalDeepLearningHolomorphic_2024, 
adcock_LearningSmoothFunctionsHigh_2024,
adcock_NearoptimalLearningBanachvaluedHighdimensional_2025}.
We mention in passing that algebraic NN expression rates and exponential sample complexity estimates (in probability) have also been derived for classes of (non-holomorphic) solution operators of certain PDEs~\citep{ryck_GenericBoundsApproximationError_2022, boulle_EllipticPDELearningProvably_2023}. Moreover, algebraic complexity estimates are also available for infinite-dimensional functionals (with one-dimensional codomain) with mixed regularity~\citep{dung_HyperbolicCrossApproximationInfinite_2016}.

\subsubsection{Analysis of Lipschitz operators}

However, many operators fail to be holomorphic. Among families of non-holomorphic operators, Lipschitz operators---which arise, for example, as parameter-to-solution mappings for parametric elliptic variational inequalities---are an important class. Applications are numerous and include, inter alia, obstacle problems in mechanics and control problems, as well as uncertainty quantification in equilibrium models. The respective operators in these contexts generically do not exhibit any regularity beyond Lipschitz continuity. We refer to~\citet{schwab_DeepOperatorNetworkApproximation_2026} and~\citet[Ch. 6]{gwinner_UncertaintyQuantificationVariationalInequalities_2021} and references therein for further details.

As a specific example, consider the following obstacle problem on the unit square \( \Omega = (0,1)^2 \): Given an obstacle \( v \in H_0^1(\Omega) \) and an external force \( f \in L^2(\Omega) \), find a function \( u \in H_0^1(\Omega) \) with minimal energy \( \frac{1}{2} \int_{\Omega} \nm{ \nabla u }^2 \D \bm{x} - \int_{\Omega} f u \D \bm{x} \) which stays above \( v \). This problem asks for the resulting shape of an elastic membrane \( u \) that is spanned over an obstacle \( v \) under an external load \( f \) and vanishing Dirichlet boundary conditions~\citep[Ch. 1]{rodrigues_ObstacleProblemsMathematicalPhysics_1987}. The associated obstacle-to-solution operator
\begin{align}
\label{eq: obstacle-to-solution operator}
\begin{split}
    F :  H_0^1(\Omega) &\to H_0^1(\Omega), \\
         v & \mapsto F(v) = \argmin{} \left\{ \frac{1}{2} \int_{\Omega} \nm{\nabla u}^2 \D \bm{x} - \int_{\Omega} f u \D \bm{x} : u \in H_0^1(\Omega), u \geq v \textup{ a.e. in } \Omega \right\}
\end{split}
\end{align}
maps any obstacle \( v \) to the corresponding solution \( u = F(v) \) of the obstacle problem. It is known that this operator is Lipschitz continuous but not Fréchet differentiable~\citep{mignot_ControleDansInequationsVariationelles_1976}.

Obstacle-to-solution operators similar to~\eqref{eq: obstacle-to-solution operator} appear in practice in, for example, contact problems in elasticity, fluid filtration through a porous medium, and ingot solidification in continuous casting. We refer to~\citet{rodrigues_ObstacleProblemsMathematicalPhysics_1987} for further details. Other practical applications of obstacle problems include, among many others, ice sheet modeling~\citep{jouvet_SteadyShallowIceSheets_2012} as well as option pricing in mathematical finance~\citep{jaillet_VariationalInequalitiesPricingAmerican_1990}.

The study of approximating Lipschitz operators has recently attracted increased research attention, as they exhibit fundamentally different properties from those of holomorphic operators in terms of learnability. It was shown in~\citet{lanthaler_OperatorLearningPCANetUpper_2023} that bounded (with respect to the uniform norm) Lipschitz (and \(C^k\)-Fréchet differentiable) operators cannot be approximated with algebraic parametric complexity using PCA-Net. More specifically, the number of PCA-Net parameters scales exponentially with the inverse of the approximation error. This result, termed the \emph{curse of parametric complexity}, can be interpreted as the infinite-dimensional analogue to the classical curse of dimensionality in finite dimensions, see also~\citet{lanthaler_ParametricComplexityOperatorLearning_2026}. It can seemingly be overcome by neural operators which use hyper-expressive activation functions or non-standard NN architectures~\citep{schwab_DeepOperatorNetworkApproximation_2026}. In practical implementations, however, each real-valued parameter can only be represented by a sequence of bits of finite length. 
In~\citet{lanthaler_OperatorLearningLipschitzOperators_2024}, the cost model of counting real-valued parameters was therefore replaced by instead counting the number of bits that are necessary to digitally encode all parameters to some finite accuracy. The resulting cost-accuracy scaling law reveals a curse of parametric complexity that is independent of the activation functions used in any NN approximation. It states that the number of bits required to encode all NN parameters, in fact, still scales exponentially with the inverse of the approximation error. 

It was shown in ~\citet{kovachki_DataComplexityEstimatesOperator_2024} that bounded Lipschitz and \(C^k\)-operators also exhibit a \emph{curse of data complexity}. That is, the \(L^2\)-error with respect to a Gaussian measure with at most algebraically decaying PCA eigenvalues converges at most logarithmically in the number of samples for any learning algorithm which is based on i.i.d. pointwise samples. 
However, Lipschitz operators that arise in practical applications as mentioned above are usually not bounded and the just reviewed results are thus not directly applicable. For example, it is immediate to see that the obstacle-to-solution operator in~\eqref{eq: obstacle-to-solution operator} is not bounded.

\subsubsection{Analysis of Gaussian measures}

In the present paper, we prove a \emph{curse of sample complexity} which generalizes the result of~\citet{kovachki_DataComplexityEstimatesOperator_2024} to unbounded Lipschitz operators and arbitrary (centered, nondegenerate) Gaussian measures. Unlike in previous studies, we tightly characterize the sample complexity of learning such operators in terms of the PCA eigenvalues of the (covariance operator of the) underlying Gaussian measure. We then show that algebraic convergence cannot be achieved, regardless of the decay of these eigenvalues. Nevertheless, we show that with sufficiently fast spectral decay, error decay rates which are arbitrarily close to any algebraic rate can be achieved.
We mention the recent work of~\citet{liu_DeepNonparametricEstimationOperators_2024} for further results in the statistical theory of deep nonparametric estimation of Lipschitz operators. Therein, however, the authors work with probability measures with compact support. Consequently, their results are not directly applicable to Gaussian measures.

Our work focuses on Gaussian measures, as they are the typical choice of input measures. They also allow us to draw on results from infinite-dimensional analysis~\citep{bogachev_GaussianMeasures_1998, daprato_IntroductionInfiniteDimensionalAnalysis_2006,daprato_IntroductionStochasticAnalysisMalliavin_2014, lunardi_InfiniteDimensionalAnalysis_2015}. The theory of Gaussian Sobolev spaces is key in our analysis as it is well-known that Lipschitz functionals are Gaussian Sobolev functionals. This connection yields explicit control over error bounds in terms of the spectral properties of the Gaussian measure.
The Gaussian setting has been considered previously to prove expression rates for NN approximations of operators, see, e.g.,~\citet{schwab_DeepLearningHighDimension_2023} and~\citet{dung_DeepReLUNeuralNetwork_2023}. It can also be studied within the abstract framework developed in~\citet{griebel_StableSplittingsHilbertSpaces_2017}, see Example 1 therein, to derive dimension-independent results for the approximation of high- and infinite-dimensional function(al)s. Its connection to Lipschitz regularity, however, has, to the best of our knowledge, not yet been exploited to derive sample complexity estimates for Lipschitz operators.

\subsection{Contributions}
\label{subsec: contributions}

Let \(\cX, \cY\) be separable Hilbert spaces with \(\dim(\cX) = \infty\) and let \(\mu\) be a Gaussian measure on \(\cX\) with PCA eigenvalues \( \bm{\lambda} = ( \lambda_i)_{i = 1}^{\infty} \). We now give an (informal) overview of our three main contributions. \\

\noindent\textit{1.}
In Section~\ref{sec: Gaussian Sobolev and Lipschitz operators}, we extend standard results from infinite-dimensional analysis and define the (weighted) Gaussian Sobolev space \(W_{\mu, \bm{b}}^{1, 2}(\cX; \cY)\) by means of a sequence \(\bm{b} = (b_i)_{i = 1}^{\infty}\) of positive bounded weights. As our first main contribution, we show that this space contains the set \(\Lip(\cX; \cY)\) of (potentially unbounded) Lipschitz operators which map from \(\cX\) to \(\cY\):
\begin{result}[Lipschitz operators are Gaussian Sobolev operators (Theorem~\ref{thm: Lipschitz operators are Gaussian Sobolev operators})]
\label{res: Lipschitz operators are Gaussian Sobolev operators}
     If \(\cY\) is finite-dimensional, then \(\Lip(\cX; \cY) \subset W_{\mu, \bm{b}}^{1,2}(\cX; \cY)\).
     If \(\cY\) is infinite-dimensional and if \(\bm{b} \in \ell^2(\bbN)\), then \(\Lip(\cX; \cY) \subset W_{\mu, \bm{b}}^{1,2}(\cX; \cY)\). In both cases, the space \(C^{0,1}(\cX; \cY)\) of bounded Lipschitz operators is continuously embedded in \(W_{\mu, \bm{b}}^{1,2}(\cX; \cY)\).
\end{result}
This result is known in the unweighted case (\( b_i = 1 \) for all \( i \)) if \( \cY = \bbR \) and can be easily generalized to the vector-valued case \( \cY = \bbR^n \). For \( \dim(\cY) = \infty \), the weighted setting provides the correct structure for our analysis: The condition \( \bm{b} \in \ell^2(\bbN) \) gives control of the degree of (weak) differentiability of the operators in all (infinitely many) coordinate directions.

Result~\ref{res: Lipschitz operators are Gaussian Sobolev operators} is crucial in our subsequent analysis. Elements in \(W_{\mu, \bm{b}}^{1, 2}(\cX; \cY)\) are characterized as operators whose polynomial expansion coefficients with respect to the (infinite-dimensional, \(\bm{\lambda}\)-rescaled) Hermite polynomials \(\{H_{\bm{\gamma}, \bm{\lambda}}\}_{\bm{\gamma} \in \Gamma}\), with countable index set \(\Gamma\), are weighted \(\ell^2\)-summable. The corresponding weights \(\bm{u} = (u_{\bm{\gamma}})_{\bm{\gamma} \in \Gamma}\) are given in terms of the (\(\bm{b}\)-weighted) PCA eigenvalues \(\lambda_{\bm{b}, i} = \lambda_i / b_i^2\). As a result, we can study the approximation of a given Lipschitz operator by considering its Hermite polynomial expansions. \\

\noindent\textit{2.}
In Section~\ref{sec: polynomial s-term approximation}, we give upper and lower bounds for the convergence of these expansions in terms of the PCA eigenvalues. 
In particular, we show the following \emph{curse of parametric complexity}: No \(s\)-term Hermite polynomial expansion can converge with an algebraic rate uniformly for all Lipschitz operators as \(s \to \infty\). This holds regardless of the decay rate of the PCA eigenvalues.
More specifically, let \(S \subset \Gamma\) be a finite index set with at most \(s\) elements and let \(F_S\) denote the polynomial approximation of an operator \(F \in W_{\mu, \bm{b}}^{1, 2}(\cX; \cY)\) by Hermite polynomials \( H_{\bm{\gamma}, \bm{\lambda}} \) with \( \bm{\gamma} \in S \). Moreover, let \(\pi: \bbN \to \Gamma\) be a bijection such that \((u_{\bm{\pi(i)}})_{i=1}^{\infty}\) is a nonincreasing rearrangement of \(\bm{u}\). Our second main contribution is the following result:

\begin{result}[Curse of parametric complexity (Theorems~\ref{thm: s-term: best s-term approximation error},~\ref{thm: s-term: lower bound for u_pi(s+1)},~\ref{thm: s-term: upper bounds for u_pi(s+1)})]
\label{res: best polynomial s-term error; informal}
    For every \(s \in \bbN\), we have
    \begin{align*}
        \inf_{S \subset \Gamma, \abs{S} \leq s} \sup_F
        \nm{F - F_S}_{L_{\mu}^2(\cX; \cY)} 
        = \sup_F
        \nm{F - F_{\{\bm{\pi(1)}, \dots, \bm{\pi(s)}\}}}_{L_{\mu}^2(\cX; \cY)}
        = u_{\bm{\pi(s + 1)}},
    \end{align*}
    where the suprema are taken either over the class of all Lipschitz operators or all \(W_{\mu, \bm{b}}^{1, 2}\)-operators with \(\nm{F}_{W_{\mu, \bm{b}}^{1, 2}(\cX; \cY)} \leq 1\) in both cases. 
    Moreover, \(u_{\bm{\pi(s + 1)}}\) cannot decay with algebraic order as \(s \rightarrow \infty\), regardless of the spectral properties of \(\mu\). On the positive side, the decay of \(u_{\bm{\pi(s + 1)}}\) can become arbitrarily close to any algebraic rate if the PCA eigenvalues \(\lambda_{\bm{b}, i}\) decay sufficiently fast (e.g., double exponentially).
\end{result}

As we will discuss in Section~\ref{subsec: s-term: discussion}, this curse of parametric complexity is consistent with the one from~\cite{lanthaler_OperatorLearningPCANetUpper_2023}, which we mentioned in Section~\ref{subsec: motivation and literature review}.\\

\noindent\textit{3.}
Up to this point, we study best polynomial approximation of Lipschitz operators. In Section~\ref{sec: optimal sampling and adaptive m-widths}, we adopt a more general view and consider learning Lipschitz and \(W_{\mu, \bm{b}}^{1, 2}\)-operators from finitely many arbitrary linear samples. Using tools from~\emph{information-based complexity} (IBC) (see~\citet{novak_TractabilityMultivariateProblemsVolume_2008}), we define the adaptive \(m\)-width \(\Theta_m(\cK)\) of a set  \(\cK\) of operators. It quantifies the smallest worst-case error that can be achieved by learning operators in \(\cK\) from \(m\) measurements \(\cL(F) \in \cY^m\) which are generated by an adaptive sampling operator \(\cL\) and used as inputs for an arbitrary (generally nonlinear) reconstruction mapping \(\cT: \cY^m \to L_{\mu}^2(\cX; \cY)\). Specifically,
\[
\Theta_m(\cK)
:= \inf\left\{\sup_{F \in \cK} \nm{F - \cT(\cL(F))}_{L_{\mu}^2(\cX; \cY)}: \cL, \cT \textup{ as above} \right\}.
\]
Our key result is the characterization of \(\Theta_m(\cK)\) in terms of the weights \(u_{\bm{\gamma}}\):
\begin{result}[Characterization of the adaptive \(m\)-width (Theorems~\ref{thm: s-term: lower bound for u_pi(s+1)},~\ref{thm: s-term: upper bounds for u_pi(s+1)},~\ref{thm: adaptive m-width})]
\label{res: adaptive m-width; informal}
    For any number of samples \(m \in \bbN\), we have
    \[
    \Theta_m(\cK) = u_{\bm{\pi(m + 1)}},
    \]
    where \(\cK\) is either the set of all Lipschitz or all \(W_{\mu, \bm{b}}^{1,2}\)-operators of at most unit \(W_{\mu, \bm{b}}^{1,2}\)-norm in both cases. Again, \(u_{\bm{\pi(m + 1)}}\) cannot decay with algebraic order as \(m \rightarrow \infty\), regardless of the spectral properties of \(\mu\). But its decay can become arbitrarily close to any algebraic rate if the PCA eigenvalues \(\lambda_{\bm{b}, i}\) of \(\mu\) decay sufficiently fast (e.g., double exponentially).
\end{result}
This result tightly characterizes the sample complexity of learning Lipschitz operators and gives rise to the following \emph{curse of sample complexity}: No procedure (e.g., NNs, polynomial approximation, random features, kernel methods, etc.) for learning Lipschitz operators can achieve algebraic convergence rates for the smallest worst-case \(L_{\mu}^2\)-approximation error. This holds for general (centered, nondegenerate) Gaussian measures \(\mu\).
In light of Result~\ref{res: adaptive m-width; informal}, notice that Result~\ref{res: best polynomial s-term error; informal} shows that Hermite polynomial approximation is optimal among all possible (adaptive) sampling and reconstruction operators for approximating Lipschitz and \(W_{\mu, \bm{b}}^{1,2}\)-operators. This sharpens, in particular, the ``adaptivity does not help'' principle from IBC~\citep{novak_PowerAdaption_1996}, in that it shows that in this case adaptivity in the \(\cY\)-valued samples does not even help to improve any constants in the error bounds. Finally, we emphasize that Result~\ref{res: adaptive m-width; informal} implies that from an IBC point of view, there is no difference between the space \(\Lip(\cX; \cY)\), equipped with the \(W_{\mu, \bm{b}}^{1,2}\)-norm, and the whole space \(W_{\mu, \bm{b}}^{1,2}(\cX; \cY)\), as the adaptive \(m\)-widths of the unit balls in both spaces coincide.

\subsection{Limitations and future work}
\label{subsec: limitations and future work}

In this work, we do not consider reconstruction strategies based on i.i.d. pointwise samples but based on general linear information. 
In practice, however, recovery from pointwise samples is most relevant as it is non-intrusive. While the inherent curse of sample complexity cannot be overcome, it is of key interest to design algorithms with (near-)optimal sample complexity which achieve an optimal approximation error with the minimum number of operator evaluations. Moreover, in practice, only finite-dimensional data is available, which is usually corrupted by noise.  Hence, encoding and decoding errors and errors due to noisy observations arise. We aim to address this problem in future work by studying the approximation of Lipschitz operators from noisy pointwise samples with methods from machine learning, e.g., in terms of practical existence theorems, as studied in~\citet{adcock_LearningSmoothFunctionsHigh_2024, adcock_OptimalDeepLearningHolomorphic_2024, adcock_NearoptimalLearningBanachvaluedHighdimensional_2025} and~\citet{franco_PracticalExistenceTheoremReduced_2025}, or with techniques from statistical learning theory (see, e.g.,~\citet{schmidt-hieber_NonparametricRegressionUsingDeep_2020} and~\cite{liu_DeepNonparametricEstimationOperators_2024}). 

On the other hand, by tightly characterizing the curse of sample complexity, our results suggest that learning Lipschitz operators with arbitrary Gaussian measures may be practically challenging.
Extending and corroborating previous studies (e.g., \citet{kovachki_DataComplexityEstimatesOperator_2024, lanthaler_ParametricComplexityOperatorLearning_2026}), our results suggest that the class of Lipschitz operators may simply be too large. By contrast, the class of holomorphic operators is known to be too small since operators found in various applications are non-holomorphic. Hence, our results highlight another important problem for future work, namely, the search for spaces that better describe classes of operators found in applications.

\subsection{Basic notation}
\label{subsec: notation}

As usual, \(\bbN\) denotes the set of all positive integers, \(\bbN_0\) the set of all nonnegative integers, and \(\bbR\) the real numbers.
We denote by \((\cX, \ip{\cdot}{\cdot}_{\cX})\) and \((\cY, \ip{\cdot}{\cdot}_{\cY})\) two separable Hilbert spaces with corresponding inner products. For simplicity, we focus on real Hilbert spaces, but all results can be readily generalized to the complex case as well. We assume that \(\dim(\cX) = \infty\), unless stated otherwise. We use capital letters \(X \in \cX\) and \(Y \in \cY\) (at times also \(H, K, Z\))  for elements of the respective Hilbert spaces. Operators which map from \(\cX\) to \(\cY\) are typically denoted by the capital letters \(F\) or \(G\). For functionals, i.e., in the case \(\cY = \bbR\), we also use lowercase letters at times.
We equip \(\cX\) with a centered, nondegenerate Gaussian measure \(\mu\) with covariance operator \(\Sigma := \int_{\cX}{X \otimes X \D \mu(X)}\). We recall that \(\Sigma: \cX\to \cX\) is positive definite, self-adjoint, and trace class~\citep[Prop. 1.8]{daprato_IntroductionInfiniteDimensionalAnalysis_2006}. As such, there exists an orthonormal eigenbasis (PCA basis) \(\{\phi_i\}_{i = 1}^{\infty}\) of \(\cX\) and a sequence of corresponding PCA eigenvalues \(\bm{\lambda} = (\lambda_i)_{i = 1}^{\infty}\), which satisfy \( \sum_{i = 1}^{\infty} \lambda_i < \infty \).

%%%%%%%%%%%%%%%%%%%%%%%%%%%%%%%%%%%%%%%%%%%%%%%%%%%%%%%%%%%%%%%%%%%%%%%%%%%%%%%%%%%%%%%%%%
%%%%%%%%%%%%%%%%%%%%%%%%%%%%%%%%%%%%%%%%%%%%%%%%%%%%%%%%%%%%%%%%%%%%%%%%%%%%%%%%%%%%%%%%%%

\section{Gaussian Sobolev and Lipschitz operators}
\label{sec: Gaussian Sobolev and Lipschitz operators}

The results in this section are the basis for the approximation theoretical analysis carried out in the remainder of this paper. 
We state the definition of the weighted Gaussian Sobolev space \(W_{\mu, \bm{b}}^{1 ,2}(\cX; \cY)\) and its characterization as a weighted \(\ell^2\)-space in Section~\ref{subsec: the space W^12_mu}. %The underlying principle for the latter is that by expanding operators in a series of Hermite polynomials, Sobolev regularity can be characterized as weighted summability of the corresponding Hermite coefficients. 
We recall necessary notions in Section~\ref{subsec: preliminaries} and provide technical details in Appendix~\ref{subsec: app: the Gaussian Sobolev space}. In Section~\ref{subsec: Lipschitz operators}, we prove that all Lipschitz operators \(\cX \to \cY\) lie in \(W_{\mu, \bm{b}}^{1 , 2}(\cX; \cY)\).

\subsection{Preliminaries}
\label{subsec: preliminaries}

The space \(L_{\mu}^2(\cX; \cY)\) denotes the Lebesgue-Bochner space of (equivalence classes of) strongly measurable operators \(F: \cX \to \cY\) with finite Bochner norm
\[
\nm{F}_{L_{\mu}^2(\cX; \cY)} := \left( \int_{\cX} \nm{F(X)}_{\cY}^2 \D \mu(X) \right)^{1/2}.
\]
If \(\cY = \bbR\), the space \(L_{\mu}^2(\cX; \bbR)\) coincides with the usual Lebesgue space and we write \(L_{\mu}^2(\cX) := L_{\mu}^2(\cX; \bbR)  \). See, e.g.,~\citet[Ch. 1]{hytonen_AnalysisBanachSpacesVolume_2016} for more information.

Next, we introduce the (infinite-dimensional) Hermite polynomials. For \(n \in \bbN_0\), we first recall the \(n\)th normalized (probabilist's) Hermite polynomial on \(\bbR\), given by
\[
H_n: \bbR \to \bbR, \quad H_n(x) := \frac{(-1)^n}{\sqrt{n!}} \exp \left( \frac{x^2}{2} \right) \frac{d^n}{d x^n} \exp \left( -\frac{x^2}{2} \right).
\]
The family \(\{H_n\}_{n = 0}^{\infty}\) is an orthonormal basis of \(L_{\mu_1}^2(\bbR)\) with \( \mu_1 = \cN(0,1) \)~\citep[Prop. 9.4]{daprato_IntroductionInfiniteDimensionalAnalysis_2006}. The higher-dimensional Hermite polynomials are defined as products of the one-dimensional ones. To this end, we introduce the set of all sequences of nonnegative integers with finite support,
\[
\Gamma := \left\{\bm{\gamma}\in\bbN_0^{\bbN} : \abs{\supp(\bm{\gamma})} < \infty\right\},
\]
with \(\supp(\bm{\gamma}) := \{i \in \bbN : \gamma_i \neq 0\}\). It is easy to see that \(\Gamma\) is countable.
For \(\bm{\gamma} \in \Gamma\), \(d \in \bbN\), we set
\begin{equation}
\label{eq: definition of Hermite polynomials on bbR^d}
    H_{\bm{\gamma}, d}: \bbR^d \to \bbR, \quad 
    H_{\bm{\gamma}, d}(\bm{x}) := \prod_{i = 1}^d H_{\gamma_i}(x_i).
\end{equation}
We define Hermite polynomials on the infinite-dimensional space \(\cX\) by means of the PCA basis \(\{\phi_i\}_{i = 1}^{\infty}\) and the PCA eigenvalues \(\bm{\lambda} = (\lambda_i)_{i = 1}^{\infty}\):
\begin{equation}
\label{eq: definition of inf.-dim. Hermite polynomials on cX}
    H_{\bm{\bm{\gamma}}, \bm{\lambda}} : \cX \to \bbR, \quad 
    H_{\bm{\bm{\gamma}}, \bm{\lambda}}(X) := \prod_{i = 1}^{\infty} H_{\bm{\bm{\gamma}}_i}\left(\frac{\ip{X}{\phi_i}_{\cX}}{\sqrt{\lambda_i}}\right).
\end{equation}
The family \(\{H_{\bm{\gamma}, \bm{\lambda}}\}_{\bm{\gamma}\in\Gamma}\) is an orthonormal basis of \(L_{\mu}^2(\cX)\)~\citep[Thm. 9.7]{daprato_IntroductionInfiniteDimensionalAnalysis_2006}.
Recalling that
\(
L_{\mu}^2(\cX; \cY) = L_{\mu}^2(\cX) \otimes \cY 
\)
with Hilbertian tensor product, any \(F \in L_{\mu}^2(\cX; \cY)\) can thus be written as an unconditionally \(L_{\mu}^2(\cX; \cY)\)-convergent expansion in Hermite polynomials, also called the~\emph{Wiener-Hermite Polynomial Chaos (PC) expansion},
\begin{equation}
\label{eq: polynomial expansion of L2 operators}
    F = \sum_{\bm{\gamma} \in \Gamma} Y_{\bm{\gamma}} H_{\bm{\gamma}, \bm{\lambda}}
    \quad \textup{with} \quad Y_{\bm{\gamma}} := \int_{\cX} F(X) H_{\bm{\gamma}, \bm{\lambda}}(X) \D \mu(X) \in \cY.
\end{equation}
The \(Y_{\bm{\gamma}}\) are called the~\emph{Wiener-Hermite PC coefficients} of \(F\).
Moreover, \emph{Parseval's identity} holds, that is,
\begin{equation}\label{eq: Parseval's identity}
    \nm{F}_{L_{\mu}^2(\cX; \cY)}^2 = \sum_{\bm{\gamma} \in \Gamma} \nm{Y_{\bm{\gamma}}}_{\cY}^2.
\end{equation}

\subsection{The space \texorpdfstring{\(W_{\mu, \bm{b}}^{1, 2}(\cX; \cY)\)}{W\_\{mu, b\}\string^\{1,2\}(X;Y)}}
\label{subsec: the space W^12_mu}

The definition of the space \(W_{\mu, \bm{b}}^{1, 2}(\cX; \cY)\) is based on the Fréchet differential operator \(D_{\cX_{\bm{b}}}\) along a particular weighted space \(\cX_{\bm{b}}\) (see Appendix~\ref{sec: app: Notions of differentiability} for background on notions of differentiability). 
We commence with the definition of the latter.
Let \(\bm{b} = (b_i)_{i=1}^{\infty} \in \ell^{\infty}(\bbN)\), \(\bm{b} > \bm{0}\) be a sequence of positive bounded weights. Here and throughout the text, \( \bm{0} \) denotes the constant zero sequence and inequalities between sequences are understood componentwise. By means of the PCA basis \(\{\phi_i\}_{i=1}^{\infty}\) of \(\cX\) we define
\begin{equation}
\label{eq: weighted space cX_b}
    \cX_{\bm{b}} := \left\{X \in \cX: \sum_{i = 1}^{\infty} b_i^{-2} \abs{\ip{X}{\phi_i}_{\cX}}^2 < \infty\right\}.
\end{equation}
Note that \(\cX_{\bm{b}}\) is a Hilbert subspace of \(\cX\) with inner product and induced norm given by, respectively,
\[
\ip{X}{Z}_{\cX_{\bm{b}}} := \sum_{i = 1}^{\infty} b_i^{-2} \ip{X}{\phi_i}_{\cX} \ip{Z}{\phi_i}_{\cX} \quad \textup{and} \quad 
\nm{X}_{\cX_{\bm{b}}} := \sqrt{\ip{X}{X}_{\cX_{\bm{b}}}}
\]
for all \( X, Z \in \cX_{\bm{b}} \). It is easy to see that \(\cX_{\bm{b}}\) has an orthonormal basis \(\{\eta_i\}_{i=1}^{\infty}\), given by
\begin{equation}
\label{eq: orthonormal basis of cX_b}
    \eta_i := b_i \phi_i, \quad i\in\bbN.
\end{equation}
The role of \( \cX _{\bm{b}} \) will become clear in light of Theorem~\ref{thm: Lipschitz operators are Gaussian Sobolev operators} below. See also Remark~\ref{rmk: weighted Sobolev spaces}. It allows us to control the \( L^2_{\mu} \)-norm of the weak derivatives of a Sobolev operator \( \cX \to \cY \) in the case of infinite-dimensional output spaces \( \cY \).
\begin{example}
\label{ex: important cases of cX_b}
    If \(\bm{b} = \bm{1}\), we recover the full space \(\cX_{\bm{1}} = \cX\). If \( \bm{b} = \sqrt{\bm{\lambda}} := (\sqrt{\lambda_i})_{i = 1}^{\infty} \), we obtain the Cameron-Martin space \(\cX_{\sqrt{\bm{\lambda}}} = \cH\) of \(\mu\), see~\eqref{eq: orthonormal basis of cH}. We remark, however, that the choice \( \bm{b} = \sqrt{\bm{\lambda}} \) is excluded by the assumptions we put on \( \bm{b} \), see Assumption~\ref{ass: properties of b} and Example~\ref{ex: choices for b}. While the choice \( \bm{b} = \sqrt{\bm{\lambda}}  \) is not needed for our analysis, the connection to the Cameron-Martin space of \( \mu \) is nevertheless essential as it allows us to use results from infinite-dimensional analysis. See Appendix~\ref{sec: app: results from infinite-dimensional analysis} for details.
\end{example}

Let us now turn to the definition of the space \(W_{\mu, \bm{b}}^{1, 2}(\cX; \cY)\). We start with defining \(D_{\cX_{\bm{b}}}\) as an operator which maps from the set \(\cF C_b^1(\cX; \cY)\) of boundedly differentiable cylindrical operators to \(L_{\mu}^2(\cX; \HS(\cX_{\bm{b}}; \cY))\). Here and throughout the text, \( \HS(\cX_{\bm{b}}; \cY) \) denotes the space of all Hilbert-Schmidt operators from \( \cX_{\bm{b}} \) to \( \cY \), and we write \( \HS(\cX_{\bm{b}}) := \HS(\cX_{\bm{b}}; \bbR) \). The theory of the Cameron-Martin space \(\cH\) of \(\mu\) allows us to show that \(D_{\cX_{\bm{b}}}\) is closable in \(L_{\mu}^2(\cX; \cY)\). Details are provided in Appendix~\ref{sec: app: results from infinite-dimensional analysis}. See, in particular, Proposition~\ref{prop: closability of differential operator}. Relevant background on the closability and closure of operators is recalled in Appendix~\ref{sec: app: closability and closure of operators}.

\begin{definition}[The Sobolev space \(W_{\mu, \bm{b}}^{1,2}(\cX; \cY)\)]
\label{def: Gaussian Sobolev space}
    We define the space \(W_{\mu, \bm{b}}^{1,2}(\cX; \cY)\) as the domain of the closure of the operator \(D_{\cX_{\bm{b}}}: \cF C_b^1(\cX; \cY) \to L_{\mu}^2(\cX; \HS(\cX_{\bm{b}}; \cY))\) in \(L_{\mu}^2(\cX; \cY)\) (still denoted by \(D_{\cX_{\bm{b}}}\)).
\end{definition}

The space \(W_{\mu, \bm{b}}^{1,2}(\cX; \cY)\) is a Hilbert space with norm
\[
\nm{F}_{W_{\mu, \bm{b}}^{1,2}(\cX;\cY)} 
:= \left(\int_{\cX} \nm{F(X)}_{\cY}^2 \D \mu(X)
+ \int_{\cX} \nm{D_{\cX_{\bm{b}}} F(X)}_{\HS(\cX_{\bm{b}}; \cY)}^2 \D \mu(X)\right)^{1/2},
\]
which is induced by the inner product
\[
\ip{F}{G}_{W_{\mu, \bm{b}}^{1,2}(\cX; \cY)} 
:= \int_{\cX} \ip{F(X)}{G(X)}_{\cY} \D \mu(X) + \int_{\cX} \ip{D_{\cX_{\bm{b}}} F(X)}{D_{\cX_{\bm{b}}} G(X)}_{\HS(\cX_{\bm{b}};\cY)} \D \mu(X).
\]
As usual, we write \(W_{\mu, \bm{b}}^{1, 2}(\cX) := W_{\mu, \bm{b}}^{1, 2}(\cX; \bbR)\) for the space of Sobolev functionals.

\begin{remark}
    Defining Gaussian Sobolev spaces as the domain of the closure of a suitable differential operator is standard. See, e.g.,~\citet{chojnowska-michalik_GeneralizedOrnsteinUhlenbeckSemigroups_2001, daprato_IntroductionInfiniteDimensionalAnalysis_2006, daprato_IntroductionStochasticAnalysisMalliavin_2014, lunardi_InfiniteDimensionalAnalysis_2015}. For an equivalent definition via the completion of \(\cF C_b^1(\cX; \cY)\) under a suitable Sobolev norm, see~\citet[Ch. 5.2]{bogachev_GaussianMeasures_1998}.
\end{remark}

\begin{remark}
\label{rmk: weighted Sobolev spaces}
    Weighted Gaussian Sobolev spaces with \( \cY = \bbR \) have been considered in the literature in the study of continuous and compact Sobolev embeddings~\citep{shigekawa_SobolevSpacesWienerSpace_1992, daprato_RegularDensitiesInvariantMeasures_1995, chojnowska-michalik_GeneralizedOrnsteinUhlenbeckSemigroups_2001}.
    Recently,~\cite{luo_SobolevEmbeddingsInfiniteDimensions_2023} defined weighted Gaussian Sobolev spaces of functionals on \(\ell^r(\bbN)\), \(r\geq 1\), by weighting the partial derivatives with weights \( \bm{b} = (b_i)_{i = 1}^{\infty} \in \ell^{\infty}(\bbN) \). They remarked that their construction unifies various definitions of Gaussian Sobolev spaces via different choices of \(\bm{b}\). Our definition of \(W_{\mu, \bm{b}}^{1,2}(\cX)\) is equivalent to this construction in the case \(r = 2\). However, our approach via the differential operator along the space \(\cX_{\bm{b}}\) highlights the role of the latter as the underlying differential structure of \(W_{\mu, \bm{b}}^{1,2}(\cX)\).
    For \(\bm{b} = \sqrt{\bm{\lambda}}\), we obtain the same space as defined in~\citet{bogachev_GaussianMeasures_1998, lunardi_InfiniteDimensionalAnalysis_2015, daprato_IntroductionStochasticAnalysisMalliavin_2014}, which is commonly used in Malliavin calculus. For \(\bm{b} = \bm{1}\), we obtain a space as defined in~\citet{daprato_IntroductionInfiniteDimensionalAnalysis_2006}, which is strictly smaller. 
\end{remark}

\begin{remark}
    In the case of finite-dimensional domains \( \cX = \bbR^n \), \( n \in \bbN \), we have \( \cX_{\bm{b}} = \bbR^n \) for any weights \( \bm{b} = (b_i)_{i=1}^n >\bm{0} \). If additionally \( \cY = \bbR \), there is a strong relation between Gaussian and classical Sobolev spaces. In fact, 
    \[
    W_{\mu, \bm{b}}^{1,2}(\bbR^n)
    = \left\{ f \in W_{\textup{loc}}^{1,2}(\bbR^n) : f \in L^2_{\mu}(\bbR^n), \ \nabla f \in L^2_{\mu}(\bbR^n; \bbR^n) \right\},
    \]
    and the Gaussian and classical weak derivatives coincide~\citep[Prop. 1.5.2]{bogachev_GaussianMeasures_1998}. 
\end{remark}

Next, we use the Wiener-Hermite PC expansion~\eqref{eq: polynomial expansion of L2 operators} of \(L_{\mu}^2(\cX; \cY)\)-operators to characterize the space \(W_{\mu, \bm{b}}^{1,2}(\cX; \cY)\) by a weighted \(\ell^2\)-space.
More specifically, Gaussian Sobolev regularity of an operator translates to weighted summability of its Wiener-Hermite PC coefficients and vice versa. We start with some notation. 
Define the \emph{\(\bm{b}\)-weighted PCA eigenvalues} 
\begin{equation}
\label{eq: definition of lambda_b,i}
    \bm{\lambda_b} = (\lambda_{\bm{b}, i})_{i=1}^{\infty}, \quad \lambda_{\bm{b}, i} := \lambda_i / b_i^2,
\end{equation}
and the weights 
\begin{equation}
\label{eq: Sobolev weights}
    \bm{u}
    = (u_{\bm{\gamma}})_{\bm{\gamma} \in \Gamma}, \quad 
    u_{\bm{\gamma}} 
    = u_{\bm{\gamma}}(\bm{\lambda}_{\bm{b}})
    := \Bigg(1 + \sum_{i = 1}^{\infty} \frac{\gamma_i}{\lambda_{\bm{b},i}}\Bigg)^{-1/2}.
\end{equation}
We denote by \(\ell^2_{\bm{u}}(\Gamma; \cY) \) the space of all \(\cY\)-valued sequences \( \bm{Y} = (Y_{\bm{\gamma}})_{\bm{\gamma} \in \Gamma} \) whose norm 
\[
\nm{\bm{Y}}_{\ell^2_{\bm{u}}(\Gamma; \cY)} := \left( \sum_{\bm{\gamma} \in \Gamma}  u_{\bm{\gamma}}^{-2} \nm{Y_{\bm{\gamma}}}_{\cY}^2 \right)^{1/2}
\]
is finite.
The following result is an immediate consequence of Proposition~\ref{prop: l2-characterization of Gaussian Sobolev space}.

\begin{theorem}[{\(\ell^2\)-characterization of \(W_{\mu, \bm{b}}^{1,2}(\cX; \cY)\)}]
\label{thm: l2-characterization of Gaussian Sobolev space}
    The mapping 
    \[
    \ell_{\bm{u}}^2(\Gamma; \cY) \to W_{\mu, \bm{b}}^{1,2}(\cX; \cY), \quad \bm{Y} = (Y_{\bm{\gamma}})_{\bm{\gamma} \in \Gamma} \mapsto \sum_{\bm{\gamma} \in \Gamma} Y_{\bm{\gamma}} H_{\bm{\gamma}, \bm{\lambda}}
    \]
    is an isometric isomorphism. In particular, using the representation~\eqref{eq: polynomial expansion of L2 operators}, we have
    \[
    \nm{F}_{W_{\mu, \bm{b}}^{1,2}(\cX; \cY)}^2 = \sum_{\bm{\gamma }\in \Gamma} u_{\bm{\gamma}}^{-2} \nm{\int_{\cX} F H_{\bm{\gamma}, \bm{\lambda}} \D \mu}_{\cY}^2,
    \quad \forall F \in W_{\mu, \bm{b}}^{1, 2}(\cX; \cY).
    \]
\end{theorem}
Note that by setting \( \cX = \bbR^n \), \( \cY = \bbR \), and \( \lambda_1 = \dots = \lambda_n = 1 \), we immediately recover the classical result for the isotropic Gaussian Sobolev space~\citep{lopez_SobolevSpacesPotentialSpaces_2018}, namely,
\[
\nm{F}_{W^{1,2}_{\mu, \bm{1}}(\bbR^n)}^2 = \sum_{\bm{\gamma} \in \bbN_0^n} \left( 1 + \sum_{i = 1}^n \gamma_i \right) \abs{\int_{\bbR^n} F H_{\bm{\gamma}, \bm{1}} \D \mu}^2.
\]

\begin{assumption}[Properties of \(\bm{b}\)]
\label{ass: properties of b}
    The sequence \(\bm{b} = (b_i)_{i=1}^{\infty} \in \ell^{\infty}(\bbN) \), \(\bm{b} > \bm{0} \), is a sequence of positive bounded real numbers such that the sequence of weighted PCA eigenvalues \(\bm{\lambda_b} = (\lambda_{\bm{b}, i})_{i=1}^{\infty}\), defined in~\eqref{eq: definition of lambda_b,i}, is nonincreasing and satisfies \(\lim_{i \to \infty} \lambda_{\bm{b},i} = 0\). If \(\dim(\cY) = \infty\), we assume in addition that \(\bm{b} \in \ell^2(\bbN)\).
\end{assumption}

The weights \(u_{\bm{\gamma}}\) in~\eqref{eq: Sobolev weights} are key to our analysis.
By Assumption~\ref{ass: properties of b}, every superlevel set \( \{ \bm{\gamma} \in \Gamma : u_{\bm{\gamma}} \geq t \} \), \( t > 0 \), is finite. We can thus order the weights \(u_{\bm{\gamma}}\) in a nonincreasing way, that is, there exists a bijection $\pi: \bbN \to \Gamma$ that gives a \emph{nonincreasing rearrangement} of \(\bm{u}\) so that
\begin{equation}
\label{eq: nonincreasing rearrangement pi}
    u_{\bm{\pi(1)}} \geq u_{\bm{\pi(2)}} \geq \cdots > 0.
\end{equation}
The mapping \(\pi\) is unique up to permutations of weights of the same value. 
The additional requirement in Assumption~\ref{ass: properties of b} that \( \bm{b} \in \ell^2(\bbN) \) in the case \(\dim(\cY) = \infty\) is necessary in our analysis to show that the set of Lipschitz operators is a subset of $W_{\mu, \bm{b}}^{1, 2}(\cX; \cY)$, see Theorem~\ref{thm: Lipschitz operators are Gaussian Sobolev operators}.

\begin{example}
\label{ex: choices for b}
    If \(\cY\) is finite-dimensional and the \( \lambda_i \) are assumed to be nonincreasing, we can choose \(\bm{b} = \bm{1}\), so that \(\bm{\lambda_b} = \bm{\lambda}\). This is not possible if \(\cY\) is infinite-dimensional, as \(\bm{1} \not\in \ell^2(\bbN) \). Note that \( \bm{b} = \sqrt{\bm{\lambda}} \) is not a valid choice either, as it yields \( \bm{\lambda_b} = \bm{1} \) and thus violates the limit condition in Assumption~\ref{ass: properties of b}. However, since \(\bm{\lambda} \in \ell^1(\bbN)\), a valid choice in any case is given, for example, by \(b_i := \lambda_i^{1/2} / (\sum_{j = i}^{\infty} \lambda_j)^{1/4} \), \( i \in \bbN \). In this case, \(\bm{b} > \bm{0} \) and \( \lambda_{\bm{b}, i} = (\sum_{j = i}^{\infty} \lambda_j)^{1/2} \) so that \( \lim_{i \to \infty} \lambda_{\bm{b}, i} = 0 \).  It is a short exercise to show that \( \bm{b} \in \ell^2(\bbN) \). For this, set \( s_i := \sum_{j = i}^{\infty} \lambda_j \) and compute
    \[
    b_i^2 = \frac{s_i - s_{i + 1}}{\sqrt{s_i}} = \frac{(\sqrt{s_i} + \sqrt{s_{i+1}})(\sqrt{s_i} - \sqrt{s_{i+1}})}{\sqrt{s_i}}
    \leq \frac{2 \sqrt{s_i} (\sqrt{s_i} - \sqrt{s_{i+1}})}{\sqrt{s_i}}
    = 2(\sqrt{s_i} - \sqrt{s_{i+1}}).
    \]
    Consequently, 
    \(
    \sum_{i = 1}^{\infty} b_i^2 \leq 2 \sqrt{s_1} < \infty.
    \)
\end{example}

\begin{remark}
\label{rmk: PCA conditions and embeddings}
    The condition \(\lim_{i \to \infty} \lambda_{\bm{b}, i} = 0\) is equivalent to the compact embedding of \(W_{\mu, \bm{b}}^{1,2}(\cX)\) in \(L_{\mu}^2(\cX)\)~\citep[Prop. 2.2]{daprato_RegularDensitiesInvariantMeasures_1995}, and, more generally, in the Orlicz space \(L^2 \log^q L(\cX, \mu)\) for \(q \in [0, 1)\)~\citep[Thm. 5.2]{luo_SobolevEmbeddingsInfiniteDimensions_2023}.
\end{remark}

\subsection{Lipschitz operators}
\label{subsec: Lipschitz operators}

We now turn to Lipschitz continuous operators and first recall their definition. 
\begin{definition}[Lipschitz operators]
    An operator \(F: \cX \to \cY\) is called \emph{(\(L\)-)Lipschitz (continuous)} if there exists a constant \(L > 0\) such that
    \[
    \nm{F(X) - F(Z)}_{\cY} \leq L \nm{X - Z}_{\cX}, \quad \forall X, Z \in \cX.
    \]
    The number \(L\) is a \emph{Lipschitz constant} of \(F\). The smallest Lipschitz constant of \(F\) is given by 
    \[
    [F]_{\Lip(\cX; \cY)} := \sup_{\substack{X, Z \in \cX \\ X \neq Z}} \frac{\nm{F(X) - F(Z)}_{\cY}}{\nm{X - Z}_{\cX}}.
    \]
    We denote the space of all Lipschitz operators from \(\cX\) to \(\cY\) by \(\Lip(\cX; \cY)\) and write \(\Lip(\cX) := \Lip(\cX; \bbR)\). 
    We further define the space of all \emph{bounded} Lipschitz operators \(C^{0,1}(\cX;\cY)\) as the set of all Lipschitz operators \(F \in \Lip(\cX; \cY)\) with finite norm 
    \[
    \nm{F}_{C^{0, 1}(\cX; \cY)} 
    := \sup_{X \in \cX} \nm{F(X)}_{\cY} + [F]_{\Lip(\cX; \cY)}.
    \]
    Notice that \(C^{0, 1}(\cX; \cY)\) is a strict subset of \(\Lip(\cX; \cY)\) as operators in \(\Lip(\cX; \cY)\) are not necessarily bounded.
\end{definition}

The next result is the main result of this section. It shows that Lipschitz operators are Gaussian Sobolev operators. This motivates Gaussian Sobolev spaces as a natural setting for the study of Lipschitz operators. Recall that on finite-dimensional domains, the classical Rademacher theorem states that Lipschitz functions \( \bbR^n \to \bbR \) are differentiable \(\cL^n\)-almost everywhere (here and throughout, \( \cL^n \) denotes the Lebesgue measure). This can be generalized to Banach-valued operators: Any Lipschitz operator \( F : \bbR^n \to \cY \) is \(\cL^n\)-almost everywhere Fréchet differentiable if \( \cY \) is a Banach space which has the Radon-Nikodym property (which is satisfied by any Hilbert space). See~\citet[Prop. 6.41]{benyamini_GeometricNonlinearFunctionalAnalysis_1999}. Theorem~\ref{thm: Lipschitz operators are Gaussian Sobolev operators} can be interpreted as a further generalization of Rademacher's theorem to operators on infinite-dimensional domains with the Lebesgue measure replaced by a Gaussian measure. In fact, the proof is based on a reduction of the infinite-dimensional setting to the finite-dimensional one by means of a suitable approximating sequence and a subsequent application of the classical Rademacher theorem. We present a sketch of the proof below and provide full details in Appendix~\ref{sec: app: Lipschitz operators are Gaussian Sobolev operators}.

\begin{theorem}[Lipschitz operators are Gaussian Sobolev operators]
\label{thm: Lipschitz operators are Gaussian Sobolev operators}
    Let \(\bm{b} = (b_i)_{i=1}^{\infty} \in \ell^{\infty}(\bbN)\), \(\bm{b} > \bm{0}\), be a sequence of positive bounded numbers.
    \begin{itemize}
        \item [(i)] If \(\cY\) is finite-dimensional, then \(\Lip(\cX; \cY) \subset W_{\mu, \bm{b}}^{1,2}(\cX; \cY)\) and 
        \[
        \nm{D_{\cX_{\bm{b}}} F}_{L_{\mu}^2(\cX; \HS(\cX_{\bm{b}}; \cY))} 
        \leq \sqrt{\dim(\cY)} \cdot \nm{\bm{b}}_{\ell^{\infty}(\bbN)} \cdot [F]_{\Lip}, 
        \quad \forall F \in \Lip(\cX; \cY).
        \]
        In particular, the embedding \(C^{0, 1}(\cX; \cY) \hookrightarrow W_{\mu, \bm{b}}^{1,2}(\cX; \cY)\) is continuous with 
        \[
        \nm{F}_{W_{\mu, \bm{b}}^{1,2}(\cX; \cY)} \leq \max\left\{1, \sqrt{\dim(\cY)} \cdot \nm{\bm{b}}_{\ell^{\infty}(\bbN)}\right\} \cdot \nm{F}_{C^{0, 1}(\cX; \cY)}, \quad \forall F \in C^{0, 1}(\cX; \cY).
        \]

        \item [(ii)] If \(\cY\) is infinite-dimensional and if \(\bm{b} \in \ell^2(\bbN)\), then \(\Lip(\cX; \cY) \subset W_{\mu, \bm{b}}^{1,2}(\cX; \cY)\) and 
        \[
        \nm{D_{\cX_{\bm{b}}} F}_{L_{\mu}^2(\cX; \HS(\cX_{\bm{b}}; \cY))} 
        \leq \nm{\bm{b}}_{\ell^2(\bbN)} \cdot [F]_{\Lip}, 
        \quad \forall F \in \Lip(\cX; \cY).
        \]
        In particular, the embedding \(C^{0, 1}(\cX; \cY) \hookrightarrow W_{\mu, \bm{b}}^{1,2}(\cX; \cY)\) is continuous with 
        \[
        \nm{F}_{W_{\mu, \bm{b}}^{1,2}(\cX; \cY)} \leq \max \left\{1, \nm{\bm{b}}_{\ell^2(\bbN)} \right\} \cdot \nm{F}_{C^{0, 1}(\cX; \cY)}, \quad \forall F \in C^{0, 1}(\cX; \cY).
        \]
    \end{itemize}
\end{theorem}

\begin{proof}[Proof (Sketch)]
    Let \(F \in \Lip(\cX; \cY)\). In order to show that \(F\) lies in \(W_{\mu, \bm{b}}^{1,2}(\cX; \cY)\), it suffices to find a sequence \((F_n)_{n = 1}^{\infty}\) of operators which converges to \(F\) in \(L_{\mu}^2(\cX; \cY)\) with the \(F_n\) uniformly bounded in \(W_{\mu, \bm{b}}^{1,2}(\cX; \cY)\) (see Lemma~\ref{lem: belonging to W^1,2}). To this end, we construct a specific sequence of operators of the form \(F_n = V_n \circ T_n\) with \(V_n: \bbR^n \to \cY\) and \(T_n: \cX \to \bbR^n\) which converges to \(F\) in \(L_{\mu}^2(\cX; \cY)\) and, in fact, in \(W_{\mu, \bm{b}}^{1, 2}(\cX; \cY)\). It can be shown that the operator \(V_n\) inherits Lipschitz continuity of \(F\). Hence, by Rademacher's theorem, \(V_n\) is differentiable \(\cL^n\)-almost everywhere. It remains to show that \(F_n\) is differentiable \(\mu\)-almost everywhere and that there exists a constant \(C > 0\) such that
    \begin{equation}
    \label{eq: Lipschitz: proof sketch: boundedness of HS norm}
        \nm{D_{\cX_{\bm{b}}} F_n(X)}_{\HS(\cX_{\bm{b}};\cY)} \leq C
    \end{equation}
    for all \(n \in \bbN\) and \(\mu\)-almost every \(X \in \cX\). 

    In case (i), let \(m := \dim({\cY})\). We fix an orthonormal basis \(\{\psi_j\}_{j = 1}^m\) of \(\cY\) and consider the (Lipschitz continuous) coordinate functions \(V_n^{(j)} := \ip{V_n}{\psi_j}_{\cY}: \bbR^n \to \bbR\). Unwinding definitions, a straightforward calculation then leads to~\eqref{eq: Lipschitz: proof sketch: boundedness of HS norm} for any bounded weight sequence \(\bm{b} > \bm{0}\). The resulting constant \(C\) is given by \(\sqrt{\dim(\cY)} \cdot \nm{\bm{b}}_{\ell^{\infty}(\bbN)} \cdot [F]_{\Lip}\).     
    The same argument does not work in case (ii) where \(\cY\) has infinite dimension. At this point, we additionally require that \(\bm{b} \in \ell^2(\bbN)\). A slight modification of the argument in (i) then yields~\eqref{eq: Lipschitz: proof sketch: boundedness of HS norm} with \(C\) given by \(\nm{\bm{b}}_{\ell^2(\bbN)} \cdot [F]_{\Lip}\).
    The continuous embedding of \(C^{0, 1}(\cX; \cY)\) in \(W_{\mu, \bm{b}}^{1, 2}(\cX; \cY)\) follows in both cases from~\eqref{eq: Lipschitz: proof sketch: boundedness of HS norm} and the fact that \(F_n \to F\) in \(W_{\mu, \bm{b}}^{1, 2}(\cX; \cY)\).
\end{proof}

In light of Theorem~\ref{thm: Lipschitz operators are Gaussian Sobolev operators}, note that Assumption~\ref{ass: properties of b} implies that \(\Lip(\cX; \cY) \subset W_{\mu, \bm{b}}^{1, 2}(\cX; \cY)\), regardless of whether \(\cY\) is finite- or infinite-dimensional. 

\begin{remark}
    For functionals (i.e., \(\cY = \bbR\)), it is well-known that Lipschitz continuity implies Gaussian Sobolev regularity. We refer to~\citet[Prop. 10.11]{daprato_IntroductionInfiniteDimensionalAnalysis_2006} and~\citet[Prop. 3.18]{daprato_IntroductionStochasticAnalysisMalliavin_2014} where it is shown that \(\Lip(\cX) \subset W_{\mu, \bm{1}}^{1,2}(\cX)\) and \(\Lip(\cX) \subset W_{\mu, \sqrt{\bm{\lambda}}}^{1,2}(\cX)\), respectively.
\end{remark}

\begin{remark}
    One can define Gaussian Sobolev spaces \(W_{\mu, \bm{b}}^{1,p}(\cX; \cY)\) for any order \(1 \leq p < \infty\) and a proof analogous to the one of Theorem~\ref{thm: Lipschitz operators are Gaussian Sobolev operators} shows that they contain \(\Lip(\cX; \cY)\) as a subset. For the case \(\cY = \bbR\), we mention~\citet[Ex. 5.4.10(i)]{bogachev_GaussianMeasures_1998} and~\citet[Prop. 10.1.4]{lunardi_InfiniteDimensionalAnalysis_2015}. However, only in the Hilbert space case \(p = 2\) is there a simple characterization of \(W_{\mu, \bm{b}}^{1,p}(\cX; \cY)\) in terms of a weighted \(\ell^2\)-space as given by Theorem~\ref{thm: l2-characterization of Gaussian Sobolev space}.
\end{remark}

%%%%%%%%%%%%%%%%%%%%%%%%%%%%%%%%%%%%%%%%%%%%%%%%%%%%%%%%%%%%%%%%%%%%%%%%%%%%%%%%%%%%%%%%%%
%%%%%%%%%%%%%%%%%%%%%%%%%%%%%%%%%%%%%%%%%%%%%%%%%%%%%%%%%%%%%%%%%%%%%%%%%%%%%%%%%%%%%%%%%%

\section{Polynomial \texorpdfstring{\(s\)}{s}-term approximation}
\label{sec: polynomial s-term approximation}

The \(\ell^2\)-characterization of \(W_{\mu, \bm{b}}^{1, 2}(\cX; \cY)\) via Wiener-Hermite PC expansions (see Theorem~\ref{thm: l2-characterization of Gaussian Sobolev space}) motivates studying polynomial \(s\)-term approximations of \(W_{\mu, \bm{b}}^{1, 2}\)-operators and quantifying the smallest achievable worst-case \(s\)-term error.
To this end, for any index set \(S \subset \Gamma\), we define the space of \(\cY\)-valued polynomials
\[
\cP_{S; \cY} := \left \{ \sum_{\bm{\gamma} \in S} Y_{\bm{\gamma}} H_{\bm{\gamma}, \bm{\lambda}} : Y_{\bm{\gamma}} \in \cY \right \}
\]
and the corresponding orthogonal \(L_{\mu}^2\)-projection
\begin{equation}
\label{eq: L2 projection}
    (\cdot)_S: L_{\mu}^2(\cX; \cY) \to \cP_{S; \cY}, \quad F \mapsto F_S := \sum_{\bm{\gamma} \in S} \left( \int_{\cX} F H_{\bm{\gamma}, \bm{\lambda}} \D \mu \right) H_{\bm{\gamma}, \bm{\lambda}}.
\end{equation}
We further define the \emph{Sobolev unit ball} and the \emph{Sobolev unit Lipschitz ball} as, respectively,
\begin{align}
\begin{split}
\label{eq: Sobolev unit (Lipschitz) ball}
    B_{\mu, \bm{b}}(\cX; \cY)
    &:= \left\{ F \in W_{\mu, \bm{b}}^{1,2}(\cX; \cY): \nm{F}_{W_{\mu, \bm{b}}^{1,2}(\cX; \cY)} \leq 1 \right\},
    \\
    B_{\mu, \bm{b}}^{\Lip}(\cX; \cY)
    &:= \left\{ F \in \Lip(\cX; \cY): \nm{F}_{W_{\mu, \bm{b}}^{1,2}(\cX; \cY)} \leq 1 \right\}.
\end{split}
\end{align}
As usual, when \( \cY = \bbR \), we drop it from the notation.

Next, let \(F \in W_{\mu, \bm{b}}^{1, 2}(\cX; \cY)\) and let \(S \subset \Gamma\) be finite with \(\abs{S} \leq s\). Setting \(Y_{\bm{\gamma}} := \int_{\cX} F H_{\bm{\gamma}, \bm{\lambda}} \D \mu\), it follows from Parseval's identity~\eqref{eq: Parseval's identity} and Theorem~\ref{thm: l2-characterization of Gaussian Sobolev space} that
\begin{align*}
    \nm{F - F_S}_{L_{\mu}^2(\cX; \cY)}^2 
    &= \sum_{\bm{\gamma} \in \Gamma \setminus S} \nm{Y_{\bm{\gamma}}}_{\cY}^2 
    \leq \big( \max_{\bm{\gamma} \in \Gamma \setminus S} u_{\bm{\gamma}}^2 \big) \sum_{\bm{\gamma} \in \Gamma} u_{\bm{\gamma}}^{-2} \nm{Y_{\bm{\gamma}}}_{\cY}^2  
    = \big( \max_{\bm{\gamma} \in \Gamma \setminus S} u_{\bm{\gamma}}^2 \big) \nm{F}_{W_{\mu, \bm{b}}^{1, 2}(\cX; \cY)}^2.
\end{align*}
Let us recall from~\eqref{eq: nonincreasing rearrangement pi} the mapping \(\pi: \bbN \to \Gamma\) which gives a nonincreasing rearrangement of \(\bm{u} = (u_{\bm{\gamma}})_{\bm{\gamma} \in \Gamma}\). For $s \in \bbN$, we set \(S = \pi([s]) = \{\bm{\pi(1)}, \dots, \bm{\pi(s)}\}\) and conclude for \(\cK \in \{ B_{\mu, \bm{b}}(\cX; \cY), B_{\mu, \bm{b}}^{\Lip}(\cX; \cY) \}\) that
\begin{align}
\label{eq: s-term: approximation error <= u_pi(s+1)}
\begin{split}
    \inf_{S \subset \Gamma, \abs{S} \leq s} \sup_{F \in \cK} \nm{F - F_S}_{L_{\mu}^2(\cX; \cY)}
    &\leq \sup_{F \in \cK} \nmd{F - F_{\pi([s])}}_{L_{\mu}^2(\cX; \cY)} 
    \leq \max_{\bm{\gamma} \in \Gamma \setminus \pi([s])} u_{\bm{\gamma}} 
    = u_{\bm{\pi(s + 1)}}.
\end{split}
\end{align}
Our first main result in this section shows that this chain of inequalities can, in fact, be improved to equality and hence gives a tight characterization of the best polynomial \(s\)-term error.
The proof is an immediate consequence of Theorem~\ref{thm: adaptive m-width} in the special case \(\cV = L_{\mu}^2(\cX; \cY)\).

\begin{theorem}[Best polynomial \(s\)-term error]
\label{thm: s-term: best s-term approximation error}
    For \(\cK \in \{ B_{\mu, \bm{b}}(\cX; \cY), B_{\mu, \bm{b}}^{\Lip}(\cX; \cY) \}\) and every \(s \in \bbN\), we have
    \[
    \inf_{S \subset \Gamma, \abs{S} \leq s} \sup_{F \in \cK} 
    \nm{F - F_S}_{L_{\mu}^2(\cX; \cY)} 
    = \sup_{F \in \cK} \nmd{F - F_{\pi([s])}}_{L_{\mu}^2(\cX; \cY)} 
    = u_{\bm{\pi(s + 1)}}.
    \]
\end{theorem}

Motivated by Theorem~\ref{thm: s-term: best s-term approximation error}, we study in the rest of this section the decay of \(u_{\bm{\pi(s + 1)}}\) as \(s \to \infty\).
The proofs are based on the relation of the set \(\pi([s])\) to an anisotropic total degree index set, which we discuss in Section~\ref{subsec: s-term: relation to anisotropic total degree index sets}. Subsequently, we prove lower and upper bounds for \(u_{\bm{\pi(s + 1)}}\) in Sections~\ref{subsec: s-term: lower bound} and~\ref{subsec: s-term: upper bounds}.

\subsection{Relation to anisotropic total degree index sets}
\label{subsec: s-term: relation to anisotropic total degree index sets}

We first recall the notion of anisotropic total degree (TD) index sets and provide lower and upper size bounds. We then identify a specific such set to which we can relate \(\pi([s])\).

\begin{definition}[Anisotropic TD index set]
    For \(d\in\bbN\) and \(\bm{a}=(a_1,\dots,a_d)\in\bbR^d\), \(\bm{a} > \bm{0}\), we define the anisotropic TD index set in \(d\) dimensions with weight \(\bm{a}\) by
    \[
    \Lambda_{d,\bm{a}}^{\textup{TD}} := \left\{\nu\in\bbN_0^d : \sum_{i=1}^d a_i \nu_i \leq 1\right\}.
    \]
\end{definition}

\begin{lemma}[Lower and upper size bounds for anisotropic TD index sets]
\label{lem: size bounds}
    Let \(d \in \bbN\) and \(\bm{a}=(a_1,\dots,a_d)\in\bbR^d\) with \(0 < a_1\leq\dots\leq a_d\). We have
    \begin{equation}
    \label{eq: anisotropic TD index set}
               \prod_{i=1}^d \frac{1}{a_i i} \leq \absd{\Lambda_{d,\bm{a}}^{\textup{TD}}} \leq \prod_{i=1}^d \left(\frac{1}{a_i i} + 1\right).
    \end{equation}
\end{lemma}

\begin{proof}
    The upper bound is Lemma 5.3 in~\citet{haji-ali_NovelResultsAnisotropicSparse_2018}. The lower bound is proved in~\citet{beged-dov_LowerUpperBoundsNumber_1972}. See~\citet{griebel_TensorProductApproximationAnalytic_2016} for further discussion.
\end{proof}

For any \(\varepsilon > 0\), we define the set
\begin{equation}
\label{eq: s-term: definition of S_varepsilon}
    S(\varepsilon) := \left\{\bm{\gamma} \in \Gamma: u_{\bm{\gamma}}^{-2} \leq 1 + \frac{1}{\varepsilon^2} \right\}
    = \left\{\bm{\gamma} \in \Gamma: \sum_{i = 1}^{\infty} \frac{\gamma_i}{\lambda_{\bm{b}, i}} \leq \frac{1}{\varepsilon^2} \right\}
\end{equation}
as well as the quantity 
\begin{equation}
\label{eq: s-term: effective dimension d}
    d(\varepsilon)
    := \min \left\{l \in \bbN_0: \lambda_{\bm{b}, l+1} < \varepsilon^2 \right\} \in \bbN_0.
\end{equation}
By definition, we have \(S(\varepsilon) = \pi([\absd{S(\varepsilon)}])\). If \( d(\varepsilon) > 0 \), then \(S(\varepsilon)\) is isomorphic to an anisotropic TD index set via the mapping
\begin{equation}
\label{eq: s-term: S_varepsilon isomorphic to anisotropic TD index set}
    S(\varepsilon) \overset{\simeq}{\longrightarrow} \Lambda_{d(\varepsilon), \bm{a}}^{\textup{TD}}, \quad \bm{\gamma} \mapsto (\gamma_1, \dots, \gamma_{d(\varepsilon)}),
\end{equation}
with weights \(a_i := \varepsilon^2/\lambda_{\bm{b}, i}\), \(i \in [d(\varepsilon)]\).
Observe that by the monotonicity condition in Assumption~\ref{ass: properties of b} we have \(0 < a_1 \leq a_2 \leq \cdots \leq a_{d(\varepsilon)}\).

\begin{remark}[Effective dimension]
    The number \(d(\varepsilon)\), defined in~\eqref{eq: s-term: effective dimension d}, can be interpreted as the \emph{effective dimension} of the approximation problem in the following sense: For \(i > d(\varepsilon)\), we have \( \lambda_i \leq \lambda_{\bm{b}, i} \nm{\bm{b}}_{\ell^{\infty}(\bbN)}^2 < \varepsilon^2 \nm{\bm{b}}_{\ell^{\infty}(\bbN)}^2 \). That is, the variance of \(\mu\) in the \(i\)th coordinate direction (with respect to the PCA basis \(\{\phi_i\}_{i=1}^{\infty}\)) essentially vanishes for very small values of \(\varepsilon\). Consequently, \(\cN(0, \lambda_i) \approx \delta_0\) for \(i > d(\varepsilon)\), where \(\delta_0\) is the Dirac delta measure centered at \(0\). Measuring an operator \(F\) on \(\cX\) with respect to \(\mu\) thus essentially reduces to measuring \(F\) in its first \(d(\varepsilon)\) coordinates with respect to the Gaussian product measure \(\bigotimes_{i = 1}^{d(\varepsilon)} \cN(0, \lambda_i)\).
\end{remark}

\subsection{Lower bound}
\label{subsec: s-term: lower bound}

We now prove the second main result of this section which is a lower bound for \(u_{\bm{\pi(s + 1)}}\). It states that, regardless of the (unweighted) PCA eigenvalues \(\bm{\lambda}\) and choice of \(\bm{b}\), one cannot achieve an algebraic decay of \(u_{\bm{\pi(s + 1)}}\) as \(s \to \infty\). The proof is based on relation~\eqref{eq: s-term: S_varepsilon isomorphic to anisotropic TD index set} and the lower size bound in Lemma~\ref{lem: size bounds}.

\begin{theorem}[Impossibility of algebraic decay of \(u_{\bm{\pi(s + 1)}}\)]
\label{thm: s-term: lower bound for u_pi(s+1)}
    For any \(p \in \bbN\), there exists \(\bar{s} \in \bbN\), depending on \(\lambda_{\bm{b}, 1}, \dots, \lambda_{\bm{b}, p}\), and \(p\), such that
    \[
    u_{\bm{\pi(s+1)}} \geq C s^{-\frac{1}{2p}}, \quad \forall s \geq \bar{s},
    \]
    with constant 
    \(
    C = C(\lambda_{\bm{b},1}, \dots, \lambda_{\bm{b},p}, p) 
    := \frac{1}{2} \left(\prod_{i=1}^p \frac{\lambda_{\bm{b},i}}{i}\right)^{\frac{1}{2p}}.
    \)
\end{theorem}

\begin{proof}  
    Fix \(\varepsilon > 0\), whose exact value will be chosen later, and define for \(n \in \bbN\) the set
    \[
    S(\varepsilon, n) := \left\{\bm{\gamma} \in \Gamma: \sum_{i = 1}^{\infty} \frac{\gamma_i}{\lambda_{\bm{b},i}} \leq \frac{1}{\varepsilon^2}, \ \supp(\bm{\gamma}) \subset [n] \right\}.
    \]
    We first make a couple of simple but important observations. First note that \(S(\varepsilon) = S(\varepsilon, d(\varepsilon))\), where \(S(\varepsilon)\) and \(d(\varepsilon)\) are defined in~\eqref{eq: s-term: definition of S_varepsilon} and~\eqref{eq: s-term: effective dimension d}, respectively. Second, we have \(S(\varepsilon, n') \subset S(\varepsilon, n)\) for every \(1 \leq n' \leq n\). Third, \(S(\varepsilon, n)\) is isomorphic to the anisotropic TD index set \(\Lambda_{n, \bm{a}}^{\textup{TD}}\) with weight \(\bm{a} = (a_1, \dots, a_n)\), \(a_i := \varepsilon^2/\lambda_{\bm{b}, i}\), via the mapping \( \bm{\gamma} \mapsto (\gamma_1, \dots, \gamma_n) \). 
    We now combine these observations with the lower size bound in~\eqref{eq: anisotropic TD index set} to conclude
    \begin{equation}
    \label{eq: s-term: lower bound on size of S_varepsilon}
        \abs{S(\varepsilon)} 
        = \abs{S(\varepsilon, d(\varepsilon))} 
        \geq \abs{S(\varepsilon, d')} 
        \geq \prod_{i=1}^{d'} \frac{\lambda_{\bm{b},i}}{\varepsilon^2 i}, 
        \quad \forall 1 \leq d' \leq d(\varepsilon).
    \end{equation}
    Analogously to the definition of \(d(\varepsilon)\) in~\eqref{eq: s-term: effective dimension d}, we set
    \[
    \wtd(\varepsilon) 
    := \min \left\{ l \in \bbN_0: \frac{\lambda_{\bm{b}, l+1}}{l+1} < \varepsilon^2 \right\} \in \bbN_0.
    \]
    Note that we have \(\wtd(\varepsilon) \leq d(\varepsilon)\) and \(\wtd(\varepsilon) \to \infty\) as \(\varepsilon \to 0\).
    
    Next, let \(p \in \bbN\) be arbitrary. We fix some \(0 < \bar{\varepsilon} = \bar{\varepsilon}(\lambda_{\bm{b}, p}, p) \leq \min\{ \sqrt{\lambda_{\bm{b}, p}/p}, 1 \}\). By definition of \(\wtd(\varepsilon)\), we have \(\wtd(\varepsilon) \geq p\) for every \(0 < \varepsilon \leq \bar{\varepsilon}\). Since \(\lambda_{\bm{b}, i}/(\varepsilon^2 i) \geq 1\) for every \(1 \leq i \leq \wtd(\varepsilon)\), it follows from~\eqref{eq: s-term: lower bound on size of S_varepsilon} that
    \[
    \abs{S(\varepsilon)} 
    \geq \prod_{i = 1}^{\wtd(\varepsilon)} \frac{\lambda_{\bm{b}, i}}{\varepsilon^2 i} 
    \geq \prod_{i = 1}^{p} \frac{\lambda_{\bm{b}, i}}{\varepsilon^2 i} 
    = \wtC \varepsilon^{-2p},
    \quad \forall 0 < \varepsilon \leq \bar{\varepsilon},
    \]
    with constant 
    \(
    \wtC 
    = \wtC(\lambda_{\bm{b}, 1}, \dots, \lambda_{\bm{b}, p}, p) 
    := \prod_{i = 1}^p \frac{\lambda_{\bm{b}, i}}{i}.
    \)
    We choose \(\bar{s} = \bar{s}(\lambda_{\bm{b}, 1}, \dots, \lambda_{\bm{b}, p}, p) \in \bbN\) sufficiently large such that \(s + 1 \geq \lceil \wtC \bar{\varepsilon}^{-2 p} \rceil\) for every \(s \geq \bar{s}\). We then fix some arbitrary \(s \geq \bar{s}\) and pick \(0 < \widetilde{\varepsilon}\leq \bar{\varepsilon}\) such that 
    \(
    \wtC \widetilde{\varepsilon}^{-2p} = s + 1.
    \)
    Solving for \(\widetilde{\varepsilon}^2\) yields
    \[
    \widetilde{\varepsilon}^2 
    = \widetilde{\varepsilon}^2 (\lambda_{\bm{b}, 1}, \dots, \lambda_{\bm{b}, p}, p, s) 
    = \wtC^{1/p} (s + 1)^{-1/p}.
    \]
    By our choice of \(\widetilde{\varepsilon}\), we have \(\absd{S(\widetilde{\varepsilon})} \geq s + 1\). Since \(S(\widetilde{\varepsilon}) = \pi([\absd{S(\widetilde{\varepsilon})}])\), we conclude \(\bm{\pi(s + 1)} \in S(\widetilde{\varepsilon})\) and therefore
    \[
    u_{\bm{\pi(s+1)}}^2 
    \geq \left(\frac{1}{\widetilde{\varepsilon}^2} + 1\right)^{-1} 
    \geq \frac{1}{2}\widetilde{\varepsilon}^2 
    = \frac{1}{2} \wtC^{1/p} (s+1)^{-1/p}
    \geq \frac{1}{4} \wtC^{1/p} s^{-1/p}, 
    \quad \forall s \geq \bar{s},
    \]
    where the second inequality holds because \(\widetilde{\varepsilon} \leq \bar{\varepsilon} \leq 1\). This completes the proof.
\end{proof}

\subsection{Upper bounds}
\label{subsec: s-term: upper bounds}

We have now seen that \(u_{\bm{\pi(s + 1)}}\) cannot decay algebraically as \(s \to \infty\), regardless of the decay of the \(\bm{b}\)-weighted PCA eigenvalues \(\lambda_{\bm{b},i}\). We now study the decay of \(u_{\bm{\pi(s + 1)}}\) for three different decays of these eigenvalues: Algebraic and exponential decays, which are typically considered in the context of (functional) PCA (see, e.g.,~\citet{reiss_NonasymptoticUpperBoundsReconstruction_2020, milbradt_HighprobabilityBoundsReconstructionError_2020} and references therein), as well as double exponential decays.
In principle, the proof of the following result can be adapted to any other decay of the \(\lambda_{\bm{b},i}\) as long as \( (\lambda_{\bm{b}, i})_{i = 1}^{\infty} \) is nonincreasing and \( \lim_{i \to \infty} \lambda_{\bm{b}, i} = 0 \). It is based on relation~\eqref{eq: s-term: S_varepsilon isomorphic to anisotropic TD index set} and the upper size bound in Lemma~\ref{lem: size bounds}.

\begin{theorem}[Typical decays of \(u_{\bm{\pi(s + 1)}}\)]
\label{thm: s-term: upper bounds for u_pi(s+1)}
Let \(\alpha, \beta > 0\).
\begin{itemize}  
    \item [(a)] \emph{Algebraic spectral decay:} Let \(\lambda_{\bm{b}, i} = i^{-\alpha}\) for every \(i \in \bbN\). Then, for every \(\delta, \eta > 0\), there exists \(\bar{s} = \bar{s}(\alpha, \delta, \eta) \in \bbN\) such that for every \(s \geq \bar{s}\),
    \begin{equation}
    \label{eq: s-term: upper bound: algebraic}
        u_{\bm{\pi(s + 1)}} \leq \eta \log(s)^{-\frac{1}{2(1/\alpha + \delta)}}.
    \end{equation}

    \item [(b)] \emph{Exponential spectral decay:} Let \(\lambda_{\bm{b}, i} = e^{-\alpha i^{\beta}}\) for every \(i \in \bbN\). Then, for every \(\delta > 0\), there exists \(\bar{s} = \bar{s}(\alpha, \beta, \delta) \in \bbN\) such that for every \(s \geq \bar{s}\),
    \begin{equation}
    \label{eq: s-term: upper bound: exponential}
        u_{\bm{\pi(s + 1)}} \leq e^{- \frac{1}{2} \alpha^{\frac{1}{\beta + 1}} \left( \frac{\beta + 1}{\beta + \delta}  \right)^{\frac{1}{1 + 1/\beta}} \log(s)^{\frac{1}{1 + 1/\beta}}}.
    \end{equation}
    
    \item [(c)] \emph{Double exponential spectral decay:} Let \(\lambda_{\bm{b}, i} = e^{-e^{\alpha i}}\) for every \(i \in \bbN\). Then, for every \(\delta, \eta > 0\), there exists \(\bar{s} = \bar{s}(\alpha, \delta, \eta) \in \bbN\) such that for every \(s \geq \bar{s}\),
    \begin{equation}
    \label{eq: s-term: upper bound: double exponential}
        u_{\bm{\pi(s + 1)}} \leq e^{- \eta \log(s)^{\frac{1}{1 + \delta}}}.
    \end{equation}
\end{itemize}
\end{theorem}

\begin{proof}
    All rates can be derived by suitably choosing the parameter \(\varepsilon\) in the set \(S(\varepsilon)\), defined in~\eqref{eq: s-term: definition of S_varepsilon}.
    Suppose that \( \varepsilon \leq \sqrt{\lambda_{\bm{b}, 1}} \), so that \(\lambda_{\bm{b}, i} \geq \varepsilon^2\) for every \(i \in [d(\varepsilon)]\) by definition of \(d(\varepsilon)\) (see~\eqref{eq: s-term: effective dimension d}). By the relation~\eqref{eq: s-term: S_varepsilon isomorphic to anisotropic TD index set}, we may use the upper size bound in~\eqref{eq: anisotropic TD index set} to compute
    \begin{equation}
    \label{eq: upper size bound for S_varepsilon}
        \abs{S(\varepsilon)} \leq \prod_{i=1}^{d(\varepsilon)} \left(\frac{\lambda_{\bm{b},i}}{i \varepsilon^2} + 1\right) 
        \leq \prod_{i=1}^{d(\varepsilon)} \frac{\lambda_{\bm{b},i}}{\varepsilon^2} \left(\frac{1}{i} + 1\right) 
        \leq (2 \varepsilon^{-2})^{d(\varepsilon)} \prod_{i=1}^{d(\varepsilon)} \lambda_{\bm{b},i}.
    \end{equation}
    %In the second step we used the fact that \(\lambda_{\bm{b}, i} \geq \varepsilon^2\) for \(i \in [d(\varepsilon)]\) by definition of \(d(\varepsilon)\) (see~\eqref{eq: s-term: effective dimension d}). 
    For brevity, we write in the following \(d = d(\varepsilon)\).

    \emph{Case (a).}
    Using~\eqref{eq: upper size bound for S_varepsilon} as well as the Stirling type estimate \(d^d \leq e^d d!\), we obtain
    \begin{equation}
    \label{eq: upper size bound for S_varepsilon: algebraic}
        \abs{S(\varepsilon)} \leq (2\varepsilon^{-2})^d \prod_{i=1}^d i^{-\alpha}
        = (2\varepsilon^{-2})^d (d!)^{-\alpha}
        \leq (2\varepsilon^{-2})^d e^{\alpha d} d^{-\alpha d}.
    \end{equation}
    Let \(0 < \varepsilon \leq 2^{- \alpha / 2}\). By definition of \(d\), we have \(d \leq \varepsilon^{-2/\alpha} < d + 1 \leq 2d\). We take the logarithm on both sides in~\eqref{eq: upper size bound for S_varepsilon: algebraic} and compute
    \begin{align*}
        \log(\abs{S(\varepsilon)}) 
        &\leq d \log(2 \varepsilon^{-2}) + \alpha d - \alpha d \log(d) \\
        &\leq \varepsilon^{-2/\alpha} \log(2\varepsilon^{-2}) 
        + \alpha \varepsilon^{-2/\alpha} 
        - \frac{1}{2} \alpha \varepsilon^{-2/\alpha} \log\left( \frac{1}{2} \varepsilon^{-2/\alpha} \right) \\
        &= \frac{1}{2} \varepsilon^{-2/\alpha} \log(2 \varepsilon^{-2})
        + \left(\alpha + \frac{1}{2} \log(2^{\alpha + 1})\right) \varepsilon^{-2/\alpha}.
    \end{align*}
    Next, let \(\delta, \eta > 0\) be arbitrary. There exists \(0 < \bar{\varepsilon} = \bar{\varepsilon} (\alpha, \delta, \eta) \leq 2^{- \alpha / 2}\) such that for every \(0 < \varepsilon \leq \bar{\varepsilon}\), we have \(\log(2 \varepsilon^{-2}) \leq \eta^{2(1/\alpha + \delta)} \varepsilon^{-2\delta}\) as well as \(\alpha + \frac{1}{2} \log(2^{\alpha + 1}) \leq \frac{1}{2} \eta^{2(1/\alpha + \delta)} \varepsilon^{-2 \delta}\). It follows that
    \begin{equation}
    \label{eq: log-size bound for S_varepsilon: algebraic}
        \log(\abs{S(\varepsilon)}) 
        \leq \eta^{2(1/\alpha + \delta)} \varepsilon^{-2/\alpha - 2\delta},
        \quad \forall \varepsilon \leq \bar{\varepsilon}.
    \end{equation}
    We set the right-hand side equal to \(\log(s)\) and solve for \(\varepsilon^2\), which gives
    \begin{equation}
    \label{eq: s-term: equation for varepsilon and s: algebraic}
        \varepsilon^2 = \varepsilon(s)^2 = \eta^2 \log(s)^{-\frac{1}{1/\alpha + \delta}}.
    \end{equation}
    We can now choose \(\bar{s} = \bar{s}(\alpha, \delta, \eta) \in \bbN\) sufficiently large such that the right-hand side in~\eqref{eq: s-term: equation for varepsilon and s: algebraic} is smaller than \(\bar{\varepsilon}^2\) for every \(s \geq \bar{s}\). Then,~\eqref{eq: log-size bound for S_varepsilon: algebraic} holds with \(\varepsilon = \varepsilon(s)\) and we conclude \(\absd{S(\varepsilon(s))} \leq s\) for every \(s \geq \bar{s}\).
    Since \(S(\varepsilon) = \pi([\absd{S(\varepsilon)}])\), it follows that \(\bm{\pi(s + 1)} \not\in S(\varepsilon)\) and consequently, 
    \[
    u_{\bm{\pi(s + 1)}}^2 
    \leq (\varepsilon(s)^{-2} + 1)^{-1} 
    \leq \varepsilon(s)^2
    \leq \eta^2 \log(s)^{-\frac{1}{1/\alpha + \delta}},\quad \forall s \geq \bar{s}.
    \]
   
    \emph{Case (b).} By~\eqref{eq: upper size bound for S_varepsilon}, we find
    \begin{equation}
    \label{eq: upper size bound for S_varepsilon: super-exponential}
        \abs{S(\varepsilon)} \leq (2 \varepsilon^{-2})^d \prod_{i=1}^d e^{-\alpha i^{\beta}}
        = (2 \varepsilon^{-2})^d e^{-\alpha \sum_{i=1}^d i^{\beta}}.
    \end{equation}
    Let \(0 < \varepsilon \leq e^{-\alpha / 2}\). By definition of \(d\), we have \(d \leq \alpha^{-1/\beta} \log(\varepsilon^{-2})^{1/\beta} < d + 1\). We take the logarithm on both sides of~\eqref{eq: upper size bound for S_varepsilon: super-exponential} and compute
    \begin{align*}
        \log(\abs{S(\varepsilon)}) 
        &\leq d \log(2\varepsilon^{-2}) - \alpha \sum_{i=1}^d i^{\beta} 
        \leq d \log(2\varepsilon^{-2}) - \alpha \int_0^d t^{\beta} dt \\
        &\leq \alpha^{-1/\beta} \log(\varepsilon^{-2})^{1 + 1/\beta} 
        + \alpha^{-1/\beta} \log(2) \log(\varepsilon^{-2})^{1/\beta} \\
        &\hspace{30ex} - \frac{\alpha}{\beta + 1} ( \alpha^{-1/\beta} \log(\varepsilon^{-2})^{1/\beta} - 1 )^{\beta + 1}.
    \end{align*}
    Next, let \(\widetilde{\delta} > 0\) and \(0 < \eta < 1\) be arbitrary. There exists \(0 < \bar{\varepsilon} = \bar{\varepsilon}(\alpha, \beta, \widetilde{\delta}, \eta) \leq e^{-\alpha/2}\) such that \(\log(2) \leq \widetilde{\delta} \log(\varepsilon^{-2})\) and \(\alpha^{-1/\beta} \log(\varepsilon^{-2})^{1/\beta} - 1 \geq \eta^{\frac{1}{\beta + 1}} \alpha^{-1/\beta} \log(\varepsilon^{-2})^{1/\beta}\) for every \(0 < \varepsilon \leq \bar{\varepsilon}\). Hence,
    \[
    \log(\abs{S(\varepsilon)}) 
    \leq \alpha^{-1/\beta} \left( 1 + \widetilde{\delta} -  \frac{\eta}{\beta + 1} \right) \log(\varepsilon^{-2})^{1 + 1/\beta}, \quad \forall \varepsilon \leq \bar{\varepsilon}.
    \]
    We now set \(\delta = 1 + \widetilde{\delta} (\beta + 1) - \eta\) and conclude similarly as in case (a).
    
    \emph{Case (c).} By~\eqref{eq: upper size bound for S_varepsilon}, we find that
    \begin{equation}
    \label{eq: upper size bound for S_varepsilon: double exponential}
        \abs{S(\varepsilon)} \leq (2 \varepsilon^{-2})^d \prod_{i=1}^d e^{-e^{\alpha i}}
        = (2 \varepsilon^{-2})^d  e^{-\sum_{i=1}^d  e^{\alpha i}}.
    \end{equation}
    Let \(0 < \varepsilon \leq e^{-e^{\alpha} / 2}\). Then, by definition of \(d\), we have \(d \leq \log(\log( \varepsilon^{-2})) / \alpha < d + 1\). We take the logarithm on both sides in~\eqref{eq: upper size bound for S_varepsilon: double exponential} and compute
    \begin{equation}
    \label{eq: upper size bound for S_varepsilon: double exponential II}
        \log(\abs{S(\varepsilon)}) 
        \leq d \log(2 \varepsilon^{-2}) - \sum_{i=1}^d e^{\alpha i} 
        \leq \frac{1}{\alpha} \log(\log(\varepsilon^{-2})) \log(2\varepsilon^{-2}) - \frac{\log(\varepsilon^{-2}) - 1}{e^{\alpha} - 1}  + 1.
    \end{equation}
    Next, let \(\delta,\eta > 0\) be arbitrary. There exists \(0 < \bar{\varepsilon} = \bar{\varepsilon}(\alpha, \delta, \eta) \leq e^{-e^{\alpha} / 2}\) such that for every \(0 < \varepsilon \leq \bar{\varepsilon}\), the right-hand side in~\eqref{eq: upper size bound for S_varepsilon: double exponential II} is dominated by \((2 \eta)^{-(1 + \delta)} \log(\varepsilon^{-2})^{1+\delta}\). We can now conclude similarly as in case (a).
\end{proof}

Similar computations (which we omit for brevity) based on the lower size bound in~\eqref{eq: anisotropic TD index set} show that the bounds in Theorem~\ref{thm: s-term: upper bounds for u_pi(s+1)} are asymptotically sharp in the limit \(s \to \infty\)---at least in the exponential and double exponential case. Indeed, in case (b), for every \(0 < \delta < \beta\), there exists \(\bar{s} = \bar{s}(\alpha, \beta, \delta) \in \bbN\) such that
\begin{equation}
\label{eq: s-term: upper bound: lower bound: exponential}
    u_{\bm{\pi(s + 1)}} \gtrsim e^{- \frac{1}{2} \alpha^{\frac{1}{\beta + 1}} \left( \frac{\beta + 1}{\beta - \delta} \right)^{\frac{1}{1 + 1/\beta}} \log(s+1)^{\frac{1}{1 + 1/\beta}}},
    \quad \forall s \geq \bar{s}.
\end{equation}
Here and henceforth, we write \(x \gtrsim y\) for \(x, y \in \bbR\) if there exists a global constant \(C > 0\), independent of any parameters, such that \(x \geq C y\).
In case (c), for every \(\eta > 0\), there exists \(\bar{s} = \bar{s}(\alpha,\eta) \in \bbN\) such that
\[
u_{\bm{\pi(s + 1)}} \gtrsim (s+1)^{-\eta},
\quad \forall s \geq \bar{s}.
\]
In case (a) with algebraic spectral decay, the lower size bound~\eqref{eq: anisotropic TD index set} is too weak and leads to inconclusive results which do not asymptotically match the upper bound~\eqref{eq: s-term: upper bound: algebraic}.

\subsection{Discussion}
\label{subsec: s-term: discussion}

Theorem~\ref{thm: s-term: lower bound for u_pi(s+1)} shows that \(u_{\bm{\pi(s + 1)}}\) decays subalgebraically as \(s \to \infty\). In particular, as \(p \in \bbN\) can be chosen arbitrarily large, we conclude that the decay is slower than \emph{any} algebraic decay rate for all sufficiently large~\(s\). In combination with Theorem~\ref{thm: s-term: best s-term approximation error}, we deduce the following \textbf{curse of parametric complexity}: \emph{No \(s\)-term Wiener-Hermite PC expansion can converge with an algebraic rate uniformly for all operators in the Sobolev unit (Lipschitz) ball as \(s \to \infty\). This holds regardless of the decay rate of the PCA eigenvalues.}

We have now seen that approximating Lipschitz operators by truncating Wiener-Hermite PC expansions cannot be done efficiently with algebraic convergence rates. 
On the other hand, Theorem~\ref{thm: s-term: upper bounds for u_pi(s+1)} shows the connection between the decay of the eigenvalues \(\lambda_{\bm{b}, i}\) in \(i\) and the decay of \(u_{\bm{\pi(s + 1)}}\) in \(s\). It illustrates the obvious qualitative fact that a faster eigenvalue decay implies a faster decay of the polynomial \(s\)-term error. But it also provides quantitative decay rates and shows that the curse of parametric complexity can be overcome at least asymptotically in the sense that decay rates arbitrarily close to any algebraic rate can be attained, provided the \(\lambda_{\bm{b}, i}\) decay sufficiently fast. Since the lower bound~\eqref{eq: s-term: upper bound: lower bound: exponential} asymptotically matches the upper bound~\eqref{eq: s-term: upper bound: exponential}, we see, however, that to this end, the eigenvalue decay has to be faster than exponential. Equation~\eqref{eq: s-term: upper bound: double exponential} shows that a double exponential decay is sufficient.

In the context of the above-stated curse of parametric complexity, the parameters, that is, the polynomial coefficients in the truncated Wiener-Hermite PC expansion, are elements of \(\cY\). In~\citet{lanthaler_OperatorLearningPCANetUpper_2023}, a related curse of parametric complexity for learning bounded Lipschitz operators with PCA-Net was proved, which for distinction we call~\emph{curse of parametric PCA-Net complexity}. It relates the learnability of Lipschitz operators by PCA-Nets to the size, i.e., the number of neural network parameters, of the latter. For succinctness, we refer to~\citet{lanthaler_OperatorLearningPCANetUpper_2023} for more details and definitions. The following result follows from Theorem 9 therein.

\begin{theorem}[Curse of parametric PCA-Net complexity]
\label{thm: curse of parametric complexity; PCA-Net}
    For any \(\alpha > 0\), there exists a bounded Lipschitz operator \(F \in C^{0, 1}(\cX; \cY)\) and a constant \(c_{\alpha} > 0\) such that 
    \[
    \nm{F - \Psi}_{L_{\mu}^2(\cX; \cY)}^2 \geq c_{\alpha} \left(\textup{size}(\psi)\right)^{-\alpha}
    \]
    for every PCA-Net \(\Psi = \cD_{\cY} \circ \psi \circ \cE_{\cX}\). Here, \(\psi: \bbR^{d_{\cX}} \to \bbR^{d_{\cY}}\) is a (ReLU) neural network and \(\cE_{\cX}: \cX \to \bbR^{d_{\cX}}\) and \(\cD_{\cY}: \bbR^{d_{\cY}} \to \cY\) are encoder and decoder mappings, respectively, which are defined in terms of the empirical PCA eigenvalues based on finitely many fixed sample points. 
\end{theorem} 

\begin{remark}
    Inspection of the proof of~\citet[Thm. 9]{lanthaler_OperatorLearningPCANetUpper_2023} shows that the empirical PCA eigenvalues used in the definition of \(\cE_{\cX}, \cD_{\cY}\) can be replaced by the true PCA eigenvalues, which puts Theorem~\ref{thm: curse of parametric complexity; PCA-Net} in the setting of the present paper.
\end{remark}

We now argue that the curse of parametric complexity described by Theorems~\ref{thm: s-term: best s-term approximation error} and~\ref{thm: s-term: lower bound for u_pi(s+1)} is consistent with the curse of parametric PCA-Net complexity described by Theorem~\ref{thm: curse of parametric complexity; PCA-Net}.
To this end, suppose that Hermite polynomial \(s\)-term approximations of Lipschitz operators in the Sobolev unit Lipschitz ball were possible with some algebraic rate of order \(\beta > 0\):

\begin{assumption}[Algebraic \(s\)-term convergence for Lipschitz operators]
\label{ass: Hermite expansion with algebraic rate}
    There exists \(\beta > 0\) such that for every \(F \in B_{\mu, \bm{b}}^{\Lip}(\cX; \cY)\), there are constants \(c(F) > 0\) and \(\bar{s}(F) \in \bbN\) such that 
    \begin{equation}
    \label{eq: Hermite expansion with algebraic rate}
        \nm{F - F_{\pi([s])}}_{L_{\mu}^2(\cX; \cY)} \leq c(F) s^{-\beta}, \quad \forall s \geq \bar{s}(F).
    \end{equation}
\end{assumption}

\begin{proposition}
\label{prop: algebraic convergence rate for C01}
    Grant Assumption~\ref{ass: Hermite expansion with algebraic rate}.
    Then there exists \(\alpha > 0\) with the following property: For all \(F \in C^{0, 1}(\cX; \cY)\), there exist constants \(C = C(F, \alpha) > 0\) and \(\bar{\epsilon} = \bar{\epsilon}(F, \alpha) > 0\) such that for all \(0 < \epsilon \leq \bar{\epsilon}\), there is a PCA-Net \(\Psi = \widetilde{\cD}_{\cY} \circ \psi \circ \widetilde{\cE}_{\cX}\) such that 
    \[
    \nm{F - \Psi}_{L_{\mu}^2(\cX; \cY)} \leq \epsilon \quad \textup{and} \quad 
    \textup{size}(\psi) \leq C \epsilon^{-1/\alpha}.
    \]
\end{proposition}

With the notion introduced in~\citet{lanthaler_OperatorLearningPCANetUpper_2023}, Proposition~\ref{prop: algebraic convergence rate for C01} asserts that \(\alpha\) is an algebraic convergence rate for the class of \(C^{0, 1}(\cX; \cY)\)-operators.
It is easy to see that this leads to a contradiction to Theorem~\ref{thm: curse of parametric complexity; PCA-Net}, hence implying Assumption~\ref{ass: Hermite expansion with algebraic rate} to be false.
The proof of Proposition~\ref{prop: algebraic convergence rate for C01} is based on~\citet[Thm. 3.9]{schwab_DeepLearningHighDimension_2023}, which provides expression rate bounds for the approximation of multivariate Hermite polynomials by ReLU neural networks. It is given in Appendix~\ref{subsec: app: proof algebraic convergence}.

%%%%%%%%%%%%%%%%%%%%%%%%%%%%%%%%%%%%%%%%%%%%%%%%%%%%%%%%%%%%%%%%%%%%%%%%%%%%%%%%%%%%%%%%%%
%%%%%%%%%%%%%%%%%%%%%%%%%%%%%%%%%%%%%%%%%%%%%%%%%%%%%%%%%%%%%%%%%%%%%%%%%%%%%%%%%%%%%%%%%%

\section{Optimal sampling and (adaptive) \texorpdfstring{\(m\)}{m}-widths}
\label{sec: optimal sampling and adaptive m-widths}

Up to this point, we have studied the best approximation of \(W_{\mu, \bm{b}}^{1, 2}\)- and Lipschitz operators by Hermite polynomials. This is a certain type of what in IBC is referred to as \emph{linear information}~\citep[Ch. 4.1.1]{novak_TractabilityMultivariateProblemsVolume_2008}. In this section, we consider more general sampling and reconstruction schemes based on \emph{adaptive information}, that is, (nonlinear) reconstruction from \(m\) adaptively chosen samples. We define adaptive sampling operators and the adaptive \(m\)-width and characterize the latter in terms of the weights \(u_{\bm{\gamma}}\). We follow in parts ideas from~\citet{adcock_OptimalApproximationInfinitedimensionalHolomorphic_2024}, who studied holomorphic operators.

\subsection{Adaptive sampling operators}

\begin{definition}[{Adaptive sampling operator; scalar-valued case}]
\label{def: adaptive sampling operator; scalar-valued}
    Let \((\cV,\nm{\cdot}_\cV)\) be a normed vector space and \(m\in\bbN\). A (scalar-valued) \emph{adaptive sampling operator} is a mapping of the form 
    \[
    \cL: \cV \to \bbR^m, \quad \cL(F) = 
    \begin{pmatrix}
        \cL_1(F) \\
        \cL_2(F; \cL_1(F))\\
        \vdots\\
        \cL_m(F; \cL_1(F), \dots, \cL_{m-1}(F))
    \end{pmatrix},
    \]
    where \(\cL_1: \cV \to \bbR\) is a bounded linear functional and \(\cL_i: \cV \times \bbR^{i-1} \to \bbR\) is bounded and linear in its first component for \(i = 2, \dots, m\).
\end{definition}

Trivially, any bounded linear mapping \(\cL: \cV \to \bbR^m\) is an adaptive sampling operator (which, in fact, generates non-adaptive information).
Different choices for \(\cV\) lead to important special cases. If \(\cV = L_{\mu}^2(\cX)\) and \(\bm{\gamma}^{\bm{(1)}}, \dots, \bm{\gamma}^{\bm{(m)}} \in \Gamma\), we can define a sampling operator, generating (non-adaptive) \emph{linear information}, by
\begin{equation}
\label{eq: adaptive sampling: linear sampling operator}
    \cL(F) := \left(\int_{\cX} F H_{\bm{\gamma}^{\bm{(i)}}, \bm{\lambda}} \D \mu \right)_{i = 1}^m \in \bbR^m, \quad \forall F \in L_{\mu}^2(\cX).
\end{equation}
Throughout the text, we denote by \( C_b(\cX; \cY) \) the space of all bounded continuous operators \( \cX \to \cY \), equipped with the uniform norm. We write \(C_b(\cX) := C_b(\cX; \bbR) \). If \(\cV = C_b(\cX)\) and \(X_1, \dots, X_m \in \cX\), we can define a pointwise sampling operator, generating (non-adaptive) \emph{standard information}~\citep[Ch. 4.1.1]{novak_TractabilityMultivariateProblemsVolume_2008}, by
\begin{equation}
\label{eq: adaptive sampling: pointwise sampling operator}
    \cL(F) := \left(F(X_i)\right)_{i = 1}^m \in \bbR^m, \quad \forall F \in C_b(\cX).
\end{equation}
In both cases above, the \(\bm{\gamma}^{\bm{(i)}}\) and the \(X_i\) can, in principle, also be chosen adaptively based on previous measurements \(\int_{\cX} F H_{\bm{\gamma}^{\bm{(j)}}, \bm{\lambda}} \D \mu\) and \(F(X_j)\), respectively, for \(j \in [i-1]\).

Next, we consider the Hilbert-valued case. For any \(Y \in \cY\), \(\bm{v} = (v_i)_{i = 1}^m \in \bbR^m\), and \(F \in L_{\mu}^2(\cX)\), we write \(Y \bm{v}\) for the vector \((Y v_i)_{i = 1}^m \in \cY^m\) and \(Y F\) for the mapping \(X \mapsto Y F(X)\). 

\begin{definition}[{Adaptive sampling operator; Hilbert-valued case}]
\label{def: adaptive sampling operator; Hilbert-valued}
    Let \(\cV \subset L_{\mu}^2(\cX; \cY)\) be a vector subspace with norm \(\nm{\cdot}_{\cV}\) and consider a mapping
    \[
    \cL: \cV \to \cY^m, \;\; \cL(F) = 
    \begin{pmatrix}
        \cL_1(F) \\
        \cL_2(F; \cL_1(F))\\
        \vdots\\
        \cL_m(F; \cL_1(F), \dots, \cL_{m-1}(F))
    \end{pmatrix},
    \]
    where \(\cL_1: \cV \to \cY\) is a bounded linear operator and \(\cL_i: \cV \times \cY^{i-1} \to \cY\) is bounded and linear in its first component for \(i = 2, \dots, m\). Then \(\cL\) is a \emph{Hilbert-valued adaptive sampling operator} if the following condition holds: There exist \(Y, \wtY \in \cY \setminus \{0\}\), a normed vector space \(\widetilde{\cV} \subset L_{\mu}^2(\cX)\), and a scalar-valued adaptive sampling operator \(\widetilde{\cL}: \widetilde{\cV} \to \bbR^m\) such that, if \(Y F \in \cV\) for some \(F \in L_{\mu}^2(\cX)\), then \(F \in \widetilde{\cV}\) and  \(\cL(Y F) = \wtY \widetilde{\cL}(F)\).
\end{definition}

This definition involves a technical assumption which links the Hilbert-valued case to the scalar-valued case and which we will use to establish a lower bound for the adaptive \(m\)-width. However, this condition is not too strong. It holds, for example, in the case of adaptive pointwise sampling. Here, we choose \(\cV = C_b(\cX; \cY)\) (seen as a subspace of \(L_{\mu}^2(\cX; \cY)\) and equipped with the \(L^{\infty}\)-norm) and define
\[
\cL(F) := \left(F(X_i)\right)_{i = 1}^m \in \cY^m, \quad \forall F \in \cV,
\]
where the \(i\)th sample point \(X_i\) is potentially chosen based on the previous measurements \(F(X_1),\dots,F(X_{i-1})\). We then have
\[
\cL(YF) = Y \widetilde{\cL}(F), \quad \forall F \in \widetilde{\cV} := C_b(\cX), \forall Y \in \cY,
\]
where \(\widetilde{\cL}: \widetilde{\cV} \to \bbR^m\) is the adaptive pointwise sampling operator in~\eqref{eq: adaptive sampling: pointwise sampling operator}. 
As another example, we will see in the proof of the upper bound of the adaptive \(m\)-width that the Hilbert-valued version of linear sampling, as defined in~\eqref{eq: adaptive sampling: linear sampling operator}, is a Hilbert-valued sampling operator in the sense of Definition~\ref{def: adaptive sampling operator; Hilbert-valued} (under a mild condition on \(\cV\)), see Section~\ref{subsubsec: m-width: upper bound}.

\subsection{Adaptive \texorpdfstring{\(m\)}{m}-widths and main result}
\label{subsec: adaptive widths and main result}

We now formally define the adaptive \(m\)-width and state our main result.

\begin{definition}[Adaptive \(m\)-width]
    Let \((\cV,\nm{\cdot}_{\cV})\) be a normed vector subspace of \(L_{\mu}^2(\cX;\cY)\) and let \(\cK \subset \cV\) be a subset.
    The adaptive \(m\)-width of \(\cK\) in \(\cV\) is given by
    \begin{align}
    \begin{split}
    \label{eq: m-width: definition of adaptive m-width}
        &\Theta_m(\cK;\cV,L_{\mu}^2(\cX; \cY)) \\
        &:= \inf\left\{
        \sup_{F \in \cK} \nm{F - \cT(\cL(F))}_{L_{\mu}^2(\cX; \cY)} : \ \cL: \cV \to \cY^m \textup{ adaptive}, \cT: \cY^m \to L_{\mu}^2(\cX; \cY)
        \right\}.
    \end{split}
    \end{align}
\end{definition}

The adaptive \(m\)-width describes the smallest worst-case error that can be achieved when we reconstruct all operators in a set \(\cK\) by a reconstruction mapping \(\cT\) from \(m\) samples that have been generated by an adaptive Hilbert-valued sampling operator \(\cL\). Note that in~\eqref{eq: m-width: definition of adaptive m-width} we allow for any (possibly nonlinear) reconstruction mappings. The choice of \(\cV\), however, determines which sampling operators are allowed. If \(\cV = C_b(\cX; \cY)\), we can use pointwise sampling, whereas, if \(\cV = L_{\mu}^2(\cX; \cY)\), this is not possible.

We remark that the term ``adaptive \(m\)-width'' is not standard terminology in the literature. For better understanding, we put it into the context of other widths in nonlinear approximation. 
The adaptive \(m\)-width is the Hilbert-valued version of the adaptive compressive \(m\)-width~\citep{foucart_MathematicalIntroductionCompressiveSensing_2013}, which we introduce in Section~\ref{subsubsec: m-width: results about widths} where we also discuss its relation to the Gelfand and Kolmogorov width.
If we set \( \cY = \bbR \) in~\eqref{eq: m-width: definition of adaptive m-width} and assume \(\cL\) to be a general continuous mapping \( \cK \to \bbR^m\) instead of a (scalar-valued) adaptive sampling operator on \( \cV \), then we obtain the~\emph{continuous nonlinear \(m\)-width} as introduced in~\citet{devore_OptimalNonlinearApproximation_1989}. If we assume in addition that \( \cT \) is continuous (and \( \cK \) is compact), then we obtain the~\emph{nonlinear \(m\)-width} (or~\emph{manifold width}) in~\citet{devore_NonlinearApproximation_1998}.
If \( \cL : \cK \to \bbR^m \) and \( \cT : \bbR^m \to L^2_{\mu}(\cX) \) are instead required to be bounded linear mappings, the resulting quantity is the~\emph{linear \(m\)-width}~\citep[Def. 4.1]{pinkus_NwidthsApproximationTheory_1985}.
The various \(m\)-width notions therefore differ in two aspects: the type of information available about the target operator, and how that information may be used to reconstruct it. In this work, we study the adaptive \(m\)-width, since (adaptively chosen) linear samples combined with general reconstruction strategies are of high importance in practical applications.

Recently,~\citet{kovachki_DataComplexityEstimatesOperator_2024} studied the so-called~\emph{sampling nonlinear \(m\)-width} \(s_m(\cK)_{L_{\mu}^2(\cX)}\) of a set \(\cK \subset C_b(\cX)\), which is based on standard information. In principle, it results from~\eqref{eq: m-width: definition of adaptive m-width} in the case \( \cY = \bbR \) by replacing the adaptive sampling operator \( \cL : \cV \to \bbR^m \) by a pointwise sampling operator \(\delta_{\bm{X}}: \cK \to \bbR^m\). More specifically, for fixed \(\bm{X} = (X_1, \dots, X_m) \in \cX^m\), define \(\delta_{\bm{X}}(F) = (F(X_1), \dots, F(X_m)) \in \bbR^m\) for every \(F \in \cK\), and set
\begin{equation}
\label{eq: sampling nonlinear width}
    s_m(\cK)_{L_{\mu}^2(\cX)} := 
    \inf \left\{ \sup_{F \in \cK} \nm{F - \cT(\delta_{\bm{X}}(F))}_{L_{\mu}^2(\cX)}: \bm{X} \in \cX^m, \cT: \bbR^m \to L_{\mu}^2(\cX) \right\}.
\end{equation}
The definition in~\citet{kovachki_DataComplexityEstimatesOperator_2024} is slightly more general and we refer to the original article for more details.
Since the adaptive \( m \)-width includes recovery strategies based on standard information if we choose \( \cV = C_b(\cX) \), it constitutes a lower bound for the sampling nonlinear \(m\)-width: \( s_m(\cK)_{L_{\mu}^2(\cX)} \geq \Theta_m(\cK; C_b(\cX), L_{\mu}^2(\cX)) \). This relation allows us to use lower bounds on the adaptive \(m\)-width (Theorems~\ref{thm: adaptive m-width} and~\ref{thm: curse of data complexity}) to generalize some results in~\citet{kovachki_DataComplexityEstimatesOperator_2024}. Further details are discussed in Section~\ref{subsec: m-width: discussion}. 

Our main result in this section is the tight characterization of the adaptive \(m\)-width of the Sobolev unit (Lipschitz) ball, defined in~\eqref{eq: Sobolev unit (Lipschitz) ball}, in terms of the weights \(u_{\bm{\gamma}}\), given by~\eqref{eq: Sobolev weights}. The lower bound for the adaptive \(m\)-width pertains to arbitrary \(\cV\) which contain the Sobolev unit (Lipschitz) ball. For the upper bound, we require the additional assumption that \(\cV\) is continuously embedded in \(L_{\mu}^2(\cX; \cY)\).

\begin{theorem}[Tight characterization of adaptive \(m\)-width]
\label{thm: adaptive m-width}
    Let \(\cK \in \{ B_{\mu, \bm{b}}(\cX; \cY), B_{\mu, \bm{b}}^{\Lip}(\cX; \cY) \}\) and let \( (\cV, \nm{\cdot}_{\cV}) \) be a normed vector subspace of \( L^2_{\mu}(\cX; \cY) \) containing \( \cK \). For every \(m \in \bbN\), we have the lower bound 
    \[
    \Theta_m(\cK; \cV, L_{\mu}^2(\cX; \cY)) \geq u_{\bm{\pi(m+1)}},
    \]
    where \(\pi: \bbN \to \Gamma\) is a bijection which gives a nonincreasing rearrangement of \(\bm{u} = (u_{\bm{\gamma}})_{\bm{\gamma} \in \Gamma}\), see~\eqref{eq: nonincreasing rearrangement pi}.
    If, in addition, \(\cV\) is continuously embedded in \(L_{\mu}^2(\cX; \cY)\), we have for every \(m \in \bbN\) the matching upper bound
    \begin{align*}
        \Theta_m(\cK; \cV, L_{\mu}^2(\cX; \cY)) 
        &\leq \inf_{S \subset \Gamma, \abs{S} \leq m} \sup_{F \in \cK} \nm{F - F_S}_{L_{\mu}^2(\cX; \cY)} \\
        &\leq \sup_{F \in \cK} \nm{F - F_{\{\bm{\pi(1)}, \dots, \bm{\pi(m)}\}}}_{L_{\mu}^2(\cX; \cY)} 
        \leq u_{\bm{\pi(m+1)}}.
    \end{align*}
\end{theorem}

Suppose in the following that \( \cV \) is continuously embedded in \( L^2_{\mu}(\cX; \cY) \).
It is worth noting that in this case the adaptive \(m\)-width of the Sobolev unit (Lipschitz) ball is not only bounded from above and below by \( u_{\bm{\pi(m + 1)}} \) up to constants, as proven in~\citet{adcock_OptimalApproximationInfinitedimensionalHolomorphic_2024} for holomorphic operators, but it is exactly equal to \( u_{\bm{\pi(m + 1)}} \). 
A similar characterization holds for the linear \(m\)-width of unit balls of weighted Hilbert spaces~\citep[Thm. 2.2]{pinkus_NwidthsApproximationTheory_1985}. See also~\cite[Thm. 11.11.3]{pietsch_OperatorIdeals_1980} for an equivalent formulation in terms of approximation numbers of diagonal operators between \(\ell^p\)-spaces.
Hence, Theorem~\ref{thm: adaptive m-width} can be interpreted as an extension of this result to the setting of adaptive sampling and general (possibly nonlinear) reconstruction strategies. We highlight, however, two aspects that show that, besides this extension, the reach of Theorem~\ref{thm: adaptive m-width} is considerably broader. First, it covers the recovery of operators from \(\cY\)-valued samples (not only scalar-valued ones). Second, it also applies to the Sobolev unit Lipschitz ball, which is the unit ball of a pre-Hilbert space, but not of a Hilbert space (as required by the linear \(m\)-width characterization).

Theorem~\ref{thm: adaptive m-width} has important implications for the role of adaptivity in our approximation problem. It is well-known in IBC that ``adaptivity does not help'', that is, adaptive methods in approximation problems can be better than non-adaptive methods in terms of the smallest worst-case error only by a factor of at most \(2\).
We refer to~\citet{novak_PowerAdaption_1996} and~\citet{krieg_PowerAdaptionRandomization_2025} and references therein. For non-empty, convex, symmetric subsets of pre-Hilbert spaces, the factor \(2\) can in fact be reduced to \(1\)~\citep[Thm. 4.5]{novak_TractabilityMultivariateProblemsVolume_2008}. That is, adaptivity does not even help to improve the constants in the error bounds. 
We emphasize that these ``adaptivity does not help'' results hold for recovery strategies based on scalar-valued information. Theorem~\ref{thm: adaptive m-width} extends this principle to the recovery of operators from \(\cY\)-valued information which is generated by Hilbert-valued adaptive sampling operators, as defined in Definition~\ref{def: adaptive sampling operator; Hilbert-valued}. Indeed, it shows that the smallest worst-case error in our approximation problem is equal to \( u_{\bm{\pi(m + 1)}} \), and it is attained by the non-adaptive linear projection \( F \mapsto F_S \), see~\eqref{eq: L2 projection}, with \( S = \{ \bm{\pi(1)}, \dots, \bm{\pi(m)} \} \). 

Finally, we remark that Theorem~\ref{thm: adaptive m-width} implies that the adaptive \(m\)-width of both the Sobolev unit ball and the Sobolev unit Lipschitz ball coincide. From an IBC point of view, there is thus no difference between the space \( W_{\mu, \bm{b}}^{1,2}(\cX; \cY) \) and the space \( \Lip(\cX;\cY) \), equipped with the \( W_{\mu, \bm{b}}^{1,2}(\cX; \cY) \)-norm, in terms of how well elements in these spaces can be approximated by (adaptively chosen) linear samples.

\subsection{Proof of Theorem~\ref{thm: adaptive m-width}}

We briefly outline the idea of the proof of Theorem~\ref{thm: adaptive m-width}. The upper bound follows standard computations. It suffices to show that the \(L^2_{\mu}\)-projection \( F \mapsto F_S \), defined in~\eqref{eq: L2 projection}, can be written as the composition of a (non-adaptive) sampling operator and a suitable reconstruction mapping. Details are provided in Section~\ref{subsubsec: m-width: upper bound}. 

The lower bound is less obvious. We first reduce the continuous problem to a discrete one in \(N\) dimensions by considering a suitable Lipschitz operator. The adaptive \(m\)-width of the continuous problem can be bounded from below by the adaptive compressive \(m\)-width of the discrete problem. The latter, in turn, is lower bounded by the Gelfand width of certain weighted balls in \( \bbR^N \) with weights given by the \( u_{\bm{\gamma}} \). We then use known duality relations between the Gelfand and Kolmogorov width and an explicit characterization of the latter in terms of the \( u_{\bm{\gamma}} \) to conclude the proof. We recall relevant notions and results about widths in Section~\ref{subsubsec: m-width: results about widths}. Further information can be found in~\citet[Ch. 10]{foucart_MathematicalIntroductionCompressiveSensing_2013}. The proof of the lower bound is provided in Section~\ref{subsubsec: m-width: lower bound}.

\subsubsection{Results about widths}
\label{subsubsec: m-width: results about widths} 

Let \(\cK\) be a subset of a normed vector space \((\cZ, \nm{\cdot}_{\cZ})\) and let \(m \in \bbN\).
The \emph{Gelfand \(m\)-width} of \(\cK\) is defined~by
\[
d^m(\cK, \cZ) := \inf\left\{\sup_{Z \in \cK \cap L^m} \nm{Z}_\cZ: L^m \textup{ subspace of } \cZ \textup{ with codim}(L^m) \leq m\right\}.
\]
An equivalent characterization is given by
\[
d^m(\cK, \cZ) = \inf\left\{\sup_{Z \in \cK \cap \ker(A)} \nm{Z}_{\cZ}: A: \cZ \to \bbR^m \textup{ linear} \right\}.
\]
We also recall the \emph{adaptive compressive \(m\)-width} of \(\cK\),
\[
E_{\textup{ada}}^m(\cK, \cZ) := \inf\left\{\sup_{Z \in \cK} \nm{Z - \Delta(\Xi(Z))}_{\cZ}: \Xi: \cZ \to \bbR^m \textup{ adaptive, } \Delta: \bbR^m \to \cZ \right\},
\]
and the \emph{Kolmogorov \(m\)-width} of \(\cK\),
\[
d_m(\cK,\cZ) := 
\inf\left\{
\sup_{K \in \cK} \inf_{Z \in \cZ_m} \nm{Z - K}_{\cZ}: \cZ_m \textup{ subspace of } \cZ \textup{ with } \dim(\cZ_m) \leq m 
\right\}.
\]
Next, we state some standard results which relate these various \(m\)-width notions. 
\begin{theorem}[{\citet[Thm. 10.4]{foucart_MathematicalIntroductionCompressiveSensing_2013}}]
\label{thm: Gelfand width <= adaptive compressive width}
    If \(\cK\) is symmetric with respect to the origin, i.e., \(- \cK = \cK\), then 
    \(
    d^m(\cK, \cZ) \leq E_{\textup{ada}}^m(\cK, \cZ).
    \)
\end{theorem}

Pietsch~\citep[Thm. 7.2]{pietsch_SNumbersOperatorsBanachSpaces_1974} and Stesin~\citep[Thm. 3]{stesin_AleksandrovDiametersFinitedimensionalSets_1975} found explicit characterizations of the Kolmogorov \(m\)-width in finite sequence spaces. For this, let us introduce the following notation.
For \(1 \leq p \leq \infty\), an index set \(I\), and positive weights \(\bm{w} = (w_i)_{i \in I} > \bm{0}\), we define the space \(\ell_{\bm{w}}^p(I)\) as the set of all sequences of real numbers \(\bm{x} = (x_i)_{i \in I}\) whose norm \(\nm{\bm{x}}_{\ell_{\bm{w}}^p(I)}\) is finite, where
\[
\nm{\bm{x}}_{\ell_{\bm{w}}^p(I)} :=
\begin{cases}
    \left( \sum_{i \in I} w_i^{-p} \abs{x_i}^p \right)^{1/p} & \textup{ if } 1 \leq p < \infty, \\
    \sup_{i \in I} \left\{w_i^{-1} \abs{x_i}\right\} & \textup{ if } p = \infty.
\end{cases}
\]
We denote the closed unit ball in \(\ell_{\bm{w}}^p(I)\) by
\[
B_{\bm{w}}^p(I) := \left\{\bm{x} \in \ell_{\bm{w}}^p(I): \nm{\bm{x}}_{\ell_{\bm{w}}^p(I)} \leq 1 \right\}.
\]
If \(\bm{w} = \bm{1}\), we write \((\ell^p(I), \nm{\cdot}_{\ell^p(I)})\) and~\(B^p(I)\).

\begin{theorem}[Characterization of Kolmogorov width]
\label{thm: Stesin}
    Let \(N \in \bbN\) with \(N > m\), \(1\leq q < p \leq \infty\), and \(\bm{w}\in\bbR^N\) be a vector of positive weights. Then,
    \[
    d_m(B_{\bm{w}}^p([N]), \ell^q([N])) = \left(\max_{\substack{i_1,\dots,i_{N-m}\in[N] \\ i_k \neq i_j}} \left(\sum_{j=1}^{N-m} w_{i_j}^{\frac{pq}{p-q}}\right)^{\frac{1}{p} - \frac{1}{q}}\right)^{-1}.
    \]
\end{theorem}
Pietsch showed that the Kolmogorov width and the Gelfand width are dual to each other~\citep[Thm. 6.2]{pietsch_SNumbersOperatorsBanachSpaces_1974}. For a direct proof of this fact in our setting we also refer to~\citet[Thm. B.3]{adcock_OptimalApproximationInfinitedimensionalHolomorphic_2024}. For a (possibly finite) sequence \(\bm{w} = (w_i)_{i \in I}\) of nonvanishing real numbers, we write \(1/\bm{w} := (1/w_i)_{i \in I}\) for the sequence of reciprocals.
\begin{theorem}[Duality of Kolmogorov width and Gelfand width]
\label{thm: Gelfand width dual Kolmogorov width}
    For \(1\leq p,q \leq \infty\), let \(\bm{w}\in\bbR^N\) be a vector of positive weights and let \(1 \leq p^*, q^* \leq \infty\) be such that \(1/p + 1/p^* = 1\) and \(1/q + 1/q^* = 1\). Then
    \(
    d_m(B^p([N]), \ell_{\bm{w}}^q([N])) = d^m(B_{1/\bm{w}}^{q^*}([N]), \ell^{p^*}([N])). 
    \)
\end{theorem}

Finally, we need one more technical lemma which states that suitably changing weights in sequence spaces does not change the Kolmogorov width.

\begin{lemma}[{\citet[Lem. B.4]{adcock_OptimalApproximationInfinitedimensionalHolomorphic_2024}}]
\label{lem: Kolmogorov width: equality}
    Let \(\bm{w}\in\bbR^N\) be a vector of positive weights and \( 1 \leq p,q \leq \infty \). Then
    \(
    d_m(B^p([N]), \ell_{\bm{w}}^q([N])) = d_m(B_{1/\bm{w}}^p([N]), \ell^q([N])).
    \)
\end{lemma}

\subsubsection{Lower bound}
\label{subsubsec: m-width: lower bound}

Since \(B_{\mu, \bm{b}}^{\Lip}(\cX; \cY)\) is a subset of \(B_{\mu, \bm{b}}(\cX; \cY)\), it suffices to prove the lower bound for the adaptive \(m\)-width in the case \(\cK = B_{\mu, \bm{b}}^{\Lip}(\cX; \cY)\), that is,
\begin{equation}
\label{eq: m-width: lower bound II}
    \Theta_m(B_{\mu, \bm{b}}^{\Lip}(\cX; \cY); \cV, L_{\mu}^2(\cX; \cY)) \geq u_{\bm{\pi(m+1)}},
    \quad \forall m \in \bbN.
\end{equation}
With the notions and results from Section~\ref{subsubsec: m-width: results about widths}, we can concretize the proof strategy that we outlined previously. We proceed in two main steps. We first reduce the continuous problem to a discrete one which can be analyzed by the adaptive compressive \(m\)-width of the unit ball in a space of suitably weighted finite sequences, see the proof of Lemma~\ref{lem: m-width: discrete problem}. In the discrete setting, we apply Theorems~\ref{thm: Gelfand width <= adaptive compressive width} and~\ref{thm: Gelfand width dual Kolmogorov width} as well as Lemma~\ref{lem: Kolmogorov width: equality} to relate the adaptive compressive \(m\)-width to the Kolmogorov \(m\)-width. In the last step, we use the characterization of the latter, provided by Theorem~\ref{thm: Stesin}, in combination with a limiting argument to conclude the claim.

For the next result, let us introduce the notation \((c \bm{u})_{I}\) with \(\bm{u} = (u_{\bm{\gamma}})_{\bm{\gamma} \in \Gamma}\), \(c \in \bbR\), and \(I \subset \Gamma\) to denote the scaled subsequence \((c u_{\bm{\gamma}})_{\bm{\gamma} \in I}\).

\begin{lemma}[Reduction to discrete problem]
\label{lem: m-width: discrete problem}
    Let \(I\subset\Gamma\) be a finite index set. Then, for every constant \(c \in (0,1)\), we have
    \[
    \Theta_m(B_{\mu, \bm{b}}^{\Lip}(\cX; \cY); \cV, L_{\mu}^2(\cX; \cY)) \geq d^m(B_{(c\bm{u})_I}^2(I), \ell^2(I)).
    \]
\end{lemma}

The proof of this lemma is based on the construction of a suitable Lipschitz operator. To this end, let \(R > 0\) and \(n \in \bbN\), and consider the capped one-dimensional Hermite polynomials
\begin{equation}
\label{eq: m-width: definition wtH_n,R}
    \wtH_{n,R}(x) := 
    \begin{cases}
        H_n(x) & \textup{ if } -R \leq x \leq R, \\
        H_n(R) & \textup{ if } x > R, \\
        H_n(-R) & \textup{ if } x < -R.
    \end{cases}
\end{equation}
For \(\bm{\gamma}\in\Gamma\) and \(d \in \bbN\), we define
\begin{equation}
\label{eq: m-width: definition of wtH_gamma,R,d}
    \wtH_{\bm{\gamma}, R, d}: \bbR^d \to \bbR, \quad 
    \wtH_{\bm{\gamma}, R, d}(\bm{x}) := \prod_{i=1}^d \wtH_{\gamma_i, R} (x_i),
\end{equation}
as well as 
\begin{equation}
\label{eq: m-width: definition of wtH_gamma,R,lambda}
    \wtH_{\bm{\gamma}, R, \bm{\lambda}}: \cX \to \bbR, \quad 
    \wtH_{\bm{\gamma}, R, \bm{\lambda}}(X)
    := \prod_{i = 1}^{\infty} \wtH_{\gamma_i, R} \left(\frac{\ip{X}{\phi_i}_{\cX}}{\sqrt{\lambda_i}}\right).
\end{equation}

Before proving Lemma~\ref{lem: m-width: discrete problem}, we need a couple of preliminary results.

\begin{lemma}[Lipschitz continuity]
\label{lem: Lipschitz continuity of capped Hermite polynomials}
    For every \(R > 0\) and every \(\bm{\gamma} \in \Gamma\), the functional \(\wtH_{\bm{\gamma}, R, \bm{\lambda}} : \cX \to \bbR\), defined in~\eqref{eq: m-width: definition of wtH_gamma,R,lambda}, is Lipschitz continuous.
\end{lemma}

\begin{proof}
    Fix \(R > 0\) and \(\bm{\gamma} \in \Gamma\) with \(\supp(\bm{\gamma}) \subset [d]\) for some \(d \in \bbN\). We define the scaling functional
    \[
    S_{\bm{\lambda}, d}: \cX \to \bbR^d, \quad S_{\bm{\lambda}, d}(X) := \left(\frac{\ip{X}{\phi_i}_{\cX}}{\sqrt{\lambda_i}}\right)_{i = 1}^d,
    \]
    and note that
    \(
    \wtH_{\bm{\gamma}, R, \bm{\lambda}} 
    = \wtH_{\bm{\gamma}, R, d} \circ S_{\bm{\lambda}, d},
    \)
    with \(\wtH_{\bm{\gamma}, R, d}\) given by~\eqref{eq: m-width: definition of wtH_gamma,R,d}.
    As \(S_{\bm{\lambda}, d}\) is Lipschitz continuous, it suffices to show that \(\wtH_{\bm{\gamma}, R, d}\) is Lipschitz continuous. This, in turn, follows by a simple induction argument on \(d\).
\end{proof}

By Lemma~\ref{lem: Lipschitz continuity of capped Hermite polynomials} and Theorem~\ref{thm: Lipschitz operators are Gaussian Sobolev operators}, we have \(\wtH_{\bm{\gamma}, R, \bm{\lambda}} \in W_{\mu, \bm{b}}^{1,2}(\cX) \) for every \(R>0\). By Lemma~\ref{lem: Hermite polynomials in W12_mu}, we also know that \(H_{\bm{\gamma}, \bm{\lambda}} \in W^{1,2}_{\mu, \bm{b}}(\cX) \). The next result establishes the connection between \(\wtH_{\bm{\gamma}, R, \bm{\lambda}}\) and \(H_{\bm{\gamma}, \bm{\lambda}}\) in the limit \(R \to \infty\). The proof is given in Appendix~\ref{subsec: app: proof convergence}.

\begin{lemma}[Convergence in \(W_{\mu, \bm{b}}^{1,2}(\cX)\)]
\label{lem: convergence in W^{1,2}}
    For every \(\bm{\gamma} \in \Gamma\), we have 
    \[
    \lim_{R \to \infty} \wtH_{\bm{\gamma}, R, \bm{\lambda}} 
    = H_{\bm{\gamma}, \bm{\lambda}} \quad \textup{in } W_{\mu, \bm{b}}^{1,2}(\cX).
    \]
\end{lemma}

The next lemma shows that an approximate version of Parseval's identity also holds for systems of finitely many of the capped infinite-dimensional Hermite polynomials \(\wtH_{\bm{\gamma}, R, \bm{\lambda}}\). The proof is given in Appendix~\ref{subsec: app: proof Riesz basis}.

\begin{lemma}[Riesz basis]
\label{lem: Riesz basis}
    Let \(I\subset\Gamma\) be finite. Then, for every \(\varepsilon>0\), there exists \(\bar{R} > 0\) such that for every \(R\geq \bar{R}\), we have
    \begin{equation}
    \label{eq: Riesz basis condition}
        (1-\varepsilon) \nm{\bm{x}}_{\ell^2(I)}^2 
        \leq \nm{\sum_{\bm{\gamma} \in I} x_{\bm{\gamma}} \wtH_{\bm{\gamma}, R, \bm{\lambda}}}_{L_{\mu}^2(\cX)}^2 \leq (1 + \varepsilon) \nm{\bm{x}}_{\ell^2(I)}^2,
        \quad \forall \bm{x} = (x_{\bm{\gamma}})_{\bm{\gamma} \in I} \in \bbR^I.
    \end{equation}
    In particular, \(\{\wtH_{\bm{\gamma}, R, \bm{\lambda}}\}_{\bm{\gamma} \in I}\) is a Riesz basis of \(\spn\{\wtH_{\bm{\gamma}, R, \bm{\lambda}}: \bm{\gamma} \in I\}\) for every \(R \geq \bar{R}\) with Riesz constants at worst \(1 \pm \varepsilon\).
\end{lemma}

We are now ready to reduce the adaptive \(m\)-width to a suitable Gelfand \(m\)-width.  For a sequence \( \bm{u} = (u_{\bm{\gamma}})_{\bm{\gamma} \in \Gamma} \) and an index set \( I \subset \Gamma \), we henceforth denote by \( \bm{u}_I \) the subsequence~\( (u_{\bm{\gamma}})_{\bm{\gamma} \in I} \).

\begin{proof}[Proof of Lemma~\ref{lem: m-width: discrete problem}]
    Let \(\cL: \cV \to \cY^m\) be an adaptive sampling operator as in Definition~\ref{def: adaptive sampling operator; Hilbert-valued}. Then there exist \(Y, \wtY \in \cY \setminus \{0\}\) and a normed vector space \(\widetilde{\cV} \subset L_{\mu}^2(\cX)\) such that, if \(Y F \in \cV\) for some \(F \in L_{\mu}^2(\cX)\), then \(F \in \widetilde{\cV}\) and \(\cL(Y F) = \wtY \widetilde{\cL}(F)\), where \(\widetilde{\cL}: \widetilde{\cV} \to \bbR^m\) is a scalar-valued adaptive sampling operator as in Definition~\ref{def: adaptive sampling operator; scalar-valued}.
    Next, let \(I \subset \Gamma\) be a finite subset and let us fix \(c \in (0, 1)\) and \(\varepsilon > 0\). By Lemma~\ref{lem: convergence in W^{1,2}} and Lemma~\ref{lem: Riesz basis}, there exists \(R > 0\) sufficiently large such that~\eqref{eq: Riesz basis condition} holds and 
    \begin{equation}\label{eq: wtH_gamma - H_gamma}
        \nm{\wtH_{\bm{\gamma}, R, \bm{\lambda}} - H_{\bm{\gamma}, \bm{\lambda}}}_{W_{\mu, \bm{b}}^{1,2}(\cX)} \leq \frac{1-c}{c} \abs{I}^{-1/2}, \quad \forall \bm{\gamma} \in I.
    \end{equation}
    We fix some arbitrary sequence \(\bm{x} = (x_{\bm{\gamma}})_{\bm{\gamma} \in I} \in \bbR^I\) with \(\nm{\bm{x}}_{\ell_{\bm{u}_I}^2(I)} \leq c\) and define \(F_R, F \in W_{\mu, \bm{b}}^{1,2}(\cX; \cY)\) by 
    \begin{equation}
    \label{eq: F_R}
        F_R := \frac{Y}{\nm{Y}_{\cY}} \sum_{\bm{\gamma} \in I} x_{\bm{\gamma}} \wtH_{\bm{\gamma}, R, \bm{\lambda}} 
        \quad \textup{and} \quad
        F := \frac{Y}{\nm{Y}_{\cY}} \sum_{\bm{\gamma} \in I} x_{\bm{\gamma}} H_{\bm{\gamma}, \bm{\lambda}}.
    \end{equation}
    Note that \(\nm{F}_{W_{\mu, \bm{b}}^{1,2}(\cX; \cY)} = \nm{\bm{x}}_{\ell_{\bm{u}_I}^2(I)} \leq c\) by Theorem~\ref{thm: l2-characterization of Gaussian Sobolev space}. Therefore, by the triangle inequality and~\eqref{eq: wtH_gamma - H_gamma}, we have
    \[
    \nm{F_R - F}_{W_{\mu, \bm{b}}^{1,2}(\cX; \cY)}
    \leq \sum_{\bm{\gamma} \in I} \abs{x_{\bm{\gamma}}} \nm{\wtH_{\bm{\gamma}, R, \bm{\lambda}} - H_{\bm{\gamma},\bm{\lambda}}}_{W_{\mu, \bm{b}}^{1,2}(\cX)} 
    \leq \frac{1-c}{c} \abs{I}^{-1/2}  \sum_{\bm{\gamma} \in I} \abs{x_{\bm{\gamma}}} 
    \leq 1-c.
    \]
    The last step follows from Hölder's inequality and \( \nm{\bm{x}}_{\ell^2(I)} \leq \nm{\bm{x}}_{\ell_{\bm{u}_I}^2(I)} \).
    Thus, by the triangle inequality, we have \(\nm{F_R}_{W_{\mu, \bm{b}}^{1,2}(\cX; \cY)} \leq 1\). By Lemma~\ref{lem: Lipschitz continuity of capped Hermite polynomials} and since \(I\) is finite, the operator \(F_R\) is Lipschitz continuous and we conclude \(F_R \in B_{\mu, \bm{b}}^{\Lip}(\cX; \cY)\).

    By Lemma~\ref{lem: Riesz basis}, the family of functionals \(\{\wtH_{\bm{\gamma}, R, \bm{\lambda}}\}_{\bm{\gamma} \in I}\) is a Riesz basis of \(\cF := \spn\{\wtH_{\bm{\gamma}, R, \bm{\lambda}} : \bm{\gamma} \in I\}\). Hence, there exists a unique biorthogonal dual basis \(\{\whH_{\bm{\gamma}, R, \bm{\lambda}}\}_{\bm{\gamma} \in I}\) such that \(\ipd{\wtH_{\bm{\gamma}, R, \bm{\lambda}}}{\whH_{\bm{\gamma'}, R, \bm{\lambda}}}_{L^2_{\mu}(\cX)} = \delta_{\bm{\gamma}, \bm{\gamma'}}\) for every \(\bm{\gamma}, \bm{\gamma'} \in I\). Here, \( \delta_{\bm{\gamma}, \bm{\gamma'}} = \prod_{i = 1}^{\infty} \delta_{\gamma_i, \gamma'_i}\) is the product of Kronecker deltas. The orthogonal projection onto \(\cF\) is given by
    \[
    P_{\cF}: L_{\mu}^2(\cX) \to \cF, \quad 
    P_{\cF} g := \sum_{\bm{\gamma} \in I} \ip{g}{\whH_{\bm{\gamma}, R, \bm{\lambda}}}_{L_{\mu}^2(\cX)} \wtH_{\bm{\gamma}, R, \bm{\lambda}}.
    \]
    Let \(\{\psi_j\}_{j = 1}^{\infty}\) be an orthonormal basis of \(\cY\). For \(G \in L_{\mu}^2(\cX; \cY)\), we write \(g_j := \ip{G}{\psi_j}_{\cY} \in L_{\mu}^2(\cX)\) and find by~\eqref{eq: Riesz basis condition} that
    \begin{align}
    \begin{split}\label{eq: Riesz estimate}
        \nm{G}_{L_{\mu}^2(\cX; \cY)}^2 
        = \sum_{j = 1}^{\infty} \nm{g_j}_{L_{\mu}^2(\cX)}^2
        \geq \sum_{j = 1}^{\infty} \nm{P_{\cF} g_j}_{L_{\mu}^2(\cX)}^2 
        &\geq \sum_{j = 1}^{\infty} (1-\varepsilon) \sum_{\bm{\gamma} \in I} \abs{\ip{g_j}{\whH_{\bm{\gamma}, R, \bm{\lambda}}}_{L_{\mu}^2(\cX)}}^2 \\
        &= (1-\varepsilon) \sum_{\bm{\gamma} \in I} \nm{\int_{\cX} G \whH_{\bm{\gamma}, R, \bm{\lambda}} \D \mu}_{\cY}^2.
    \end{split}
    \end{align}
    We now define the scalar-valued adaptive sampling operator 
    \[
    \Xi_R: \bbR^I \to \bbR^m, \quad \Xi_R(\bm{z}) 
    := \widetilde{\cL}\left(\frac{1}{\nm{Y}_{\cY}} \sum_{\bm{\gamma} \in I} z_{\bm{\gamma}} \wtH_{\bm{\gamma}, R, \bm{\lambda}}\right).
    \]
    We need to show that it is well-defined, that is, \(\nm{Y}_{\cY}^{-1} \sum_{\bm{\gamma} \in I} z_{\bm{\gamma}} \wtH_{\bm{\gamma}, R, \bm{\lambda}} \in \widetilde{\cV}\) for every \(\bm{z} \in \bbR^I\). Since \(B_{\mu, \bm{b}}^{\Lip}(\cX; \cY) \subset \cV\) and \(\cV\) is a vector space, it suffices to observe that \(Y \nm{Y}_{\cY}^{-1} \sum_{\bm{\gamma} \in I} z_{\bm{\gamma}} \wtH_{\bm{\gamma}, R, \bm{\lambda}}\) is Lipschitz continuous as an operator from \(\cX\) to \(\cY\) and it therefore lies in \(\cV\).  
    Next, let \(\cT: \cY^m \to L_{\mu}^2(\cX; \cY)\) be an arbitrary reconstruction mapping. We define \(\widetilde{\cT}: \bbR^m \to L_{\mu}^2(\cX; \cY)\) by 
    \[
    \widetilde{\cT}: \bbR^m \to L_{\mu}^2(\cX; \cY), \quad 
    \widetilde{\cT}(\bm{z}) := \cT(\wtY \bm{z}),
    \]
    and observe that
    \(
    \cT(\cL(F_R)) = \cT(\widetilde{Y} \Xi_R(\bm{x})) = \widetilde{\cT}(\Xi_R(\bm{x})).
    \)
    We now set \(G := F_R - \cT(\cL(F_R))\) in~\eqref{eq: Riesz estimate}. We use the estimate \(\nm{Z}_{\cY} \geq \nm{Y}_{\cY}^{-1} \absd{\ipd{Z}{Y}_{\cY}}\), which holds for every \(Z \in \cY\) by the Cauchy-Schwarz inequality, and compute
    \begin{align*}
        \nm{F_R - \cT(\cL(F_R))}_{L_{\mu}^2(\cX; \cY)}^2 
        &\geq (1-\varepsilon) \sum_{\bm{\gamma} \in I} \nm{\int_{\cX} \left(F_R - \cT(\cL(F_R))\right) \whH_{\bm{\gamma}, R, \bm{\lambda}} \D \mu}_{\cY}^2  \\
        &= (1-\varepsilon) \sum_{\bm{\gamma} \in I} \nm{x_{\bm{\gamma}} \frac{Y}{\nm{Y}_{\cY}} - \int_{\cX} \widetilde{\cT}(\Xi_R(\bm{x})) \whH_{\bm{\gamma}, R, \bm{\lambda}} \D \mu}_{\cY}^2 \\
        &\geq (1-\varepsilon) \sum_{\bm{\gamma} \in I}
        \nm{Y}_{\cY}^{-2} \abs{\ip{x_{\bm{\gamma}} \frac{Y}{\nm{Y}_{\cY}} - \int_{\cX} \widetilde{\cT}(\Xi_R(\bm{x})) \whH_{\bm{\gamma}, R, \bm{\lambda}} \D \mu}{Y}_{\cY}}^2 \\
        &= (1-\varepsilon) \sum_{\bm{\gamma} \in I}
        \abs{x_{\bm{\gamma}} - \frac{1}{\nm{Y}_{\cY}} \int_{\cX} \ip{\widetilde{\cT}(\Xi_R(\bm{x}))}{Y}_{\cY} \whH_{\bm{\gamma}, R, \bm{\lambda}} \D \mu}^2.
    \end{align*}
    Finally, we define the (scalar-valued) reconstruction mapping
    \[
    \Delta_R: \bbR^m \to \bbR^I, \quad 
    \Delta_R(\bm{z}) := \left(\frac{1}{\nm{Y}_{\cY}} \int_{\cX} \ip{\widetilde{\cT}(\bm{z})}{Y}_{\cY} \whH_{\bm{\gamma}, R, \bm{\lambda}} \D \mu \right)_{\bm{\gamma} \in I},
    \]
    and conclude that
    \[
    \nm{F_R - \cT(\cL(F_R))}_{L_{\mu}^2(\cX; \cY)}^2 \geq (1-\varepsilon) \nm{\bm{x} - \Delta_R(\Xi_R(\bm{x}))}_{\ell^2(I)}^2.
    \]
We have thus shown that for any pair \((\cL, \cT)\) of a Hilbert-valued adaptive sampling operator and a reconstruction mapping, the error \(\nm{F_R - \cT(\cL(F_R))}_{L_{\mu}^2(\cX; \cY)}\) can be bounded from below by \((1-\varepsilon)^{1/2} \nm{\bm{x} - \Delta_R(\Xi_R(\bm{x}))}_{\ell^2(I)}\) for some pair \((\Xi_R, \Delta_R)\) of a scalar-valued adaptive sampling operator and reconstruction mapping. Recall that \(F_R \in B_{\mu, \bm{b}}^{\Lip}(\cX; \cY)\) and \(\bm{x} \in \bbR^I\) with \(\nm{\bm{x}}_{\ell_{\bm{u}_I}^2(I)} \leq c\) were arbitrary. Consequently,
    \begin{align*}
        &\Theta_m(B_{\mu, \bm{b}}^{\Lip}(\cX; \cY); \cV, L_{\mu}^2(\cX; \cY)) \\
        &= \inf\left\{
        \sup_{F \in B_{\mu, \bm{b}}^{\Lip}(\cX; \cY)} \nm{F - \cT(\cL(F))}_{L_{\mu}^2(\cX; \cY)}: \cL: \cV \to \cY^m \textup{ adaptive, } \cT: \cY^m \to L_{\mu}^2(\cX; \cY)\right\} \\
        &\geq (1-\varepsilon)^{1/2} \inf\left\{
        \sup_{\substack{\bm{x} \in \bbR^I \\ \nm{\bm{x}}_{\ell_{\bm{u}_I}^2(I)} \leq c}} \nm{\bm{x} - \Delta(\Xi(\bm{x}))}_{\ell^2(I)}: \Xi: \bbR^I \to \bbR^m \textup{ adaptive, } \Delta: \bbR^m \to \bbR^I \right\} \\
        &= (1 - \varepsilon)^{1/2} E_{\textup{ada}}^m(c B_{\bm{u}_I}^2(I), \ell^2(I)) = (1 - \varepsilon)^{1/2} E_{\textup{ada}}^m(B_{(c\bm{u})_I}^2(I), \ell^2(I)),
    \end{align*}
    where in the last step we used that \(c B_{\bm{u}_I}^2(I) = B_{(c\bm{u})_I}^2(I)\).
    As \(\varepsilon > 0\) was arbitrary, we can take the limit \(\varepsilon\to 0^+\).
    The claim now follows by Theorem~\ref{thm: Gelfand width <= adaptive compressive width}.
\end{proof}

We can now finally prove the desired lower bound for the adaptive \(m\)-width.

\begin{proof}[Proof of~\eqref{eq: m-width: lower bound II}]
    Let \(N \in \bbN\) with \(N > m\),  \(I = \pi([N]) = \{\bm{\pi(1)}, \dots, \bm{\pi(N)}\} \subset \Gamma\) be the index set of the \(N\) largest entries of \(\bm{u}\), and fix \(c \in (0,1)\).
    By Theorem~\ref{thm: Gelfand width dual Kolmogorov width} and Lemma~\ref{lem: Kolmogorov width: equality}, we have
    \begin{align}
        \begin{split}\label{eq: lower bound: d^m relation}
        d^m(B_{(c \bm{u})_I}^2(I) , \ell^2(I)) 
        &= d_m(B^2(I), \ell_{1/(c\bm{u})_I}^2(I)) \\
        &= d_m(B_{(c \bm{u})_I}^2(I), \ell^2(I)) 
        = c \cdot d_m(B_{\bm{u}_I}^2(I), \ell^2(I)), 
        \end{split}
    \end{align}
    where the last equality again follows from \(B_{(c\bm{u})_I}^2(I) = c B_{\bm{u}_I}^2(I)\).
    For every \(p > 2\) and \(r = r(p) := 1/2 - 1/p\), H\"older's inequality implies \(N^{-r} B_{\bm{u}_I}^p(I) \subset B_{\bm{u}_I}^2(I)\). Consequently,
    \begin{equation}\label{eq: lower bound: d_m relation}
        d_m(B_{\bm{u}_I}^2(I), \ell^2(I)) \geq d_m(N^{-r} B_{\bm{u}_I}^p(I), \ell^2(I)) 
        = N^{-r} d_m(B_{\bm{u}_I}^p(I), \ell^2(I)).
    \end{equation}
    Applying Theorem~\ref{thm: Stesin} with \(q=2\) yields
    \[
    d_m(B_{\bm{u}_I}^p(I), \ell^2(I)) =  \left(\max_{\substack{i_1,\dots,i_{N-m}\in I \\ i_k \neq i_j}} \left(\sum_{j=1}^{N-m} u_{i_j}^{\frac{2p}{p-2}}\right)^{\frac{1}{p} - \frac{1}{2}}\right)^{-1}.
    \]
    Since \((u_{\bm{\pi(i)}})_{i=1}^{\infty}\) is nonincreasing, it follows with \(q = q(p) := \frac{2p}{p-2} \in (2,\infty)\) that
    \begin{align*}
        d_m(B_{\bm{u}_I}^p(I), \ell^2(I)) 
        = \min_{\substack{i_1,\dots,i_{N-m}\in I \\ i_k \neq i_j}} \left(\sum_{j=1}^{N-m} u_{i_j}^{\frac{2p}{p-2}}\right)^{\frac{1}{2} - \frac{1}{p}}
        &= \left(\sum_{j=m+1}^{N} u_{\bm{\pi(j)}}^{q}\right)^{1/q} 
        \geq u_{\bm{\pi(m+1)}}.
    \end{align*}
    We combine this estimate with~\eqref{eq: lower bound: d^m relation},~\eqref{eq: lower bound: d_m relation}, and Lemma~\ref{lem: m-width: discrete problem}, and conclude
    \[
    u_{\bm{\pi(m+1)}} \leq c^{-1} N^{r} \Theta_m(B_{\mu, \bm{b}}^{\Lip}(\cX; \cY); \cV, L_{\mu}^2(\cX; \cY)).
    \]
    Taking the limit \(p\to 2^+\) yields \(r\to 0^+\) and therefore
    \[
    u_{\bm{\pi(m+1)}} \leq c^{-1} \Theta_m(B_{\mu, \bm{b}}^{\Lip}(\cX; \cY); \cV, L_{\mu}^2(\cX; \cY)).
    \]
    As \(c\in (0,1)\) was arbitrary, we can take the limit \(c \to 1^-\), and the claim finally follows.
\end{proof}

\subsubsection{Upper bound}
\label{subsubsec: m-width: upper bound}

Let \(\cK \in \{ B_{\mu, \bm{b}}(\cX; \cY), B_{\mu, \bm{b}}^{\Lip}(\cX; \cY) \}\). We now assume that \(\cV\) is continuously embedded in \(L_{\mu}^2(\cX; \cY)\) and prove the upper bound for the adaptive \(m\)-width, namely,
\begin{align}
    \begin{split}\label{eq: m-width: upper bound}
        \Theta_m(\cK; \cV, L_{\mu}^2(\cX; \cY)) 
        &\leq \inf_{S \subset \Gamma, \abs{S} \leq m} \sup_{F \in \cK} \nm{F - F_S}_{L_{\mu}^2(\cX; \cY)} \\
        &\leq \sup_{F \in \cK} \nm{F - F_{\{\bm{\pi(1)}, \dots, \bm{\pi(m)}\}}}_{L_{\mu}^2(\cX; \cY)}
        \leq u_{\bm{\pi(m+1)}}.
    \end{split}
\end{align}

\begin{proof}[Proof of~\eqref{eq: m-width: upper bound}]
    The second and third inequality hold by~\eqref{eq: s-term: approximation error <= u_pi(s+1)}, so we only need to prove the first inequality. 
    We fix \(m \in \bbN\) and \(S = \{\bm{\gamma}^{\bm{(1)}}, \dots, \bm{\gamma}^{\bm{(n)}}\} \subset \Gamma\) with \(n \leq m\). We define the Hilbert-valued adaptive sampling operator
    \[
    \cL: \cV \to \cY^m, \quad 
    \cL_i(F) := 
    \begin{cases}
        \int_{\cX} F H_{\bm{\gamma}^{\bm{(i)}}, \bm{\lambda}} \D \mu & \textup{if } 1 \leq i \leq n, \\
        0 & \textup{if } n + 1 \leq i \leq m,
    \end{cases}
    \]
    and the reconstruction mapping
    \[
    \cT: \cY^m \to L_{\mu}^2(\cX; \cY), \quad \cT(\bm{Y}) := \sum_{i=1}^n Y_i H_{\bm{\gamma}^{\bm{(i)}}, \bm{\lambda}}.
    \]
    Since \(\cV\) is continuously embedded in \(L_{\mu}^2(\cX; \cY)\), it is easy to see that \(\cL\) is a well-defined bounded linear operator. We need to show that it satisfies the conditions in Definition~\ref{def: adaptive sampling operator; Hilbert-valued}. It suffices to show that there exists \(Y \in \cY \setminus \{0\}\), a normed vector space \(\widetilde{\cV} \subset L_{\mu}^2(\cX)\), and a scalar-valued adaptive sampling operator \(\widetilde{\cL}: \widetilde{\cV} \to \bbR^m\) such that, if \(YF \in \cV\) for some \(F \in L_{\mu}^2(\cX)\), then \(F \in \widetilde{\cV}\) and \(\cL(Y F) = Y \widetilde{\cL}(F)\). 
    To this end, we choose some \(Y \in \cY\) with \(\nm{Y}_{\cY} = 1\) and define the space
    \(
    \widetilde{\cV} := \left\{F \in L_{\mu}^2(\cX): Y F \in \cV \right\}.
    \)
    It can be readily checked that this defines a normed vector space with norm given by \(\nm{F}_{\widetilde{\cV}} := \nm{Y F}_{\cV}\) for any \(F \in \widetilde{\cV}\). Moreover, as \(\cV\) is continuously embedded in \(L_{\mu}^2(\cX; \cY)\), there exists a constant \(C > 0\) such that
    \[
    \nm{F}_{L_{\mu}^2(\cX)} 
    = \nm{Y F}_{L_{\mu}^2(\cX; \cY)} 
    \leq C \nm{Y F}_{\cV} 
    = C \nm{F}_{\widetilde{\cV}}, \quad \forall F \in \widetilde{\cV},
    \]
    where in the first step we used the fact that \(\nm{Y}_{\cY} = 1\). This shows that \(\widetilde{\cV}\) is continuously embedded in \(L_{\mu}^2(\cX)\). We now define the operator
    \[
    \widetilde{\cL} : \widetilde{\cV} \to \bbR^m, \quad 
    \widetilde{\cL}_i(F) := 
    \begin{cases}
        \int_{\cX} F H_{\bm{\gamma}^{\bm{(i)}}, \bm{\lambda}} \D \mu & \textup{if } 1 \leq i \leq n, \\
        0 & \textup{if } n + 1 \leq i \leq m.
    \end{cases}
    \]
    Note that \(\widetilde{\cL}\) is linear and by the continuous embedding of \(\widetilde{\cV}\) in \(L_{\mu}^2(\cX)\), it is also bounded. Hence, \(\widetilde{\cL}\) is a scalar-valued adaptive sampling operator. Moreover, by construction, if \(Y F \in \cV\), then \(F \in \widetilde{\cV}\) and \(\cL(Y F) = Y \widetilde{\cL}(F)\). We conclude that \(\cL\) is indeed an adaptive Hilbert-valued sampling operator as in Definition~\ref{def: adaptive sampling operator; Hilbert-valued}.
    Consequently,
    \[
    \Theta_m(\cK; \cV, L_{\mu}^2(\cX; \cY)) 
    \leq \sup_{F \in \cK} \nm{F - \cT(\cL(F))}_{L_{\mu}^2(\cX; \cY)}
    = \sup_{F \in \cK} \nm{F - F_S}_{L_{\mu}^2(\cX; \cY)}.
    \]
    As \(S\) was arbitrary, we now take the infimum over all subsets \(S \subset \Gamma\) with \(\abs{S} \leq m\).
\end{proof}

\subsection{Discussion}
\label{subsec: m-width: discussion}

Theorem~\ref{thm: adaptive m-width} shows that linear Hermite polynomial approximation based on the index set \(S = \{\bm{\pi(1)}, \dots, \bm{\pi(m)}\}\) is optimal for the uniform approximation of \(W_{\mu, \bm{b}}^{1,2}\)- and Lipschitz operators with Sobolev norm at most one among all possible recovery strategies which are based on (adaptive) linear information.
Moreover, Theorem~\ref{thm: adaptive m-width} in combination with Theorem~\ref{thm: s-term: lower bound for u_pi(s+1)} yields the following \textbf{curse of sample complexity}: \emph{No strategy based on finitely many (potentially adaptively chosen) linear samples for the uniform recovery of all operators in the Sobolev unit (Lipschitz) ball can achieve algebraic convergence rates. This holds regardless of the decay rate of the PCA eigenvalues of the covariance operator of the underlying Gaussian measure.}

As already mentioned in Section~\ref{subsec: motivation and literature review}, a related~\emph{curse of data complexity} was shown in~\citet{kovachki_DataComplexityEstimatesOperator_2024} by means of the sampling nonlinear \(m\)-width \( s_m(\cK)_{L_{\mu}^2(\cX)} \), defined in~\eqref{eq: sampling nonlinear width}. Let \( B_{\mu, \bm{b}}^{C^{0,1}}(\cX; \cY) = B_{\mu, \bm{b}}^{\Lip}(\cX; \cY) \cap C^{0,1}(\cX; \cY) \) be the Sobolev unit \(C^{0,1}\)-ball. 
Theorem 2.12 in~\citet{kovachki_DataComplexityEstimatesOperator_2024} implies the following result.

\begin{theorem}[Curse of data complexity]
\label{thm: curse of data complexity}
    Let \(\mu\) be a centered Gaussian measure with at most algebraically decreasing (unweighted) PCA eigenvalues \(\lambda_i \gtrsim i^{-\alpha}\) for some \(\alpha > 0\). Then there exists a constant \(C = C(\alpha) > 0\) such that
    \[
    s_m(B_{\mu, \bm{1}}^{C^{0,1}}(\cX))_{L_{\mu}^2(\cX)} \geq C \log(m + 1)^{-(\alpha + 3)}, \quad \forall m \in \bbN.
    \]
\end{theorem}

Our findings in the present section generalize this result in several directions. Notice that the operator \(F_R\) in~\eqref{eq: F_R}, which we use to prove the lower bound of the adaptive \(m\)-width, is not only Lipschitz continuous, but in fact also bounded. As such, it lies in \( B_{\mu, \bm{b}}^{C^{0,1}}(\cX; \cY) \). Following the proof of Lemma~\ref{lem: m-width: discrete problem}, we deduce that \( \Theta_m(B_{\mu, \bm{b}}^{C^{0,1}}(\cX; \cY); \cV, L_{\mu}^2(\cX; \cY)) \geq u_{\bm{\pi(m + 1)}} \) for any vector subspace \( \cV \subset L^2_{\mu}(\cX; \cY) \) which contains \( B_{\mu, \bm{b}}^{C^{0,1}}(\cX; \cY) \). As already noted in Section~\ref{subsec: adaptive widths and main result}, by setting \( \cY = \bbR \), \( \cV = C_b(\cX) \), and \( \bm{b} = \bm{1} \), we can thus conclude that
\[
s_m(B_{\mu, \bm{1}}^{C^{0,1}}(\cX))_{L_{\mu}^2(\cX)} \geq \Theta_m(B_{\mu, \bm{1}}^{C^{0,1}}(\cX); C_b(\cX), L_{\mu}^2(\cX)) \geq u_{\bm{\pi(m + 1)}}.
\]
This bound pertains to \emph{general} centered, nondegenerate Gaussian measures and together with Theorem~\ref{thm: s-term: lower bound for u_pi(s+1)} we conclude that algebraic decays of the sampling \(m\)-width can never be achieved for~\emph{any} decay of the PCA eigenvalues. 
On the other hand, Theorem~\ref{thm: s-term: upper bounds for u_pi(s+1)} provides upper bounds for the adaptive \(m\)-width of the Sobolev unit (Lipschitz) ball in terms of the decay of the PCA eigenvalues \(\lambda_{\bm{b}, i}\). In particular, it implies that the curse of sample complexity described above can be overcome asymptotically in the sense that the adaptive \(m\)-width can decay with rates which are arbitrarily close to any algebraic rate---provided the decay of the \(\lambda_{\bm{b}, i}\) is sufficiently fast, e.g., double exponential.

%%%%%%%%%%%%%%%%%%%%%%%%%%%%%%%%%%%%%%%%%%%%%%%%%%%%%%%%%%%%%%%%%%%%%%%%%%%%%%%%%%%%%%%%%%
%%%%%%%%%%%%%%%%%%%%%%%%%%%%%%%%%%%%%%%%%%%%%%%%%%%%%%%%%%%%%%%%%%%%%%%%%%%%%%%%%%%%%%%%%%

\section{Conclusions}

In this article, we analyzed the approximation of Hilbert-valued Lipschitz operators from finite data. We first extended results from infinite-dimensional analysis and showed that all Lipschitz operators lie in a (weighted) Gaussian Sobolev space \(W_{\mu, \bm{b}}^{1, 2}(\cX; \cY)\).
We then studied Hermite polynomial \(s\)-term approximations and proved that they cannot achieve algebraic convergence rates. This~\emph{curse of parametric complexity} is independent of the decay of the (weighted) PCA eigenvalues \(\lambda_{\bm{b}, i}\) of the covariance operator of the Gaussian measure \(\mu\). 

Next, we analyzed the smallest worst-case error in reconstructing Lipschitz and \(W_{\mu, \bm{b}}^{1, 2}\)-operators from \(m\) arbitrary (potentially adaptively chosen) linear samples in terms of the corresponding adaptive \(m\)-width. 
We extended the IBC principle that ``adaptivity does not help'' in reducing the approximation error from scalar- to \(\cY\)-valued samples and showed that adaptivity does not even help to improve any constants in the error bounds. Furthermore, we proved that from an IBC point of view there is no difference between the space of all Lipschitz operators, equipped with the \(W_{\mu, \bm{b}}^{1, 2}\)-norm, and the space of all \(W_{\mu, \bm{b}}^{1, 2}\)-operators: The adaptive \(m\)-widths in both cases coincide. We tightly characterized the dependence of the adaptive \(m\)-width on the \(\lambda_{\bm{b}, i}\) and identified a~\emph{curse of sample complexity}: No recovery strategy based on finite (adaptive) linear information can achieve algebraic convergence rates uniformly for all Lipschitz and \(W_{\mu, \bm{b}}^{1, 2}\)-operators (of norm at most one). This holds for any (centered, nondegenerate) Gaussian measure independently of its spectral properties. 

It is an active area of research to identify classes of operators for which efficient learning in the sense of achieving algebraic (or faster) convergence rates is possible. As discussed in Section~\ref{subsec: motivation and literature review}, examples include holomorphic operators and solution operators of certain PDEs. On the positive side, we proved that \(W_{\mu, \bm{b}}^{1, 2}\)-regularity, and Lipschitz regularity in particular, suffices to achieve approximation rates which are arbitrarily close to any algebraic rate, provided that the PCA eigenvalues \(\lambda_{\bm{b}, i}\) decay sufficiently fast. As discussed in Section~\ref{subsec: limitations and future work}, there are several avenues for future work, including the construction of practical algorithms that achieve these rates and the investigation into classes that better describe operators of interest in applications.

%%%%%%%%%%%%%%%%%%%%%%%%%%%%%%%%%%%%%%%%%%%%%%%%%%%%%%%%%%%%%%%%%%%%%%%%%%%%%%%%%%%%%%%%%%
%%%%%%%%%%%%%%%%%%%%%%%%%%%%%%%%%%%%%%%%%%%%%%%%%%%%%%%%%%%%%%%%%%%%%%%%%%%%%%%%%%%%%%%%%%

\section*{Acknowledgments}

A part of this work was done when GM was visiting the Department of Mathematics at Simon Fraser University, which he thanks for providing a fruitful research environment and working conditions. He also thanks BA for his invitation and hospitality as well as the Hausdorff School for Mathematics (HSM) in Bonn for financial support. The authors would also like to thank Mario Ullrich for useful discussions.

\section*{Funding}
BA was supported by the \emph{Natural Sciences and Engineering Research Council of Canada} (NSERC) through grant RGPIN-2021-611675. MG and GM were supported by the \emph{Hausdorff Center for Mathematics} (HCM) in Bonn, funded by the Deutsche Forschungsgemeinschaft (DFG, German Research Foundation) under Germany’s Excellence Strategy -- EXC-2047/1 -- 390685813, and by the CRC 1060 \emph{The Mathematics of Emergent Effects} -- 211504053 of the Deutsche Forschungsgemeinschaft.

%%%%%%%%%%%%%%%%%%%%%%%%%%%%%%%%%%%%%%%%%%%%%%%%%%%%%%%%%%%%%%%%%%%%%%%%%%%%%%%%%%%%%%%%%%
%%%%%%%%%%%%%%%%%%%%%%%%%%%%%%%%%%%%%%%%%%%%%%%%%%%%%%%%%%%%%%%%%%%%%%%%%%%%%%%%%%%%%%%%%%

\bibliographystyle{abbrvnat}
\bibliography{SampleComplexityLipschitz-R1}

%%%%%%%%%%%%%%%%%%%%%%%%%%%%%%%%%%%%%%%%%%%%%%%%%%%%%%%%%%%%%%%%%%%%%%%%%%%%%%%%%%%%%%%%%%
%%%%%%%%%%%%%%%%%%%%%%%%%%%%%%%%%%%%%%%%%%%%%%%%%%%%%%%%%%%%%%%%%%%%%%%%%%%%%%%%%%%%%%%%%%

\appendix
\renewcommand\theHsection{appendix.\Alph{section}}

\section{Notions of differentiability}
\label{sec: app: Notions of differentiability}

We recall several notions of differentiability which we use throughout the paper, loosely following~\citet[Ch. 5.1]{bogachev_GaussianMeasures_1998}. The constructions in this section hold for general Banach spaces \(\cX\) and \(\cY\). As usual, we denote by \(L(\cX; \cY)\) the space of all bounded linear operators \(F : \cX \to \cY\) with finite operator norm \(\nm{F}_{L(\cX; \cY)} := \sup_{X \in \cX, X \neq 0} \nm{F(X)}_{\cY} / \nm{X}_{\cX}\).

\begin{definition}[Differentiability]
    Let \(\cM\) be a collection of non-empty subsets of \(\cX\), \(X \in \cX\). A mapping \(F: \cX \to \cY\) is said to be differentiable with respect to \(\cM\) at the point \(X\) if there exists a continuous linear mapping, denoted by \(D^{\cM} F(X) \in L(\cX; \cY)\), such that for every fixed set \(M\in\cM\), we have 
    \[
    \lim_{t \to 0} \sup_{Z \in M} \nm{\frac{F(X + tZ) - F(X)}{t} - D^{\cM} F(X)(Z)}_{\cY} = 0.
    \]
\end{definition}

If \(\cM\) is the class of all \emph{finite}, \emph{compact}, or \emph{bounded} subsets of \(\cX\), then we say that \(F\) is \emph{G\^ateaux}, \emph{Hadamard}, or \emph{Fréchet differentiable at \(X\)}, respectively. In the latter case, we also often just call \(F\)~\emph{differentiable} at \(X\). Note that in all three cases, the mapping \( D^{\cM} F(X) \) is unique and it is called the \emph{(G\^ateaux, Hadamard, or Fréchet) derivative of \(F\) at \(X\)}. If \(\cX\) is finite-dimensional, it is easy to see that Hadamard and Fréchet differentiability at \(X\) are equivalent and the corresponding derivatives of \(F\) at \(X\) coincide. This will become important in the proof of Theorem~\ref{thm: Lipschitz operators are Gaussian Sobolev operators} in Appendix~\ref{sec: app: Lipschitz operators are Gaussian Sobolev operators}. We henceforth drop the superscript \(\cM\) in the notation of the derivative and write \(DF(X)\) instead of \(D^{\cM} F(X)\). In the following, we will focus on Fréchet differentiability, unless stated otherwise. We call \(F\) \emph{Fréchet differentiable} (or just~\emph{differentiable}) if it is Fréchet differentiable at every point \(X \in \cX\). The resulting derivative \(D F\) is a mapping from \(\cX\) to \(L(\cX; \cY)\). 

If \(\cE\) is a linear subspace of \(\cX\) (possibly with a stronger norm), then we say that \(F\) is \emph{differentiable along \(\cE\) at the point \(X\)} if the mapping \( \cE \to \cY \), \(Z \mapsto F(X + Z)\) is differentiable at \(Z = 0\). 
We call \(F\) \emph{differentiable along \(\cE\)} if it is differentiable along \(\cE\) at every point \(X \in \cX\). The resulting derivative \(D_{\cE} F\) is a mapping from \(\cX\) to \(L(\cE; \cY)\). Moreover, the differential operator \(D_{\cE}\), mapping \(F\) to its derivative \(D_{\cE} F\), is linear in \(F\).

\begin{example}
    If \(\cE = \cX\), then \(D_{\cE} F = D F\). If \(F\) is Fréchet differentiable and we choose \(\cE\) to be the Cameron-Martin space \(\cH\) of a Gaussian measure on \(\cX\) (see Appendix~\ref{subsec: app: the Cameron-Martin space}), then \(D_{\cE} F\) is the~\emph{\(\cH\)-derivative} of \(F\) which is commonly used in infinite-dimensional analysis and Malliavin calculus~\citep[see][]{bogachev_GaussianMeasures_1998,daprato_IntroductionStochasticAnalysisMalliavin_2014,lunardi_InfiniteDimensionalAnalysis_2015}.
\end{example}

If \(\cE = \spn\{Z\}\), \(Z \in \cX \setminus \{0\}\), is one-dimensional, we obtain the usual directional derivative
\[
\frac{\partial}{\partial Z} F(X) 
:= D_{\spn\{Z\}} F(X)(Z)
=  \lim_{t \to 0} \frac{F(X + t Z) - F(X)}{t} \in \cY.
\]
In this case, the G\^ateaux, Hadamard, and Fréchet derivatives along \(\cE\) at \(X\) coincide.
For any subspaces \(\cE' \subset \cE \subset \cX\), it can be readily seen that if \(F: \cX \to \cY\) is differentiable along \(\cE\) at a point \(X \in \cX\), then \(F\) is also differentiable along \(\cE'\) at \(X\) and 
\(
D_{\cE} F(X) |_{\cE'} = D_{\cE'} F(X).
\)
In particular, if \(F\) is differentiable along \(\cE\) at \(X\), then, for any \(Z \in \cE \setminus \{0\}\), the directional derivative \(\frac{\partial}{\partial Z} F(X)\) at \(X\) exists and
\[
\frac{\partial}{\partial Z} F(X) = D_{\cE} F(X)(Z).
\]
If \(\cX = \bbR^n\) for some \( n \in \bbN \) and \(Z = \bm{e_i}\) is the \(i\)th standard unit vector, we use the standard notation \(\partial_i F(\bm{x}) := \frac{\partial}{\partial \bm{e_i}} F(\bm{x})\) for \(\bm{x} = (x_1, \dots, x_n) \in \bbR^n\).
If \(\cE\) is a Hilbert subspace of \(\cX\) and if \(F: \cX \to \bbR\) is differentiable along \(\cE\) at \(X\), then, by the Riesz representation theorem, there exists a unique \(Z \in \cE\) such that \(D_{\cE} F(X)(X') = \ip{Z}{X'}_{\cE}\) for every \(X' \in \cE\). In this case, we call \(Z\) the \emph{\(\cE\)-gradient of \(F\) at \(X\)} and write
\(
\nabla_{\cE} F(X) := Z.
\)

\section{Closability and closure of operators}
\label{sec: app: closability and closure of operators}

We define closability and the closure of operators between Hilbert spaces and state standard properties. For further details we refer to~\citet[Ch. 12]{berezanskij_FunctionalAnalysisVol2_1996}.
Let \(\cH_1, \cH_2\) be two Hilbert spaces. A linear \(\cH_2\)-valued operator (not necessarily bounded) acting on \(\cH_1\) is a linear mapping \(A: \textup{dom}(A) \to \cH_2\) from a linear subspace \(\textup{dom}(A) \subset \cH_1\) to \(\cH_2\). The set \(\textup{dom}(A)\) is called the \emph{domain} of \(A\). 
The \emph{graph} of \(A\) is defined as the set
\[
\Gamma_A := \left\{(H, AH) \in \cH_1 \oplus \cH_2: H \in \textup{dom}(A)\right\}.
\]
We can equip \( \textup{dom(A)} \) with the \emph{graph inner product},
\[
\ip{H}{K}_{\Gamma_A} := \ip{H}{K}_{\cH_1} + \ip{AH}{AK}_{\cH_2}, \quad H,K\in \textup{dom}(A),
\]
which turns it into a pre-Hilbert space with \emph{graph norm}
\[
\nm{(H, AH)}_{\Gamma_A} 
:= \sqrt{\ip{H}{H}_{\Gamma_A}}
= \left(\nm{H}_{\cH_1}^2 + \nm{AH}_{\cH_2}^2\right)^{1/2}.
\]

\begin{definition}[Closability and closure]
    A linear operator \(A: \textup{dom}(A) \to \cH_2\) is called \emph{closable} (in \(\cH_1\)) if the closure of its graph \(\overline{\Gamma}_A\) in \(\cH_1 \oplus \cH_2\) is the graph of some (necessarily unique) linear operator, that is, there exists a linear operator \(\overline{A} : \textup{dom}(\overline{A}) \to \cH_2\) such that \(\overline{\Gamma}_A = \Gamma_{\overline{A}}\). In this case, we call \(\overline{A}\) the \emph{closure} of \(A\).
\end{definition}

If \(A\) is closable, then the domain of its closure is given by 
\[
\textup{dom}(\overline{A}) = \left\{H \in \cH_1: \exists (H_i)_{i = 1}^{\infty} \subset \textup{dom}(A):
\lim_{i \to \infty} H_i = H, 
\ (A H_i)_{i = 1}^{\infty} \textup{ converges in } \cH_2\right\}.
\]
For \( H \in \textup{dom}(\overline{A}) \) and any sequence \( (H_i)_{i = 1}^{\infty} \subset \textup{dom}(A) \) with \( \lim_{i \to \infty} H_i = H \) in \( \cH_1 \) and \( (A H_i)_{i = 1}^{\infty} \) being convergent in \( \cH_2 \), we have \( \lim_{i \to \infty} A H_i = \overline{A} H\) in \( \cH_2 \). In other words, closability of \( A \) guarantees that the limit \( \lim_{i \to \infty} A H_i \) is independent of the sequence \( (H_i)_{i = 1}^{\infty} \), see~\citet[Thm. 2.1]{berezanskij_FunctionalAnalysisVol2_1996}.
If \(A\) is closable, we equip \(\textup{dom}(\overline{A})\) with the graph inner product, which turns \((\textup{dom}(\overline{A}), \ip{\cdot}{\cdot}_{\Gamma_{\overline{A}}})\) into a Hilbert space.

\section{Results from infinite-dimensional analysis}
\label{sec: app: results from infinite-dimensional analysis}

We recall some well-known results from infinite-dimensional analysis which are used to define the Gaussian Sobolev space \(W_{\mu, \bm{b}}^{1,2}(\cX; \cY)\) (see Definition~\ref{def: Gaussian Sobolev space}) and to prove Theorem~\ref{thm: Lipschitz operators are Gaussian Sobolev operators} in Appendix~\ref{sec: app: Lipschitz operators are Gaussian Sobolev operators}. 
We first define the Cameron-Martin space \(\cH\) of \(\mu\) in \( \cX \) in Appendix~\ref{subsec: app: the Cameron-Martin space} and then discuss the construction of \(W_{\mu, \bm{b}}^{1,2}(\cX; \cY)\) as well as some of its important properties in Appendix~\ref{subsec: app: the Gaussian Sobolev space}.
We mainly follow~\citet{lunardi_InfiniteDimensionalAnalysis_2015} and~\citet{daprato_IntroductionInfiniteDimensionalAnalysis_2006}, who consider \(\cY = \bbR\) and the cases \(\bm{b} = \sqrt{\bm{\lambda}}\) and \(\bm{b} = \bm{1}\), respectively, and generalize the proofs therein to general Hilbert-valued operators in the case \(\bm{b} \in \ell^{\infty}(\bbN)\), \(\bm{b} > \bm{0}\). More information can also be found in~\citet{bogachev_GaussianMeasures_1998}. Throughout this section, we use notation as introduced in the main text.

\subsection{The Cameron-Martin space}
\label{subsec: app: the Cameron-Martin space}

\begin{theorem}[{Fernique's theorem~\citep[Thm. 2.8.5]{bogachev_GaussianMeasures_1998}}]
\label{thm: Fernique theorem}
    There exists \(\alpha > 0\) such~that
    \[
    \int_{\cX} \exp(\alpha \nm{X}_{\cX}^2) \D \mu(X) < \infty.
    \]
\end{theorem}
Fernique's theorem implies that any mapping \(\cX \to \cY\) which grows at most polynomially at infinity belongs to \(L_{\mu}^2(\cX; \cY)\). In particular, the inclusion mapping
\[
j: \cX^* \to L_{\mu}^2(\cX), \quad F \mapsto F,
\]
is well-defined. We use it to define the Cameron-Martin space:

\begin{definition}[Cameron-Martin space]
    The \emph{Cameron-Martin space of \(\mu\) (in \(\cX\))} is the set \(\cH\) of all \(X \in \cX\) whose \(\cH\)-norm is finite, where
    \[
    \nm{X}_{\cH} := \sup\left\{F(X): F\in \cX^*, \nm{j(F)}_{L_{\mu}^2(\cX)} \leq 1 \right\}.
    \]
\end{definition}

To further describe the structure of \(\cH\), we define the \emph{reproducing kernel Hilbert space \(\cX_{\mu}^*\) of \(\mu\)} as the closure of \(j(\cX^*)\) in \(L_{\mu}^2(\cX)\), equipped with the inner product from \( L^2_{\mu}(\cX) \),
as well as the mapping
\[
R_{\mu}: \cX_{\mu}^* \to \cX, \;\; F \mapsto \int_\cX X F(X) \D \mu(X),
\]
where the integral is in the sense of Bochner. Notice that \(R_{\mu}\) is well-defined by Fernique's~theorem.

\begin{proposition}[{Relation between \(\cH\) and \(\cX_{\mu}^*\)~\citep[Prop. 3.1.2]{lunardi_InfiniteDimensionalAnalysis_2015}}]
    An element \(H \in \cX\) belongs to \(\cH\) if and only if there exists \(\hat{H} \in \cX_{\mu}^*\) such that \(H = R_{\mu}(\hat{H})\). In this case, we have
    \(
    \nm{H}_{\cH} = \nmd{\hat{H}}_{L_{\mu}^2(\cX)}.
    \)
    Hence, \(R_{\mu}: \cX_{\mu}^* \to \cH\) is an isometric isomorphism that turns \(\cH\) into a Hilbert space with inner product 
    \(
    \ip{H}{K}_{\cH} := \ipd{\hat{H}}{\hat{K}}_{L_{\mu}^2(\cX)}
    \)
    whenever \(H = R_{\mu}(\hat{H})\) and~\(K = R_{\mu}(\hat{K})\).
\end{proposition}

In our case, where \(\cX\) is a separable Hilbert space, the Cameron-Martin space has a particularly simple structure. Indeed, it follows from~\citet[Thm. 4.2.7]{lunardi_InfiniteDimensionalAnalysis_2015} that the family of vectors \(\{\xi_i\}_{i=1}^{\infty}\) with 
\begin{equation}\label{eq: orthonormal basis of cH}
    \xi_i := \sqrt{\lambda_i} \phi_i, \quad i \in \bbN,
\end{equation}
is an orthonormal basis of \(\cH\). The corresponding elements \(\hat{\xi}_i \in \cX_{\mu}^*\) are given by
\begin{equation}
\label{eq: form of hat(xi_i)}
    \hat{\xi}_i(\cdot) = \lambda_i^{-1/2} \ip{\cdot}{\phi_i}_{\cX}, \quad i \in \bbN.
\end{equation}
The right-hand side in~\eqref{eq: form of hat(xi_i)} is an element of \(\cX^*\) and we henceforth identify each \(\hat{\xi}_i\) with its version in \(\cX^*\).

\subsection[The Gaussian Sobolev space]{The Gaussian Sobolev space \(W_{\mu, \bm{b}}^{1,2}(\cX; \cY)\)}
\label{subsec: app: the Gaussian Sobolev space}

Recall the weighted space \(\cX_{\bm{b}}\) with weight sequence \(\bm{b} = (b_i)_{i=1}^{\infty} \in \ell^{\infty}(\bbN)\), \( \bm{b} > \bm{0}\), and orthonormal basis \(\{\eta_i\}_{i=1}^{\infty}\), \(\eta_i := b_i \phi_i\), as defined in~\eqref{eq: weighted space cX_b} and~\eqref{eq: orthonormal basis of cX_b}, respectively.

\subsubsection{Construction}

The construction of Gaussian Sobolev spaces is based on \emph{cylindrical functionals}:

\begin{definition}[Cylindrical functionals and operators]
    A functional \(\varphi: \cX \to \bbR\) is called a \emph{cylindrical functional} if there exist \(n\in\bbN\), \(\ell_1, \dots,\ell_n \in \cX^*\), and a function \(\omega: \bbR^n \to \bbR\) such that
    \[
    \varphi(X) = \omega(\ell_1(X), \dots, \ell_n(X)), \quad \forall X \in \cX.
    \]
    We call \(\varphi\) a \emph{cylindrical boundedly Fréchet differentiable functional} if, with the above notation, \(\omega \in C_b^1(\bbR^n)\). The space of all such functionals is denoted by \(\cF C_b^1(\cX)\).
    Moreover, we define the set of all \emph{cylindrical boundedly Fréchet differentiable \(\cY\)-valued operators} by 
    \[
    \cF C_b^1(\cX; \cY) := \spn\left\{\cX \ni X \mapsto \varphi(X) Y \in \cY: \varphi \in \cF C_b^1(\cX), Y \in \cY\right\},
    \]
    so that every \(F \in \cF C_b^1(\cX; \cY)\) can be written as
    \(
    F = \sum_{i = 1}^n \varphi_i Y_i
    \)
    for some \(\varphi_i \in \cF C_b^1(\cX)\), \(Y_i \in \cY\), and \(n \in \bbN\).
\end{definition}

\begin{proposition}[Closability of \(D_{\cX_{\bm{b}}}\)]
\label{prop: closability of differential operator}
    The (Fréchet) differential operator along \(\cX_{\bm{b}}\), \(D_{\cX_{\bm{b}}}: \cF C_b^1(\cX;\cY) \to L_{\mu}^2(\cX; \HS(\cX_{\bm{b}}; \cY))\), is closable in \(L_{\mu}^2(\cX; \cY)\). 
\end{proposition}

\begin{proof}
    First, a short computation shows that for every \(F \in \cF C_b^1(\cX; \cY)\), the derivative \(D_{\cX_{\bm{b}}} F\) lies in \(L_{\mu}^2(\cX; \HS(\cX_{\bm{b}}; \cY))\). Hence, the mapping \(D_{\cX_{\bm{b}}}: \cF C_b^1(\cX;\cY) \to L_{\mu}^2(\cX; \HS(\cX_{\bm{b}}; \cY))\) is well-defined.
    The proof of closability is a straightforward modification of the proof of~\citet[Lem. 10.2.4]{lunardi_InfiniteDimensionalAnalysis_2015}, replacing the derivative along the Cameron-Martin space of \( \mu \) by the derivative along the space \(\cX_{\bm{b}}\).
\end{proof}

\subsubsection{Properties}

We first observe that the Hermite polynomials \( \{ H_{\bm{\gamma}, \bm{\lambda}} \}_{\bm{\gamma} \in \Gamma} \) belong to \( W_{\mu, \bm{b}}^{1, 2}(\cX; \cY) \). This is not immediate because \( H_{\bm{\gamma}, \bm{\lambda}} \) is cylindrical and smooth but not bounded. For this, we require some notation. For \( \bm{\gamma} \in \Gamma \) and \( i \in \bbN \), we define \( \bm{\gamma}^{\bm{(i)}} = (\gamma_k^{(i)})_{k = 1}^{\infty} \in \Gamma \) as follows: If \( \gamma_i = 0 \), we set  \( \bm{\gamma}^{\bm{(i)}} := \bm{0} \), and if \( \gamma_i > 0 \), we set
\[
\gamma_k^{(i)} := 
\begin{cases}
    \gamma_k - 1 & \textup{if } k = i, \\
    \gamma_k & \textup{if } k \neq i.
\end{cases}
\]

\begin{lemma}[Hermite polynomials are Gaussian Sobolev functionals]
\label{lem: Hermite polynomials in W12_mu}
    For every \( \bm{\gamma} \in \Gamma \), we have \( H_{\bm{\gamma}, \bm{\lambda}} \in W^{1,2}_{\mu, \bm{b}}(\cX) \) with weak Gaussian derivatives \( \frac{\partial}{\partial \eta_i} H_{\bm{\gamma}, \bm{\lambda}} = b_i \sqrt{\frac{\gamma_i}{\lambda_i}} H_{\bm{\gamma^{(i)}}, \bm{\lambda}} \) for every \( i \in \bbN \).
\end{lemma}

\begin{proof}
    Let \( d \in \bbN \) such that \( \supp(\bm{\gamma}) \subset [d] \). Let \( \omega \in C_c^{\infty}(\bbR^d) \) be a smooth bump function with \( 0 \leq \omega \leq 1 \), \( \omega = 1 \) on \( [-1,1]^d \), and \( \nm{\nabla \omega}_{\infty} \leq C \) for some constant \( C > 0 \). We define for \( n \in \bbN \) and \( X \in \cX \),
    \[
    \Upsilon_n(X) := \omega \left( \frac{\ip{X}{\phi_1}_{\cX}}{n}, \dots, \frac{\ip{X}{\phi_d}_{\cX}}{n} \right) \quad \textup{and} \quad \varphi_n(X) := H_{\bm{\gamma}, \bm{\lambda}}(X) \Upsilon_n(X).
    \]
    Notice that \( \varphi_n \in \cF C_b^1(\cX) \) with \( \abs{\varphi_n} \leq \absd{H_{\bm{\gamma}, \bm{\lambda}}} \). Also, since \( \lim_{n \to \infty} \Upsilon_n = 1 \) pointwise, we have \( \lim_{n \to \infty} \varphi_n = H_{\bm{\gamma}, \bm{\lambda}} \) pointwise. Since \( H_{\bm{\gamma}, \bm{\lambda}} \) is a polynomial, Fernique's theorem (Theorem~\ref{thm: Fernique theorem}) implies that \( H_{\bm{\gamma}, \bm{\lambda}} \in L^2_{\mu}(\cX) \). By the dominated convergence theorem (DCT), we conclude that \( \lim_{n \to \infty} \varphi_n = H_{\bm{\gamma}, \bm{\lambda}} \) in \( L^2_{\mu}(\cX) \).
    
    Now consider the derivatives. By~\eqref{eq: orthonormal basis of cX_b}, we have \( \frac{\partial}{\partial \eta_i}  = b_i \frac{\partial}{\partial \phi_i} \). Hence, by the product rule,
    \[
    \frac{\partial}{\partial \eta_i} \varphi_n(X) = 
    b_i \frac{\partial}{\partial \phi_i} H_{\bm{\gamma}, \bm{\lambda}}(X) \Upsilon_n(X) + b_i H_{\bm{\gamma}, \bm{\lambda}}(X) \frac{\partial}{\partial \phi_i} \Upsilon_n(X), \quad \forall X \in \cX, \forall i \in \bbN.
    \]
    Again, since \( \frac{\partial}{\partial \phi_i} H_{\bm{\gamma}, \bm{\lambda}} \) is a polynomial, it lies in \( L^2_{\mu}(\cX) \). In fact, by equation (10.16) in~\cite{daprato_IntroductionInfiniteDimensionalAnalysis_2006}, we have
    \[
    \frac{\partial}{\partial \phi_i} H_{\bm{\gamma}, \bm{\lambda}} = \sqrt{\frac{\gamma_i}{\lambda_i}} H_{\bm{\gamma^{(i)}}, \bm{\lambda}}.
    \]
    Since \( \abs{\Upsilon_n} \leq 1 \), it follows from the DCT that \( \lim_{n \to \infty} b_i \frac{\partial}{\partial \phi_i} H_{\bm{\gamma}, \bm{\lambda}} \Upsilon_n = b_i \sqrt{\frac{\gamma_i}{\lambda_i}} H_{\bm{\gamma^{(i)}}, \bm{\lambda}} \) in \( L^2_{\mu}(\cX) \).
    Further, for \( i \in [d] \), notice that \( \frac{\partial}{\partial \phi_i} \Upsilon_n(X) = \frac{1}{n} \partial_i \omega \left( \frac{\ip{X}{\phi_1}_{\cX}}{n}, \dots, \frac{\ip{X}{\phi_d}_{\cX}}{n} \right) \) and hence, \( \absd{\frac{\partial}{\partial \phi_i} \Upsilon_n} \leq C / n \). Again by the DCT, we find that \( \lim_{n \to \infty} b_i H_{\bm{\gamma}, \bm{\lambda}} \frac{\partial}{\partial \phi_i} \Upsilon_n = 0 \) in \( L^2_{\mu}(\cX) \) for every \( i \in [d] \).
    For \(i > d \), we have \( \frac{\partial}{\partial \eta_i} \varphi_n = 0 \). We conclude that for every \( i \in \bbN \), we have
    \begin{equation}
    \label{eq: Gaussian derivatives of Hermite polynomials}
        \lim_{n \to \infty} \frac{\partial}{\partial \eta_i} \varphi_n = b_i \sqrt{\frac{\gamma_i}{\lambda_i}} H_{\bm{\gamma^{(i)}}, \bm{\lambda}} \quad \textup{ in } L^2_{\mu}(\cX).
    \end{equation}
    
    Altogether, \( \lim_{n \to \infty} \varphi_n =  H_{\bm{\gamma}, \bm{\lambda}} \) in \( L^2_{\mu}(\cX) \) and \( (\nabla_{\cX_{\bm{b}}} \varphi_n)_{n = 1}^{\infty} \) is convergent in \( L^2_{\mu}(\cX; \cX_{\bm{b}}) \). This implies by definition that \( H_{\bm{\gamma}, \bm{\lambda}} \in W^{1,2}_{\mu, \bm{b}}(\cX) \) and it follows from~\eqref{eq: Gaussian derivatives of Hermite polynomials} that the weak Gaussian derivatives are given by \( \frac{\partial}{\partial \eta_i} H_{\bm{\gamma}, \bm{\lambda}} = b_i \sqrt{\frac{\gamma_i}{\lambda_i}} H_{\bm{\gamma^{(i)}}, \bm{\lambda}} \).
\end{proof}

We proceed with an important criterion for an operator to belong to \( W_{\mu, \bm{b}}^{1, 2}(\cX; \cY)\):

\begin{lemma}
\label{lem: belonging to W^1,2}
    Suppose that \((F_n)_{n = 1}^{\infty} \subset W_{\mu, \bm{b}}^{1,2}(\cX; \cY)\) with \(\lim_{n \to \infty} F_n = F\) in \(L_{\mu}^2(\cX; \cY)\).
    \begin{itemize}
        \item [(i)] If \(\sup_{n\in\bbN} \nm{D_{\cX_{\bm{b}}} F_n}_{L^2_{\mu}(\cX; \HS(\cX_{\bm{b}}; \cY))} < \infty\), then \(F \in W_{\mu, \bm{b}}^{1,2}(\cX; \cY)\). 
        \item [(ii)] If \( \lim_{n \to \infty} D_{\cX_{\bm{b}}} F_n = G \) in \( L^2_{\mu}(\cX; \HS(\cX_{\bm{b}}; \cY)) \), then \(F \in W_{\mu, \bm{b}}^{1,2}(\cX; \cY)\) with \( D_{\cX_{\bm{b}}} F = G \).
    \end{itemize}
\end{lemma}

\begin{proof}
    Since \(W_{\mu, \bm{b}}^{1,2}(\cX; \cY)\) is a Hilbert space, it is reflexive. Notice that in both cases the assumptions imply that \((F_n)_{n = 1}^{\infty}\) is bounded in \(W_{\mu, \bm{b}}^{1,2}(\cX; \cY)\). Hence, there exists a subsequence \((F_{n_k})_{k = 1}^{\infty}\) which converges weakly in \(W_{\mu, \bm{b}}^{1,2}(\cX; \cY)\) to some \(\wtF\) as \(k \to \infty\). Since \( \lim_{k \to \infty} F_{n_k} = F \) in \(L_{\mu}^2(\cX; \cY)\) by assumption, we conclude that \(F = \wtF\). This shows the claim in case (i). In case (ii), the assumptions imply additionally that \( D_{\cX_{\bm{b}}} \wtF = G \).
\end{proof}

Next, we study the \(\ell^2\)-characterization of \(W_{\mu, \bm{b}}^{1, 2}(\cX; \cY)\), see Theorem~\ref{thm: l2-characterization of Gaussian Sobolev space}.

\begin{proposition}
\label{prop: l2-characterization of Gaussian Sobolev space}
    Let \(F \in W_{\mu, \bm{b}}^{1,2}(\cX; \cY)\). Then,
    \begin{equation}
    \label{eq: series expansion of partial derivative}
        \frac{\partial}{\partial \eta_i} F = \sum_{\bm{\gamma} \in \Gamma} b_i \sqrt{\frac{\gamma_i}{\lambda_i}} \left( \int_{\cX} F H_{\bm{\gamma}, \bm{\lambda}} \D \mu \right) H_{\bm{\gamma}^{\bm{(i)}}, \bm{\lambda}},
        \quad \forall i \in \bbN,
    \end{equation}
    and 
    \begin{equation}
    \label{eq: W^1,2 - l^2 isometry}
        \nm{F}_{W_{\mu, \bm{b}}^{1,2}(\cX; \cY)}^2
        = \sum_{\bm{\gamma} \in \Gamma} \left(1 + \sum_{i = 1}^{\infty} b_i^2 \frac{\gamma_i}{\lambda_i}\right) \nm{\int_{\cX} F H_{\bm{\gamma}, \bm{\lambda}} \D \mu}_{\cY}^2.
    \end{equation}
    Conversely, if for a family of vectors \((Y_{\bm{\gamma}})_{\bm{\gamma} \in \Gamma} \subset \cY\) one has
    \begin{equation}
    \label{eq: l2 summability of weighted coefficients}
        \sum_{\bm{\gamma} \in \Gamma} \left(1 + \sum_{i = 1}^{\infty} b_i^2 \frac{\gamma_i}{\lambda_i}\right) \nm{Y_{\bm{\gamma}}}_{\cY}^2 < \infty,
    \end{equation}
    then 
    \(
    F := \sum_{\bm{\gamma} \in \Gamma} Y_{\bm{\gamma}} H_{\bm{\gamma}, \bm{\lambda}}\) lies in \(W_{\mu, \bm{b}}^{1,2}(\cX; \cY)\).
\end{proposition}

\begin{proof}
    First, let us fix a cylindrical functional \(\varphi \in \cF C_b^1(\cX)\). By~\citet[Lemma 10.14]{daprato_IntroductionInfiniteDimensionalAnalysis_2006}, the partial derivatives of \(\varphi\) satisfy
    \begin{equation}
    \label{eq: identity for partial derivative; smooth functions}
        \frac{\partial}{\partial \phi_i} \varphi = \sum_{\bm{\gamma} \in \Gamma} \sqrt{\frac{\gamma_i}{\lambda_i}} \left( \int_{\cX} \varphi H_{\bm{\gamma}, \bm{\lambda}} \D \mu \right) H_{\bm{\gamma}^{\bm{(i)}}, \bm{\lambda}},
        \quad \forall i \in \bbN.
    \end{equation}
    Since \(\frac{\partial}{\partial \eta_i} \varphi = b_i \frac{\partial}{\partial \phi_i} \varphi\), it follows from~\eqref{eq: identity for partial derivative; smooth functions} together with orthonormality of the Hermite polynomials that
    \begin{equation}
    \label{eq: coefficient identity for partial derivative; smooth functions}
        \int_{\cX} \left(\frac{\partial}{\partial \eta_i} \varphi\right) H_{\bm{\gamma}^{\bm{(i)}}, \bm{\lambda}}
        \D \mu = b_i \sqrt{\frac{\gamma_i}{\lambda_i}} \int_{\cX} \varphi H_{\bm{\gamma}, \bm{\lambda}} \D \mu, \quad \forall i \in \bbN.
    \end{equation}
    By linearity, the identities~\eqref{eq: identity for partial derivative; smooth functions} and~\eqref{eq: coefficient identity for partial derivative; smooth functions} hold in fact for every cylindrical operator \(\varphi \in \cF C_b^1(\cX; \cY)\).
    Now suppose that \(F \in W_{\mu, \bm{b}}^{1,2}(\cX; \cY)\). It suffices to prove~\eqref{eq: series expansion of partial derivative}, as~\eqref{eq: W^1,2 - l^2 isometry} then follows by Parseval's identity~\eqref{eq: Parseval's identity}. For this, it is enough to show that
    \begin{equation}
    \label{eq: coefficient identity for partial derivative}
        \int_{\cX} \left(\frac{\partial}{\partial \eta_i} F\right) H_{\bm{\gamma}^{\bm{(i)}}, \bm{\lambda}}
        \D \mu = b_i \sqrt{\frac{\gamma_i}{\lambda_i}} \int_{\cX} F H_{\bm{\gamma}, \bm{\lambda}} \D \mu, \quad \forall i \in \bbN.
    \end{equation}
    By definition of \(W_{\mu, \bm{b}}^{1,2}(\cX; \cY)\), there exists a sequence \((F_n)_{n = 1}^{\infty} \subset \cF C_b^1(\cX; \cY)\) such that \(\lim_{n \to \infty} F_n = F\) and \(\lim_{n \to \infty} \frac{\partial}{\partial \eta_i} F_n = \frac{\partial}{\partial\eta_i} F\) in \(L_{\mu}^2(\cX; \cY)\). We set \(\varphi = F_n\) in~\eqref{eq: coefficient identity for partial derivative; smooth functions} and take the limit \(n \to \infty\) to obtain~\eqref{eq: coefficient identity for partial derivative}.

    Conversely, suppose that~\eqref{eq: l2 summability of weighted coefficients} holds for a family of vectors \( (Y_{\bm{\gamma}})_{\bm{\gamma} \in \Gamma} \subset \cY \).
    We fix an enumeration \( \tau: \bbN \to \Gamma \) of \( \Gamma \) and define 
    \[
    F_n := \sum_{j = 1}^n Y_{\bm{\tau(j)}} H_{\bm{\tau(j)}, \bm{\lambda}}, \quad \forall n \in \bbN.
    \]
    We deduce from Parseval's identity and~\eqref{eq: l2 summability of weighted coefficients} that \( (F_n)_{n = 1}^{\infty} \) is a Cauchy sequence in \( L^2_{\mu}(\cX; \cY) \) with limit \( F := \sum_{i = 1}^{\infty} Y_{\bm{\tau(i)}} H_{\bm{\tau(i)}, \bm{\lambda}} \). Notice that the limit is independent of the enumeration \( \tau \). We can thus write \( F = \sum_{\bm{\gamma} \in \Gamma} Y_{\bm{\gamma}} H_{\bm{\gamma}, \bm{\lambda}} \).
    Lemma~\ref{lem: Hermite polynomials in W12_mu} implies that \( F_n \in W^{1,2}_{\mu, \bm{b}}(\cX; \cY) \) with 
    \[
    \frac{\partial}{\partial \eta_i} F_n = \sum_{j = 1}^n b_i \sqrt{\frac{\tau(j)_i}{\lambda_i}} Y_{\bm{\tau(j)}} H_{\bm{\tau(j)}^{\bm{(i)}}, \bm{\lambda}}, \quad \forall i \in \bbN.
    \]
    Consequently, we can bound the \( L_{\mu}^2(\cX; \HS(\cX_{\bm{b}}; \cY)) \)-norm of the derivatives \( D_{\cX_{\bm{b}}} F_n \) by
    \begin{align*}
        &\int_{\cX} \nm{D_{\cX_{\bm{b}}} F_n(X)}_{\HS(\cX_{\bm{b}};\cY)}^2 \D \mu(X)
        = \int_{\cX} \sum_{i = 1}^{\infty} \nm{\frac{\partial}{\partial \eta_i} F_n(X)}_{\cY}^2 \D \mu(X) \\
        &= \sum_{i = 1}^{\infty} \int_{\cX} \nm{\sum_{j = 1}^n b_i \sqrt{\frac{\tau(j)_i}{\lambda_i}} Y_{\bm{\tau(j)}} H_{\bm{\tau(j)}^{\bm{(i)}}, \bm{\lambda}}(X)}_{\cY}^2 \D \mu(X)
        = \sum_{i = 1}^{\infty} \sum_{j = 1}^n b_i^2 \frac{\tau(j)_i}{\lambda_i} \nm{Y_{\bm{\tau(j)}}}_{\cY}^2 \\
        &\leq \sum_{j = 1}^{\infty} \sum_{i = 1}^{\infty} b_i^2 \frac{\tau(j)_i}{\lambda_i} \nm{Y_{\bm{\tau(j)}}}_{\cY}^2.
    \end{align*}
    The right-hand side is finite by~\eqref{eq: l2 summability of weighted coefficients}. Lemma~\ref{lem: belonging to W^1,2}(i) now implies that \( F \in W_{\mu, \bm{b}}^{1, 2}(\cX; \cY) \).
\end{proof}

Next, we consider operators with special approximation properties that will become important in the proof of Theorem~\ref{thm: Lipschitz operators are Gaussian Sobolev operators} in Appendix~\ref{sec: app: Lipschitz operators are Gaussian Sobolev operators}. To this end, recall from~\eqref{eq: orthonormal basis of cH},~\eqref{eq: form of hat(xi_i)} the orthonormal basis \(\{\xi_i\}_{i=1}^{\infty}\) of \(\cH\) with \(\xi_i = \sqrt{\lambda_i} \phi_i\), \(R_{\mu} (\hat{\xi}_i) = \xi_i\), and \(\hat{\xi}_i \in \cX^*\). For every \(F \in L_{\mu}^2(\cX; \cY)\) and \(n \in \bbN\), we define 
\begin{equation}\label{eq: definition of conditional expectation}
    \bbE_n F : \cX \to \cY, \quad 
    \bbE_n F := \bbE [F \mid \hat{\xi}_1, \dots, \hat{\xi}_n] ,
\end{equation}
to be the conditional expectation of \(F\) with respect to the \(\sigma\)-algebra generated by the random variables \(\hat{\xi}_1,\dots,\hat{\xi}_n\). Furthermore, for \(n \in \bbN\), we define the mapping
\begin{equation}\label{eq: definition of P_n}
    P_n: \cX \to \spn\{\xi_i: i \in [n]\}, \quad 
    P_n X := \sum_{i = 1}^n \hat{\xi}_i(X) \xi_i.
\end{equation}

\begin{proposition}[Properties of \(\bbE_n F\) in \(L_{\mu}^2\)]
\label{prop: properties of conditional expectation in L2}
    For \(F \in L_{\mu}^2(\cX; \cY)\), let \(\bbE_n F\) and \(P_n\) be given as in~\eqref{eq: definition of conditional expectation} and~\eqref{eq: definition of P_n}, respectively. Then the following properties hold true:
    \begin{itemize}
        \item [(i)] For every \(F \in L_{\mu}^2(\cX; \cY)\) and \(n \in \bbN\), we have
        \[
        \bbE_n F(X) = \int_{\cX} F(P_n X + (\textup{Id} - P_n)Z) \D \mu(Z) \quad \text{for } \mu \text{-a.e. } X \in \cX.
        \]
        In particular, \(\bbE_n F\) can be identified with an operator \(V_n\) on \(P_n(\cX)\) by setting \(V_n(Z) := \bbE_n F(X)\) for \(Z = P_n X\).
    
        \item [(ii)] For every \(F \in L_{\mu}^2(\cX; \cY)\), the sequence \((\bbE_n F)_{n = 1}^{\infty}\) converges to \(F\) in \(L_{\mu}^2(\cX; \cY)\).
    \end{itemize}
\end{proposition}

\begin{proof}
    We refer to the proofs of Proposition 7.4.1 and Proposition 7.4.4, respectively, in~\citet{lunardi_InfiniteDimensionalAnalysis_2015}, which can be adopted almost verbatim, only changing the Lebesgue integrals to Bochner integrals.
\end{proof}

\begin{proposition}[Properties of \(\bbE_n F\) in \(W_{\mu, \bm{b}}^{1, 2}\)]
\label{prop: properties of conditional expectation in W12}
    Let \(F \in W_{\mu, \bm{b}}^{1, 2}(\cX; \cY)\) and \(\bbE_n F\) be defined as in~\eqref{eq: definition of conditional expectation}. Then, \(\bbE_n F \in W_{\mu, \bm{b}}^{1, 2}(\cX; \cY)\) for every \(n \in \bbN\) and 
    \(
    \lim_{n \to \infty} \bbE_n F = F
    \)
    in \(W_{\mu, \bm{b}}^{1, 2}(\cX; \cY)\).
\end{proposition}

We need the following preliminary result for cylindrical operators:
\begin{proposition}
\label{prop: properties of conditional expectation in W12; cylindrical functionals}
    Let \( \varphi \in \cF C_b^1(\cX;\cY) \). Then, for every \( n \in \bbN \), we have \( \bbE_n \varphi \in W_{\mu, \bm{b}}^{1, 2}(\cX;\cY) \) and the following properties hold true:
    \begin{itemize}
        \item [(i)] For every \( i \in \bbN \), the \( i \)th partial derivative of \( \bbE_n \varphi \) is given by
        \[
        \frac{\partial}{\partial \eta_i} \bbE_n \varphi = 
        \begin{cases}
            \bbE_n (\frac{\partial}{\partial \eta_i} \varphi) & \textup{if } i \leq n, \\
            0 & \textup{if } i > n.
        \end{cases}
        \]
        \item [(ii)] We have \( \lim_{n \to \infty} \bbE_n \varphi = \varphi  \) in \( W_{\mu, \bm{b}}^{1, 2}(\cX;\cY) \).
    \end{itemize}
\end{proposition}

\begin{proof}
    In the case \( \cY = \bbR \) and \( \bm{b} = \sqrt{\bm{\lambda}} \) (where \( \eta_ i = \xi_i \)), the claim follows from~\citet[Prop. 10.1.2]{lunardi_InfiniteDimensionalAnalysis_2015}. The case of general \( \bm{b} \) can be readily checked by using the identity \( \frac{\partial}{\partial \eta_i} = b_i \lambda_i^{-1/2} \frac{\partial}{\partial \xi_i} \), see~\eqref{eq: orthonormal basis of cX_b} and~\eqref{eq: orthonormal basis of cH}. The lifting to general \( \cY \) is straightforward.
\end{proof}

\begin{proof}[Proof of Proposition~\ref{prop: properties of conditional expectation in W12}]
    Let \( F \in W^{1,2}_{\mu, \bm{b}}(\cX; \cY) \). Let \( (\varphi_m)_{m = 1}^{\infty} \subset \cF C_b^1(\cX; \cY) \) be a sequence of cylindrical operators which converges to \( F \) in \( W^{1,2}_{\mu, \bm{b}}(\cX; \cY) \). Using Proposition~\ref{prop: properties of conditional expectation in L2}(i), it can be readily checked that \( \bbE_n \varphi_m \in \cF C_b^1(\cX; \cY) \). By the contraction property of the conditional expectation, we find
    \[
    \nm{ \bbE_n \varphi_m - \bbE_n F}_{L^2_{\mu}(\cX; \cY)} 
    \leq \nm{ \varphi_m - F}_{L^2_{\mu}(\cX; \cY)}, 
    \]
    and hence, \( \lim_{m \to \infty} \bbE_n \varphi_m = \bbE_n F \) in \( L^2_{\mu}(\cX; \cY) \). 
    Moreover, using Parseval's identity, Proposition~\ref{prop: properties of conditional expectation in W12; cylindrical functionals}(i), and again the contraction property, we compute for any \( \varphi \in \cF C_b^1(\cX; \cY) \)
    \begin{align*}
        \nm{D_{\cX_{\bm{b}}} \bbE_n \varphi}_{L_{\mu}^2(\cX; \HS(\cX_{\bm{b}}; \cY))}^2 
        &= \int_{\cX} \sum_{i = 1}^{\infty} \nm{\frac{\partial}{\partial \eta_i} \bbE_n \varphi(X)}_{\cY}^2 \D \mu(X)
        = \sum_{i = 1}^n \nm{\bbE_n \left(\frac{\partial}{\partial \eta_i} \varphi \right)}_{L_{\mu}^2(\cX; \cY)}^2 \\
        &\leq \sum_{i = 1}^n \nm{\frac{\partial}{\partial \eta_i} \varphi}_{L_{\mu}^2(\cX;\cY)}^2 
        \leq \int_{\cX} \sum_{i = 1}^{\infty} \nm{\frac{\partial}{\partial \eta_i} \varphi(X)}_{\cY}^2 \D \mu(X) \\
        &= \nm{D_{\cX_{\bm{b}}} \varphi}_{L_{\mu}^2(\cX; \HS(\cX_{\bm{b}}; \cY))}^2.
    \end{align*}   
    By setting \( \varphi = \varphi_m - \varphi_{m'} \) with \( m, m' \in \bbN \), we deduce that \( (D_{\cX_{\bm{b}}} \bbE_n \varphi_m)_{m = 1}^{\infty} \) is a Cauchy sequence in \( L_{\mu}^2(\cX; \HS(\cX_{\bm{b}}; \cY)) \). 
    We conclude that \( \bbE_n F \in W^{1,2}_{\mu, \bm{b}}(\cX; \cY) \) with \( \lim_{m \to \infty} \bbE_n \varphi_m = \bbE_n F \) in \( W^{1,2}_{\mu, \bm{b}}(\cX; \cY) \). 
    We know from Proposition~\ref{prop: properties of conditional expectation in L2} that \( \lim_{n \to \infty} \bbE_n F = F \) in \( L^2_{\mu}(\cX; \cY) \). Furthermore,
    \begin{align*}
        \nm{D_{\cX_{\bm{b}}} \bbE_n F - D_{\cX_{\bm{b}}} F}_{L_{\mu}^2(\cX; \HS(\cX_{\bm{b}}; \cY))} 
        &\leq \nm{D_{\cX_{\bm{b}}} \bbE_n F - D_{\cX_{\bm{b}}} \bbE_n \varphi_m}_{L_{\mu}^2(\cX; \HS(\cX_{\bm{b}}; \cY))} \\
        &\quad + \nm{D_{\cX_{\bm{b}}} \bbE_n \varphi_m - D_{\cX_{\bm{b}}} \varphi_m}_{L_{\mu}^2(\cX; \HS(\cX_{\bm{b}}; \cY))} \\
        &\quad + \nm{D_{\cX_{\bm{b}}} \varphi_m - D_{\cX_{\bm{b}}} F}_{L_{\mu}^2(\cX; \HS(\cX_{\bm{b}}; \cY))}.
    \end{align*}
    The first and third term on the right-hand side converge to zero as \( m \to \infty \) by the arguments above. The second term converges to zero as \( n \to \infty \) by Proposition~\ref{prop: properties of conditional expectation in W12; cylindrical functionals}(ii). By a diagonal argument, we conclude that the left-hand side converges to zero as \( n \to \infty \). In conclusion, we have \( \lim_{n \to \infty} \bbE_n F = F \) in \( W_{\mu, \bm{b}}^{1,2}(\cX; \cY) \).
\end{proof}

The final two results identify families of compositional operators in \( W^{1,2}_{\mu, \bm{b}}(\cX; \cY) \) that will be needed in Appendices~\ref{sec: app: Lipschitz operators are Gaussian Sobolev operators} and~\ref{subsec: app: proof convergence}. To this end, we set 
\begin{equation}
\label{eq: T_n}
    T_n : \cX \to \bbR^n, \quad X \mapsto (\hat{\xi}_1(X), \dots, \hat{\xi}_n(X)),
\end{equation}
with the functionals \(\hat{\xi}_i = \lambda_i^{-1/2} \ip{\cdot}{\phi_i}_{\cX} \in \cX^*\), \( i \in \bbN \), defined in~\eqref{eq: form of hat(xi_i)}.
\begin{lemma}
\label{lem: Lebesgue Sobolev to Gaussian Sobolev}
    Let \(F : \cX \to \cY\) be an operator of the form
    \(
    F = G \circ T_n
    \)
    with \(G \in  W^{1,2}(\bbR^n; \cY)\) for some \(n \in \bbN\) and \( T_n \) given by~\eqref{eq: T_n}.
    Then, \(F \in W_{\mu, \bm{b}}^{1,2}(\cX; \cY)\) with
    \[
    \frac{\partial}{\partial \eta_i} F(X) =
    \begin{cases}
        b_i \lambda_i^{-1/2} \partial_i G(T_n(X)) & \textup{ if } 1 \leq i \leq n, \\
        0 & \textup{ otherwise.}
    \end{cases}
    \]
    for \( \mu \)-a.e. \( X \in \cX \).
\end{lemma}

\begin{proof}
    First observe that the pushforward measure \( T_n \sharp \mu \) is given by  \( \bigotimes_{i = 1}^n \cN(0,1) \). It is then easy to see that \( F \in L^2_{\mu}(\cX; \cY) \). Next, we construct a sequence of cylindrical operators that converges to \( F \) in \( L^2_{\mu}(\cX; \cY) \) and which yields the claimed derivative expressions. By~\citet[Cor. 2.5.6]{hytonen_AnalysisBanachSpacesVolume_2016}, there exists a sequence \( (G_m)_{m = 1}^{\infty} \subset C_c^{\infty}(\bbR^n; \cY) \) such that \( \lim_{m \to \infty} G_m = G \) in \( W^{1,2}(\bbR^n; \cY) \). Let \( \{\psi_j\}_{j = 1}^{\infty} \) be an orthonormal basis of \( \cY \) and define \( G_{m,k} := \sum_{j = 1}^k \ip{G_m}{\psi_j}_{\cY} \psi_j \) as well as \( F_{m,k} := G_{m,k} \circ T_n \). By construction, we have \( F_{m,k} \in \cF C_b^1(\cX; \cY) \). Using Parseval's identity, we compute
    \begin{align*}
        \nm{F_{m,k} - F}_{L^2_{\mu}(\cX; \cY)}^2 
        &= \int_{\cX} \sum_{j = 1}^k \abs{\ip{G_m(T_n(X)) - G(T_n(X))}{\psi_j}_{\cY}}^2 \D \mu(X) \\ 
        & \quad + \int_{\cX} \sum_{j = k + 1}^{\infty} \abs{\ip{G(T_n(X))}{\psi_j}_{\cY}}^2 \D \mu(X) \\
        &= \frac{1}{(2 \pi)^{n / 2}} \int_{\bbR^n} \sum_{j = 1}^k \abs{\ip{G_m(\bm{x}) - G(\bm{x})}{\psi_j}_{\cY}}^2 e^{-\nm{\bm{x}}_2^2 / 2} \D \bm{x} \\
        & \quad + \frac{1}{(2 \pi)^{n / 2}} \int_{\bbR^n} \sum_{j = k + 1}^{\infty} \abs{\ip{G(\bm{x})}{\psi_j}_{\cY}}^2 e^{-\nm{\bm{x}}_2^2 / 2} \D \bm{x}.
    \end{align*}
    The first term on the right-hand side converges for fixed \( k \) to zero as \( m \to \infty \) by construction of the \( G_m \). The second term converges to zero as \( k \to \infty \) because \( G \in L^2(\bbR^n; \cY) \). By a diagonal argument, we can thus choose \( m = m(k) \) such that \( \lim_{k \to \infty} F_{m(k), k} = F \) in \( L^2_{\mu}(\cX; \cY) \).
    
    Next, we consider the derivatives. By the chain rule,
    \[
    \frac{\partial}{\partial \eta_i} F_{m,k}(X) = 
    \begin{cases}
        b_i \lambda_i^{-1/2} \partial_i G_{m,k}(T_n(X)) & \textup{ if } 1 \leq i \leq n, \\
        0 & \textup{ otherwise,}
    \end{cases}
    \]
    for every \( X \in \cX \). By a similar computation as above, after possibly adapting \( m = m(k) \), we find that \( \lim_{k \to \infty} (\partial_i G_{m(k),k}) \circ T_n = (\partial_i G) \circ T_n \) in \( L^2_{\mu}(\cX; \cY) \) for \( i \in [n] \), as well as \(  \lim_{k \to \infty} \frac{\partial}{\partial \eta_i} F_{m,k} = 0 \) in \( L^2_{\mu}(\cX; \cY) \) for \( i > n \). This yields the claim. 
\end{proof}

\begin{lemma}
\label{lem: Lipschitz to Gaussian Sobolev}
    Let \(F : \cX \to \cY\) be an operator of the form
    \(
    F = G \circ T_n
    \)
    with \(G \in \Lip(\bbR^n; \cY)\) for some \(n \in \bbN\) and \( T_n \) given by~\eqref{eq: T_n}.
    Then, \(F \in W_{\mu, \bm{b}}^{1,2}(\cX; \cY)\) with
    \[
    \frac{\partial}{\partial \eta_i} F(X) =
    \begin{cases}
        b_i \lambda_i^{-1/2} \partial_i G(T_n(X)) & \textup{ if } 1 \leq i \leq n, \\
        0 & \textup{ otherwise},
    \end{cases}
    \]
    for \(\mu\)-a.e. \(X \in \cX\).
\end{lemma}

\begin{proof}
    Suppose that \( G \) is \( L \)-Lipschitz for some \( L > 0 \). It follows from the generalized Rademacher theorem that \( G \) is Hadamard differentiable \(\cL^n\)-almost everywhere in \(\bbR^n\), see~\mbox{\citet[Thm. 1 in Ch. 2 \& Rmk. 2 in Ch. 1]{aronszajn_DifferentiabilityLipschitzianMappingsBanach_1976}}. Since \(\bbR^n\) is finite-dimensional, \(G\)~is in fact Fréchet differentiable \(\cL^n\)-almost everywhere---recall Appendix~\ref{sec: app: Notions of differentiability}. Moreover, the derivative is bounded by \( \nm{D G(\bm{x})}_{\HS(\bbR^n; \cY)} \leq \sqrt{n} L \) for \( \cL^{n} \)-a.e. \( \bm{x} \in \bbR^n \). Consequently, \( G \in W^{1, \infty}_{\textup{loc}}(\bbR^n; \cY) \).
    
    Next, let \( \omega \in C_c^{\infty}(\bbR^n) \) be a smooth bump function with \( 0 \leq \omega \leq 1 \), \( \omega = 1 \) on \( [-1,1]^n \), and \( \nm{\nabla \omega}_{\infty} \leq C \) for some constant \( C > 0 \). For \( m \in \bbN \), we set \( \omega_m(\bm{x}) = \omega(\bm{x} / m) \) for \( \bm{x} \in \bbR^n \). We further set \(G_m := \omega_m G\) and \( F_m := G_m \circ T_n \). Notice that \( G_m \in W^{1,2}(\bbR^n; \cY) \). Consequently, we find by Lemma~\ref{lem: Lebesgue Sobolev to Gaussian Sobolev} that \( F_m \in W^{1,2}_{\mu, \bm{b}}(\cX; \cY) \) with 
    \[
    \frac{\partial}{\partial \eta_i} F_m(X) =
    \begin{cases}
        b_i \lambda_i^{-1/2} \partial_i G_m(T_n(X)) & \textup{ if } 1 \leq i \leq n, \\
        0 & \textup{ otherwise,}
    \end{cases}
    \]
    for \( \mu \)-a.e. \( X \in \cX \). Using similar arguments as in the proof of Lemma~\ref{lem: Hermite polynomials in W12_mu}, we find that \( \lim_{m \to \infty} F_m = F \) in \( L^2_{\mu}(\cX; \cY) \). In addition, \( \lim_{m \to \infty} \frac{\partial}{\partial \eta_i} F_m = b_i \lambda_i^{-1/2} (\partial_i G) \circ T_n \) for \(i \in [n]\) and \( \lim_{m \to \infty} \frac{\partial}{\partial \eta_i} F_m = 0\) for \( i > n\) in \( L^2_{\mu}(\cX; \cY) \). The claim now follows by Lemma~\ref{lem: belonging to W^1,2}(ii).
\end{proof}

\section{Proof of Theorem~\ref{thm: Lipschitz operators are Gaussian Sobolev operators}}
\label{sec: app: Lipschitz operators are Gaussian Sobolev operators}

The argument follows along the lines of the proof of~\citet[Prop. 10.1.4]{lunardi_InfiniteDimensionalAnalysis_2015} with several important modifications. Let \(F: \cX \to \cY\) be Lipschitz continuous with \(L := [F]_{\Lip(\cX; \cY)}\).
The idea is to use Lemmas~\ref{lem: belonging to W^1,2} and~\ref{lem: Lipschitz to Gaussian Sobolev} to show that \(F \in W_{\mu, \bm{b}}^{1,2}(\cX; \cY)\).

By Lipschitz continuity, we have \(\nm{F(X)}_{\cY} \leq \nm{F(0)}_{\cY} + L \nm{X}_{\cX}\) for every \(X \in \cX\), which implies that \(F \in L_{\mu}^2(\cX; \cY)\) by Fernique's theorem (Theorem~\ref{thm: Fernique theorem}). 
For \( n \in \bbN \), we set \(F_n := \bbE_n F\), as defined in~\eqref{eq: definition of conditional expectation}. By Proposition~\ref{prop: properties of conditional expectation in L2}, we have \(\lim_{n \to \infty} \bbE_n F = F\) in \(L_{\mu}^2(\cX; \cY)\) and we can write
\(
\bbE_n F(X) = V_n(T_n(X))
\)
for some \(V_n: \bbR^n \to \cY\) and \(T_n: \cX \to \bbR^n\), \(T_n(X) := (\hat{\xi}_1(X), \dots, \hat{\xi}_n(X))\) with \(\xi_i, \hat{\xi}_i\) defined in~\eqref{eq: orthonormal basis of cH},~\eqref{eq: form of hat(xi_i)}. Notice that \(V_n\) inherits Lipschitz continuity from \(F\). Indeed, using Proposition~\ref{prop: properties of conditional expectation in L2}(i), we find for \(\bm{x}, \bm{z} \in \bbR^n\)
\begin{align}
        \nm{V_n(\bm{x} + \bm{z}) - V_n(\bm{x})}_{\cY}
        &= \nm{\bbE_n F\left(\sum_{i=1}^n x_i \xi_i + \sum_{i=1}^n z_i \xi_i\right) - \bbE_n F\left(\sum_{i=1}^n x_i \xi_i\right)}_{\cY} \nonumber \\
        &\hspace{-15ex} \leq \int_{\cX} \nm{F\left(\sum_{i=1}^n x_i \xi_i + \sum_{i=1}^n z_i \xi_i + (\textup{Id} - P_n) Z \right) - F\left(\sum_{i=1}^n x_i \xi_i + (\textup{Id} - P_n) Z \right)}_{\cY} \D \mu(Z)  \nonumber \\
        &\hspace{-15ex} \leq L \nm{\sum_{i=1}^n z_i \xi_i}_{\cX}  
        = L \nm{\sum_{i=1}^n \sqrt{\lambda_i} z_i \phi_i}_{\cX}
        = L \left(\sum_{i=1}^n \lambda_i z_i^2\right)^{1/2}. 
        \label{eq: Lipschitz continuity of V_n on Rn_lambda}
\end{align}

Next, write \( m = \dim(\cY) \) and let \(\{\psi_j\}_{j = 1}^m\) be an orthonormal basis of \(\cY\). For \(j \in [m]\), we define the function(al)s
\begin{align*}
    &V_n^{(j)}: \bbR^n \to \bbR, \quad \bm{x} \mapsto \ip{V_n(\bm{x})}{\psi_j}_{\cY} \\
    &\bbE_n^{(j)} F: \cX \to \bbR, \quad X \mapsto \ip{\bbE_n F(X)}{\psi_j}_{\cY} = V_n^{(j)}(T_n(X)).
\end{align*}
We equip \(\bbR^n\) with the rescaled Euclidean inner product
\(
\ip{\bm{x}}{\bm{y}}_{\bbR_{\bm{\lambda}}^n} := \sum_{i=1}^n \lambda_i x_i y_i,
\)
which induces the norm \(\nm{\bm{x}}_{\bbR_{\bm{\lambda}}^n} := \sqrt{\ip{\bm{x}}{\bm{x}}_{\bbR_{\bm{\lambda}}^n}}\). We denote the hereby defined space by \(\bbR_{\bm{\lambda}}^n\) and observe that \(\{\lambda_i^{-1/2} \bm{e_i}\}_{i=1}^n\) is an orthonormal basis. Recall that \( \bm{e_i} \) denotes \(i\)th standard unit vector in \( \bbR^n \).
Now notice that~\eqref{eq: Lipschitz continuity of V_n on Rn_lambda} implies that \(V_n^{(j)}\) is \(L\)-Lipschitz as a function \(\bbR_{\bm{\lambda}}^n \to \bbR\).
We can thus apply Lemma~\ref{lem: Lipschitz to Gaussian Sobolev} with \( \cY = \bbR \) (notice that it is independent of the inner product structure in \( \bbR^n \)). Consequently,
\[
D_{\cX_{\bm{b}}} (\bbE_n^{(j)}F)(X)(\eta_i) = 
\begin{cases}
    b_i \lambda_i^{-1/2} \partial_i V_n^{(j)}(T_n(X))  & \textup{if } 1 \leq i \leq n, \\
    0 & \textup{otherwise,}
\end{cases}
\]
for \( \mu \)-a.e. \(X \in \cX\).

\emph{Proof of (i).} Suppose that \( m \in \bbN \). For \(\cL^n\)-almost every \(\bm{x} \in \bbR^n\), we have
\begin{align*}
    \sum_{i=1}^n \lambda_i^{-1} \absd{\partial_i V_n^{(j)}(\bm{x})}^2
    &= \sum_{i=1}^n \absd{D V_n^{(j)}(\bm{x})(\lambda_i^{-1/2} \bm{e_i})}^2 \\
    &= \nmd{D V_n^{(j)}(\bm{x})}_{\HS(\bbR_{\bm{\lambda}}^n; \bbR)}^2 
    = \nmd{D V_n^{(j)}(\bm{x})}_{L(\bbR_{\bm{\lambda}}^n; \bbR)}^2 
    \leq L^2.
\end{align*}
Consequently, for \(\mu\)-a.e. \( X \in \cX \),
\begin{align}
    &\nm{D_{\cX_{\bm{b}}} (\bbE_n F)(X)}_{\HS(\cX_{\bm{b}};\cY)}^2 
    = \bigg\| \sum_{j = 1}^m D_{\cX_{\bm{b}}} (\bbE_n^{(j)} F)(X) \psi_j \bigg\|_{\HS(\cX_{\bm{b}};\cY)}^2 \nonumber \\
    &= \sum_{i = 1}^{\infty} \bigg\| \sum_{j = 1}^m D_{\cX_{\bm{b}}} (\bbE_n^{(j)} F)(X)(\eta_i) \psi_j \big\|_{\cY}^2
    = \sum_{i = 1}^{\infty} \sum_{j = 1}^m \absd{D_{\cX_{\bm{b}}} (\bbE_n^{(j)} F)(X)(\eta_i)}^2 \nonumber \\
    &= \sum_{i = 1}^n \sum_{j = 1}^m b_i^2 \lambda_i^{-1} \absd{\partial_i V_n^{(j)}(T_n(X))}^2 
    \leq m \nm{\bm{b}}_{\ell^{\infty}(\bbN)}^2 L^2. \label{eq: estimate of derivative of conditional expectation; finite dimension}
\end{align}
By Lemma~\ref{lem: belonging to W^1,2}(i), we may conclude that \( F \in W_{\mu, \bm{b}}^{1,2}(\cX; \cY) \).

Next, we show that \(C^{0, 1}(\cX; \cY)\) is continuously embedded in \(W_{\mu, \bm{b}}^{1,2}(\cX;\cY)\). By~\eqref{eq: estimate of derivative of conditional expectation; finite dimension}, we know that \(\nm{D_{\cX_{\bm{b}}} (\bbE_n F)(X)}_{L_{\mu}^2(\cX; \HS(\cX_{\bm{b}}; \cY))} \leq \sqrt{m} \nm{\bm{b}}_{\ell^{\infty}(\bbN)} L\). Moreover, it follows from Proposition~\ref{prop: properties of conditional expectation in L2}(i) that \(\nm{\bbE_n F}_{L_{\mu}^2(\cX; \cY)} \leq \sup_{X \in \cX} \nm{F(X)}_{\cY}\). In total, we have
\begin{align*}
    \nm{\bbE_n F}_{W_{\mu, \bm{b}}^{1,2}(\cX;\cY)} 
    &\leq \left(\sup_{X \in \cX} \nm{F(X)}_{\cY}^2 + m \nm{\bm{b}}_{\ell^{\infty}(\bbN)}^2 L^2 \right)^{1/2} \\
    &\leq \max\left\{1, \sqrt{m} \nm{\bm{b}}_{\ell^{\infty}(\bbN)}\right\} \cdot \nm{F}_{C^{0, 1}(\cX; \cY)}.
\end{align*}
As \(\lim_{n \to \infty} \bbE_n F = F\) in \(W_{\mu, \bm{b}}^{1,2}(\cX;\cY)\) by Proposition~\ref{prop: properties of conditional expectation in W12}, the claim follows.

\emph{Proof of (ii).}
Suppose that \(m = \infty\). Notice that in this case we cannot use the same argument as in the proof of (i), as~\eqref{eq: estimate of derivative of conditional expectation; finite dimension} does not give a meaningful bound for \(m = \infty\). We thus have to argue differently. To this end, assume that \(\bm{b} \in \ell^2(\bbN)\).
From~\eqref{eq: Lipschitz continuity of V_n on Rn_lambda} it follows that \(\nm{\partial_i V_n(\bm{x})}_{\cY} \leq L \sqrt{\lambda_i}\) for \(\cL^n\)-a.e. \(\bm{x} \in \bbR^n\) and every \(1 \leq i \leq n\).
Consequently, for \( \mu \)-a.e. \( X \in \cX \), we have
\begin{align*}
    \nm{D_{\cX_{\bm{b}}} (\bbE_n F)(X)}_{\HS(\cX_{\bm{b}};\cY)}^2 
    &= \sum_{i = 1}^{\infty} \nm{D_{\cX_{\bm{b}}} (\bbE_n F)(X)(\eta_i)}_{\cY}^2 \\
    &= \sum_{i = 1}^n b_i^2 \lambda_i^{-1} \nm{\partial_i V_n(T_n(X))}_{\cY}^2 
    \leq L^2 \nm{\bm{b}}_{\ell^2(\bbN)}^2.
\end{align*}
By Lemma~\ref{lem: belonging to W^1,2}(i), we conclude that \( F \in W_{\mu, \bm{b}}^{1,2}(\cX; \cY) \). In order to show that \(C^{0, 1}(\cX; \cY)\) is continuously embedded in \(W_{\mu, \bm{b}}^{1,2}(\cX;\cY)\) with the claimed bound, we can argue analogously as in the proof of (i). \hfill \qedsymbol{}

\section{Further proofs}
\label{sec: app: further proofs}

For \( d \in \bbN \cup \{\infty \} \), we write \( \mu_d := \bigotimes_{i = 1}^d \cN(0,1) \). In Sections~\ref{subsec: app: proof convergence} and~\ref{subsec: app: proof Riesz basis} we will need the complementary error function 
\( \textup{erfc}: \bbR \to \bbR\), \(x \mapsto \frac{2}{\sqrt{\pi}} \int_x^{\infty} e^{-t^2} \D t\),
which satisfies
\begin{equation}
\label{eq: growth of erfc}
    \lim_{t \to \infty} t^m \textup{erfc}(t) = 0, \quad \forall m \in \bbN.
\end{equation}

\subsection{Proof of Proposition~\ref{prop: algebraic convergence rate for C01}}
\label{subsec: app: proof algebraic convergence}

As a preliminary step, we show that 
\begin{equation}
\label{eq: bounds for ell1 and support}
    \max_{\bm{\gamma} \in \pi([s])} \nm{\bm{\gamma}}_{\ell^1(\bbN)} \leq s - 1 \leq s
    \quad \textup{and} \quad 
    \max_{\bm{\gamma} \in \pi([s])} \abs{\supp(\bm{\gamma})} \leq s - 1 \leq s.
\end{equation}
We start with the \(\ell^1\)-norm. Let \( \bm{\gamma}^* \in \pi([s]) \) with \( \nm{\bm{\gamma}^*}_{\ell^1(\bbN)} = t \geq s \). By Assumption~\ref{ass: properties of b}, we have \( \lambda_{\bm{b}, i} \leq \lambda_{\bm{b}, 1} \) for every \(i\). Hence,
\[
u_{\bm{\gamma}^*} = \left( 1 + \sum_{i = 1}^{\infty} \frac{\gamma_i^*}{\lambda_{\bm{b}, i}} \right)^{-1/2}
\leq \left( 1 + \frac{\nm{\bm{\gamma^*}}_{\ell^1(\bbN)}}{\lambda_{\bm{b}, 1}} \right)^{-1/2}
< \left( 1 + \frac{j}{\lambda_{\bm{b}, 1}} \right)^{-1/2}
= u_{j \bm{e_1}}, \quad \forall j = 0, \dots, s-1,
\]
where \( \bm{e_1} = (\delta_{1,j})_{j = 1}^{\infty} \in \Gamma \). Consequently, there exist at least \(s\) elements \( \bm{\gamma} \in \Gamma \) whose weights \( u_{\bm{\gamma}} \) exceed \( u_{\bm{\gamma}^*} \). It follows that \( \bm{\gamma}^* \not\in \pi([s]) \), as required. 
The bound on the maximum support in~\eqref{eq: bounds for ell1 and support} now follows from \( \absd{\supp(\bm{\gamma)}} \leq \nm{\bm{\gamma}}_{\ell^1(\bbN)} \).

Suppose now that Assumption~\ref{ass: Hermite expansion with algebraic rate} is true. Let \(F \in C^{0, 1}(\cX; \cY)\) with \(F \neq 0\) (otherwise there is nothing to show). We define the rescaled operator \(\wtF := r^{-1} \nm{F}_{C^{0, 1}(\cX; \cY)}^{-1} F\), where \(r := \max\{ 1, \sqrt{\dim(\cY)} \cdot \nm{\bm{b}}_{\ell^{\infty}(\bbN)}\}\) if \(\cY\) is finite-dimensional and \(r := \max\{1, \nm{\bm{b}}_{\ell^2(\bbN)}\}\) if \(\cY\) is infinite-dimensional. This implies \(\wtF \in B_{\mu, \bm{b}}^{\Lip}(\cX; \cY)\) by Theorem~\ref{thm: Lipschitz operators are Gaussian Sobolev operators}.
Next, let \(\bar{s} = \bar{s}(F)\) be the constant in Assumption~\ref{ass: Hermite expansion with algebraic rate} and set \(\bar{\epsilon} := \min\{ \bar{s}^{- \beta / 2}, e^{- \beta / (\beta + 1)} \}\) so that \(\bar{\epsilon}^{(\beta + 1) / \beta} = \min\{ \bar{s}^{- (\beta + 1) / 2}, e^{-1} \}\). Let \(0 < \epsilon \leq \bar{\epsilon}\) be arbitrary and choose \(s := \lceil \epsilon^{-2 / \beta} \rceil \) so that \( s \geq \bar{s} \) and \( \epsilon \leq 2^{\beta / 2} s^{- \beta / 2} \leq 2^{\beta / 2}  \epsilon \).

Recall from~\eqref{eq: polynomial expansion of L2 operators} that the truncated Wiener-Hermite PC expansion \(\wtF_{\pi([s])}\) is defined via the Hermite polynomials \(H_{\bm{\pi(1)}, d_1}, \dots, H_{\bm{\pi(s)}, d_s}\) with \(d_i := \max \{j: j \in \supp(\bm{\pi(i)}) \}\). 
Notice that we can interpret each \(H_{\bm{\pi(i)}, d_i}\) as a function \(H_{\bm{\pi(i)}, \infty}\) on \(\bbR^{\bbN}\) (simply by ignoring all components \(x_i\) with \(i \not \in \supp(\bm{\gamma})\)). Observe in addition that the set \( \pi([s]) \) is downward closed, i.e., if \( \bm{\gamma} \in \pi([s]) \), then \( \bm{\gamma'} \in \pi([s]) \) for every \( \bm{\gamma'} \leq \bm{\gamma} \).
Together with~\eqref{eq: bounds for ell1 and support} we can thus directly apply~\citet[Thm. 3.9]{schwab_DeepLearningHighDimension_2023} to conclude that there exists a ReLU neural network \(\psi = (\psi_1, \dots, \psi_s): \bbR^d \to \bbR^s\) with \(d := \max\{ d_1, \dots, d_s \}\) and with each \(\psi_i\) depending solely on the variables \(x_1, \dots, x_d \) such that  
\begin{equation}
\label{eq: NN bounds for Hermite polynomials}
    \max_{i \in [s]} \nm{H_{\bm{\pi(i)}, \infty} - \psi_i}_{L_{\mu_{\infty}}^2(\bbR^{\bbN})} 
    \leq \epsilon^{(\beta + 1)/\beta}.
\end{equation}
Here, we interpret the \(\psi_i\) as functionals on \(\bbR^{\bbN}\) by ignoring all components \(x_i\) with \(i > d\). Moreover,
\begin{equation}
\label{eq: size bound for PCA-Net}
    \textup{size}(\psi) \leq C' s^6 \log(s) \log(\epsilon^{- (\beta + 1)/\beta})
    \leq 2^7 C' \left(1 + \frac{1}{\beta} \right)\epsilon^{- 14/\beta - 1}
\end{equation}
with some global constant \( C' > 0 \).
We now define the encoder and decoder mappings
\begin{align*}
    \widetilde{\cE}_{\cX}: \cX \to \bbR^d, \quad 
    \widetilde{\cE}_{\cX}(X) := \left( \frac{\ip{X}{\phi_1}_{\cX}}{\sqrt{\lambda_1}}, \dots,  \frac{\ip{X}{\phi_d}_{\cX}}{\sqrt{\lambda_d}} \right), \\
    \widetilde{\cD}_{\cY}: \bbR^s \to \cY, \quad
    \widetilde{\cD}_{\cY}(\bm{x}) := \sum_{i = 1}^s \left( \int_{\cX} \wtF H_{\bm{\pi(i)}, \bm{\lambda}} \D \mu \right) x_i.
\end{align*}
The corresponding (rescaled) PCA-Net \(\Psi := r \nm{F}_{C^{0, 1}(\cX; \cY)} \left( \widetilde{\cD}_{\cY} \circ \psi \circ \widetilde{\cE}_{\cX} \right)\) satisfies
\[
\nm{F - \Psi}_{L_{\mu}^2(\cX; \cY)}
\leq r \nm{F}_{C^{0, 1}(\cX; \cY)} \bigg( \underbrace{\nm{\wtF - \wtF_{\pi([s])}}_{L_{\mu}^2(\cX; \cY)}}_{=: T_1(F)} + \underbrace{\nm{\wtF_{\pi([s])} - \widetilde{\cD}_{\cY} \circ \psi \circ \widetilde{\cE}_{\cX}}_{L_{\mu}^2(\cX; \cY)}}_{=: T_2(F)} \bigg).
\]
By Assumption~\ref{ass: Hermite expansion with algebraic rate}, we can bound \(T_1(F)\) by
\(
T_1(F) \leq c(F) s^{-\beta} 
\leq c(F) \epsilon^2
\leq c(F) \epsilon,
\)
where \(c(F)\) is the constant in~\eqref{eq: Hermite expansion with algebraic rate}.
For \(T_2(F)\), we find by an application of~\eqref{eq: NN bounds for Hermite polynomials}, the Cauchy-Schwarz inequality, and Parseval's identity that
\begin{align*}
    T_2(F) &\leq \sum_{i = 1}^s \nm{\int_{\cX} \wtF H_{\bm{\pi(i)}, \bm{\lambda}} \D \mu}_{\cY} \nm{H_{\bm{\pi(i)}, \infty} - \psi_i}_{L_{\mu_{\infty}}^2 (\bbR^{\bbN})} \\
    &\leq \epsilon^{(\beta + 1)/\beta} s^{1/2} \left( \sum_{i = 1}^s \nm{\int_{\cX} \wtF H_{\bm{\pi(i)}, \bm{\lambda}} \D \mu}_{\cY}^2 \right)^{1/2} 
    \leq \sqrt{2} \epsilon \nmd{\wtF}_{L_{\mu}^2(\cX; \cY)} 
    \leq \sqrt{2} \epsilon.
\end{align*}
Altogether, we conclude that
\(
\nm{F - \Psi}_{L_{\mu}^2(\cX; \cY)}
\leq r \nm{F}_{C^{0, 1}(\cX; \cY)} (c(F) + \sqrt{2}) \epsilon.
\)
Upon rescaling \(\epsilon\) and \(\bar{\epsilon}\) by the factor \(r \nm{F}_{C^{0, 1}(\cX; \cY)} (c(F) + \sqrt{2})\), the claim follows in light of~\eqref{eq: size bound for PCA-Net} with \(\alpha = \beta / (14 + \beta)\). \hfill \qedsymbol{}

\subsection{Proof of Lemma~\ref{lem: convergence in W^{1,2}}}
\label{subsec: app: proof convergence}

Let \(\bm{\gamma} \in \Gamma\) with \(\supp(\bm{\gamma}) \subset [d]\) for some \(d \in \bbN\). We first consider convergence in \(L_{\mu}^2(\cX)\). To this end, recall the \(d\)-dimensional Hermite polynomials \(H_{\bm{\gamma}, d} \) defined in~\eqref{eq: definition of Hermite polynomials on bbR^d}. By Fubini's theorem and a change of variables, one has 
\[
\nm{\wtH_{\bm{\gamma}, R, \bm{\lambda}} - H_{\bm{\gamma}, \bm{\lambda}}}_{L_{\mu}^2(\cX)}^2
= \int_{\bbR^d} \abs{\wtH_{\bm{\gamma}, R, d}(\bm{x}) - H_{\bm{\gamma}, d}(\bm{x})}^2
\D \mu_d(\bm{x}).
\]
It thus suffices to show that 
\begin{equation}
\label{eq: L^2 convergence of H_{gamma, d}}
    \lim_{R \to \infty} \wtH_{\bm{\gamma}, R, d} = H_{\bm{\gamma}, d} \quad \textup{in } L_{\mu_d}^2(\bbR^d), \hspace{1ex} \forall d \in \bbN.
\end{equation}
For this, we use induction on \(d\) and start with \(d = 1\). For \(n \in \bbN_0\), we compute
\begin{align*}
    &\int_{\bbR} \abs{\wtH_{n, R}(x) - H_n(x)}^2 \D \mu_1(x) \\
    &= \int_{-\infty}^{-R} \abs{H_n(-R) - H_n(x)}^2 \D \mu_1(x)
    + \int_{R}^{\infty} \abs{H_n(R) - H_n(x)}^2 \D \mu_1(x) \\
    &\leq \left(H_n(-R)^2 + H_n(R)^2\right) \textup{erfc}\left(\frac{R}{\sqrt{2}}\right) + 2 \int_{[-R,R]^c} H_n(x)^2 \D \mu_1(x) \\
    &=: T_1(R) + T_2(R),
\end{align*}
where we used the notation \([-R, R]^c := \bbR \setminus [-R, R]\). By~\eqref{eq: growth of erfc}, we have \(\lim_{R \to \infty} T_1(R) = 0\). Since \(H_n \in L_{\mu_1}^2(\bbR)\), the second term \(T_2(R)\) converges to zero as \(R \to \infty\) by the dominated convergence theorem.
Next, let \(d > 1\) and suppose that~\eqref{eq: L^2 convergence of H_{gamma, d}} holds for any \(1 \leq d' < d\). Without loss of generality we may assume \(R \geq 1\). For \( \bm{x} \in \bbR^d \), let us write \( \bm{x}_{[d-1]} = (x_1, \dots, x_{d-1}) \in \bbR^{d-1} \). Then, by Fubini's theorem, we have
\begin{align*}
    &\int_{\bbR^d} \abs{\wtH_{\bm{\gamma}, R, d}(\bm{x}) - H_{\bm{\gamma}, d}(\bm{x})}^2 \D \mu_d(\bm{x}) \\
    &\leq 2\int_{\bbR} \int_{\bbR^{d-1}} \left| \wtH_{\bm{\gamma}, R, d-1}(\bm{x}_{[d-1]}) \wtH_{\gamma_d,R}(x_d)  - \wtH_{\bm{\gamma}, R, d-1}(\bm{x}_{[d-1]}) H_{\gamma_d}(x_d) \right|^2 \D \mu_{d-1}(\bm{x}_{[d-1]}) \D \mu_1(x_d) \\
    &\hspace{2.5ex} + 2\int_{\bbR} \int_{\bbR^{d-1}} \left| \wtH_{\bm{\gamma}, R, d-1}(\bm{x}_{[d-1]}) H_{\gamma_d}(x_d)   - H_{\bm{\gamma}, d-1}(\bm{x}_{[d-1]}) H_{\gamma_d}(x_d) \right|^2 \D \mu_{d-1}(\bm{x}_{[d-1]}) \D \mu_1(x_d) \\
    & =: t_1(R) + t_2(R).
\end{align*}
The term \(t_1(R)\) can be bounded from above as 
\begin{multline*}
    t_1(R)
    \leq 2  \sup_{R \geq 1} \int_{\bbR^{d-1}} \abs{\wtH_{\bm{\gamma}, R, d-1}(\bm{x}_{[d-1]})}^2 \D \mu_{d-1}(\bm{x}_{[d-1]})   \int_{\bbR} \abs{\wtH_{\gamma_d,R}(x_d) - H_{\gamma_d}(x_d)}^2 \D \mu_1(x_d).
\end{multline*}
By induction hypothesis for \(d' = d-1\), the term \(\wtH_{\bm{\gamma}, R, d-1}\) converges in \(L_{\mu_{d-1}}^2(\bbR^{d-1})\) as \(R \to \infty\). Hence, the supremum over \(R \geq 1\) of the first integral is finite. Applying the induction hypotheses for \(d' = 1\), we conclude that the second integral over \(\bbR\) converges to zero as \(R \to \infty\).
A similar argument shows that \(\lim_{R \to \infty} t_2(R) = 0\). This completes the proof of~\eqref{eq: L^2 convergence of H_{gamma, d}}.

Next, we consider convergence of \(\nabla_{\cX_{\bm{b}}}\wtH_{\bm{\gamma}, R, \bm{\lambda}}\) in \(L_{\mu}^2(\cX; \cX_{\bm{b}})\). For this, recall the basis \(\{\eta_i\}_{i=1}^{\infty}\) of \(\cX_{\bm{b}}\) defined in~\eqref{eq: orthonormal basis of cX_b}.
Note that the capped Hermite polynomials \(\wtH_{n, R}\), as defined in~\eqref{eq: m-width: definition wtH_n,R}, are Lipschitz continuous. We write \(x_i := \lambda_i^{-1/2} \ip{X}{\phi_i}_{\cX}\) for \(X \in \cX\) and \(i \in [d]\). By Lemma~\ref{lem: Lipschitz to Gaussian Sobolev} with \( \cY = \bbR \), we have \(\wtH_{\bm{\gamma}, R, \bm{\lambda}} \in W_{\mu, \bm{b}}^{1,2}(\cX)\) and for \(\mu\)-a.e. \(X \in \cX\) and \( i \in [d] \) the weak partial derivatives are given by
\begin{align*}
    \frac{\partial}{\partial \eta_i} \wtH_{\bm{\gamma}, R, \bm{\lambda}}(X) 
    &= b_i \lambda_i^{-1/2} \partial_i \wtH_{\bm{\gamma}, R, d}(x_1, \dots, x_d)\\
    &=
    \begin{cases}
        b_i \lambda_i^{-1/2} \partial_i H_{\gamma_i}(x_i) \prod\limits_{\substack{j = 1 \\ j \neq i}}^d \wtH_{\gamma_j, R}(x_j) & \textup{if } x_i\in [-R,R], \\
        0 & \textup{if } x_i \in [-R,R]^c.
    \end{cases}
\end{align*}
Moreover, \(\frac{\partial}{\partial \eta_i} \wtH_{\bm{\gamma}, R, \bm{\lambda}} = \frac{\partial}{\partial \eta_i} H_{\bm{\gamma}, \bm{\lambda}} = 0\) for \(i > d\). Consequently,
\begin{align*}
    &\nm{\nabla_{\cX_{\bm{b}}} \wtH_{\bm{\gamma}, R, \bm{\lambda}} - \nabla_{\cX_{\bm{b}}} H_{\bm{\gamma}, \bm{\lambda}}}_{L_{\mu}^2(\cX; \cX_{\bm{b}})}^2
    = \int_{\cX} \sum_{i = 1}^d \abs{\frac{\partial}{\partial \eta_i} \wtH_{\bm{\gamma}, R, \bm{\lambda}}(X) - \frac{\partial}{\partial \eta_i} H_{\bm{\gamma}, \bm{\lambda}}(X)}^2 \D \mu(X) \\
    & = \sum_{i = 1}^d \int_{\bbR^{i-1} \times [-R,R] \times \bbR^{d-i}} 
    b_i^2 \lambda_i^{-1} \absd{\partial_i H_{\gamma_i}(x_i)}^2 \bigg| \prod_{\substack{ j = 1 \\ j \neq i}}^d  \wtH_{\gamma_j, R}(x_j) - \prod_{\substack{ j = 1 \\ j \neq i}}^d H_{\gamma_j}(x_j) \bigg|^2 \D \mu_d(\bm{x}) \\
    & \quad + \sum_{i = 1}^d \int_{\bbR^{i-1} \times [-R,R]^c \times \bbR^{d-i}} 
    b_i^2 \lambda_i^{-1} \abs{\partial_i H_{\bm{\gamma}, d}(\bm{x})}^2 \D \mu_d(\bm{x}).
\end{align*}
Since \( \partial_i H_{\gamma_i} \in L^2_{\mu_1}(\bbR) \), the first term on the right-hand side converges to zero as \( R \to \infty \) by~\eqref{eq: L^2 convergence of H_{gamma, d}} and the dominated convergence theorem.
Similarly, the second term converges to zero as \(R \to \infty\) by the dominated convergence theorem. \hfill \qedsymbol{}

\subsection{Proof of Lemma~\ref{lem: Riesz basis}}
\label{subsec: app: proof Riesz basis}

We first work in \(d = 1\) dimension and compute for \(n, m \in \bbN_0\),
\begin{align*}
    &\ip{\wtH_{n,R}}{\wtH_{m,R}}_{L_{\mu_1}^2(\bbR)} 
    = \int_{\bbR}\wtH_{n,R}(x) \wtH_{m,R}(x) \D \mu_1(x) \\
    &= \int_{-R}^R H_{n}(x) H_{m}(x) \D \mu_1(x)
    + \int_{-\infty}^{-R} H_n(-R) H_m(-R) \D \mu_1(x) 
    + \int_R^{\infty} H_n(R) H_m(R) \D \mu_1(x) \\
    &= \int_{-R}^R H_{n}(x) H_{m}(x) \D \mu_1(x)
    + \frac{1}{2} H_n(-R) H_m(-R) \textup{erfc}\left(\frac{R}{\sqrt{2}}\right)
    + \frac{1}{2} H_n(R) H_m(R) \textup{erfc}\left(\frac{R}{\sqrt{2}}\right) \\
    &=: T_1(R) + T_2(R) + T_3(R).
\end{align*}
Note that \(H_n H_m\) is a polynomial of order \(n+m\). Hence, by~\eqref{eq: growth of erfc}, we deduce \(\lim_{R \to \infty} T_2(R) = 0\) and \(\lim_{R \to \infty} T_3(R) = 0\). Moreover, the dominated convergence theorem and orthonormality of the Hermite polynomials imply that
\[
\lim_{R \to \infty} T_1(R) = \int_{-\infty}^{\infty} H_n(x) H_m(x) \D \mu_1(x) = \delta_{n,m},
\]
where \( \delta_{n,m} \) denotes the Kronecker delta.
Altogether, 
\begin{equation}
\label{eq: inner product convergence in 1D}
    \lim_{R \to \infty} \ip{\wtH_{n,R}}{\wtH_{m,R}}_{L_{\mu_1}^2(\bbR)} = \delta_{n, m}.
\end{equation}
Now, let \(\bm{\gamma}, \bm{\gamma'} \in \Gamma\) with \(\supp(\bm{\gamma})\), \(\supp(\bm{\gamma'}) \subset [d]\) for some \(d \in \bbN\). Since
\begin{align*}
    \ip{\wtH_{\bm{\gamma}, R, \bm{\lambda}}}{\wtH_{\bm{\gamma'}, R, \bm{\lambda}}}_{L_{\mu}^2(\cX)} 
    &= \int_{\bbR^d} \wtH_{\bm{\gamma}, R, d}(\bm{x}) \wtH_{\bm{\gamma'}, R, d}(\bm{x}) \D \mu_d(\bm{x}) \\
    &= \prod_{i=1}^d \int_{\bbR}\wtH_{\gamma_i, R}(x_i) \wtH_{\gamma_i', R}(x_i) \D \mu_1(x_i),
\end{align*}
we conclude by~\eqref{eq: inner product convergence in 1D} that
\[
\lim_{R \to \infty} \ip{\wtH_{\bm{\gamma}, R, \bm{\lambda}}}{\wtH_{\bm{\gamma'}, R, \bm{\lambda}}}_{L_{\mu}^2(\cX)} 
= \delta_{\bm{\gamma}, \bm{\gamma'}} := \prod_{i = 1}^{\infty} \delta_{\gamma_i, \gamma'_i}.
\]
Next, fix some arbitrary \(\varepsilon > 0\). Then there exists \(\bar{R} > 0\) such that for every \(R \geq \bar{R}\), we have 
\[
\abs{\ip{\wtH_{\bm{\gamma}, R, \bm{\lambda}}}{\wtH_{\bm{\gamma'}, R, \bm{\lambda}}}_{L_{\mu}^2(\cX)} - \delta_{\bm{\gamma}, \bm{\gamma'}}} \leq \varepsilon.
\]
Since \(I \subset \Gamma\) is finite, we obtain for any \(\bm{x} = (x_{\bm{\gamma}})_{\bm{\gamma} \in I} \in \bbR^I\),
\begin{align*}
    &\nm{\sum_{\bm{\gamma} \in I} x_{\bm{\gamma}} \wtH_{\bm{\gamma}, R,  \bm{\lambda}}}_{L_{\mu}^2(\cX)}^2 \\
    &= \sum_{\bm{\gamma} \in I} \sum_{\substack{\bm{\gamma'} \in I \\ \bm{\gamma'} \neq \bm{\gamma}}} x_{\bm{\gamma}} x_{\bm{\gamma'}} \underbrace{\ip{\wtH_{\bm{\gamma}, R, \bm{\lambda}}}{\wtH_{\bm{\gamma'}, R, \bm{\lambda}}}_{L_{\mu}^2(\cX)}}_{\leq\varepsilon}
    + \sum_{\bm{\gamma} \in I} x_{\bm{\gamma}}^2 \underbrace{\ip{\wtH_{\bm{\gamma}, R, \bm{\lambda}}}{\wtH_{\bm{\gamma}, R, \bm{\lambda}}}_{L_{\mu}^2(\cX)}}_{\leq 1 + \varepsilon} \\
    &\leq \varepsilon \nm{\bm{x}}_{\ell^1(I)}^2 + (1+\varepsilon) \nm{\bm{x}}_{\ell^2(I)}^2 \leq (\varepsilon \abs{I} + 1+ \varepsilon) \nm{\bm{x}}_{\ell^2(I)}^2.
\end{align*}
Similarly, we have
\begin{align*}
    \nm{\sum_{\bm{\gamma} \in I} x_{\bm{\gamma}} \wtH_{\bm{\gamma}, R, \bm{\lambda}}}_{L_{\mu}^2(\cX)}^2 
    \geq -\varepsilon \nm{\bm{x}}_{\ell^1(I)}^2 + (1-\varepsilon) \nm{\bm{x}}_{\ell^2(I)}^2 
    \geq (-\varepsilon \abs{I} + 1 - \varepsilon) \nm{\bm{x}}_{\ell^2(I)}^2.
\end{align*}
As \(\varepsilon > 0\) was arbitrary, the claim follows. \hfill \qedsymbol{}

\end{document}